\documentclass[10pt,letterpaper,twocolumn]{article}
\usepackage[algorithms]{wacv}      
\usepackage{pgf}
\usepackage{tikz}
\usepackage{tikzscale}
\usetikzlibrary{arrows,external}
\usetikzlibrary{spy}
\usepackage{tabularx}
\usepackage{booktabs}
\usepackage{subcaption}
\usepackage{amsmath}
\usepackage{pgfplots}
\usetikzlibrary{pgfplots.groupplots}
\usetikzlibrary{backgrounds}
\usetikzlibrary{positioning}
\graphicspath{{images/}} 
\usepackage{multirow}
\usepackage{rotating}
\usepackage{float}
\usepackage{svg}
\usepackage[font=small,skip=3pt]{caption}

\usepackage{graphicx}
\usepackage{amsmath}
\usepackage{amssymb}
\usepackage{booktabs}

\def\addlegendimage{\csname pgfplots@addlegendimage\endcsname}

\usepackage[T1]{fontenc}

\usepackage[pagebackref,breaklinks,colorlinks]{hyperref}

\usepackage[capitalize]{cleveref}
\crefname{section}{Sec.}{Secs.}
\Crefname{section}{Section}{Sections}
\Crefname{table}{Table}{Tables}
\crefname{table}{Tab.}{Tabs.}

\def\confYear{2025}
\begin{document}
\setlength{\belowdisplayshortskip}{0pt}
\title{
      Curvy: A Parametric Cross-section based Surface Reconstruction}

\author{Aradhya N. Mathur\\
IIITD\\
{\tt\small aradhyam@iiitd.ac.in}
\and
Apoorv Khattar\\
% \\
The University of Manchester\\
{\tt\small  apoorv.khattar@postgrad.manchester.ac.uk}
\and
Dr. Ojaswa Sharma\\
% \\
IIITD\\
{\tt\small  ojaswa@iiitd.ac.in}
}
\maketitle
%%%%%%%%% ABSTRACT
\begin{abstract}
   In this work, we present a novel approach for reconstructing shape point clouds using planar sparse cross-sections with the help of generative modeling. We present unique challenges pertaining to the representation and reconstruction in this problem setting. Most methods in the classical literature lack the ability to generalize based on object class and employ complex mathematical machinery to reconstruct reliable surfaces. We present a simple learnable approach to generate a large number of points from a small number of input cross-sections over a large dataset. We use a compact parametric polyline representation using adaptive splitting to represent the cross-sections and perform learning using a Graph Neural Network to reconstruct the underlying shape in an adaptive manner reducing the dependence on the number of cross-sections provided.
\end{abstract}

%%%%%%%%% BODY TEXT
\section{Introduction}
\label{sec:intro}

Surface reconstruction from cross-sections is a well-explored problem. There is a rich literature on methods demonstrating the generation of reliable surfaces from cross-sections. Little work exists that provides insights into how complex objects could be generated using cross-sections with the help of deep learning methods that could provide an added advantage of capturing semantic context associated with shapes. Deep learning-based methods can provide better generalizability qualities associated with unseen shapes of similar types. Unlike traditional methods of surface generation from incomplete point clouds, the problem of surface reconstruction from cross-sections brings unique challenges that we aim to address in this paper. Previous approaches for point cloud completion have focused on generating point clouds from images or representation learning using autoencoders. Several previous methods focused on generating surfaces using cross-sections and did not involve any learning based on the class of objects. Our method can be used with any modern encoder-decoder-based point cloud generation since it focuses on learning the latent embeddings rather than generating the point cloud directly. 
%-------------------------------------------------------------------------
Our approach introduces a novel input representation for the cross-sections, aiming to capture crucial information that would be overlooked when using surface-sampled points. Point clouds, while dense in most areas, often suffer from incomplete information in certain regions. In contrast, cross-section curves exhibit a highly non-uniform distribution of information, necessitating reconstruction methods capable of handling sparse and anisotropic data. By considering this unique characteristic of cross-sections, our approach enables a more comprehensive and accurate reconstruction of shapes. Our contributions can be summarised as follows:
% \vspace{-20pt}
\begin{enumerate}
    \item An approach for learning surface reconstruction based on parametric representation of cross-sections,
    \item A novel framework for generating a point cloud while adapting to the anisotropic and sparse nature of input cross-sections. This constitutes two attention mechanisms to focus on the local and global structure of the cross-sections and show their significance empirically through an ablation study, and
    % \item A graph neural network to capture the local adjacency of cross-sections
    \item  A new dataset for parametric representation of cross-sections. %The same can be used for experimenting with graph neural networks. %% to be refined
\end{enumerate}
%-------------------------------------------------------------------------
\section{Related work}
Surface reconstruction is a widely studied problem in computer graphics. As methods for representing 3D data change, so do the methods for shape reconstruction. The different methods for representing 3D data include a voxel-based representation that gives information pertaining to points in a discrete grid, point clouds that contain the locations of information, and meshes that have added neighborhood information in the form of an adjacency matrix corresponding to the points. Newer implicit methods directly target surface generation by learning to produce the implicit field functions. 
We divide this section based on the representation of the output for different methods. 
% \vspace{-15pt}
\subsection{Pointcloud generation}
There are two primary approaches that have been explored for point cloud reconstruction. Reconstruction of point clouds has been done using multi-view/single-view images and partial-point clouds.

A deep autoencoder network for the reconstruction of point clouds results in compact representations and can perform semantic operations, interpolations, and shape completion ~\cite{achlioptas2018learning},\cite{lin2018learning}. These networks leverage 1-D and 2-D convolutional layers to extract latent representation for the generation of point clouds.
Single image point cloud generation has also been performed hierarchically from low resolution by gradually upsampling the point cloud as explored in \cite{fan2017point}. This multi-stage process uses EMD distance \cite{fan2017point} and computes Chamfer distance for the later stages w.r.t. ground truth dense point cloud. Another approach uses a multi-resolution tree-structured network that allows to process point clouds for 3D shape understanding and generation \cite{gadelha2018multiresolution}. 
%The method chooses a locality-preserving 1D ordered list of points at multiple resolutions to represent a 3D shape that allows performing efficient feed-forward processing leading to faster convergence and a smaller memory footprint. 
Some newer methods also approach this problem from a local supervision perspective to understand the local geometry better \cite{han2019multi}.
Further, skip-attention has shown to play an important role in tasks such as point cloud completion \cite{wen2020point}. The architecture proposed consists of primarily three parts - a point cloud encoder,
%an encoder that learns the local shapes from the incomplete point cloud,
a decoder that generates the point cloud, and skip-attention layers that fuse relevant features from the encoder to the decoder at different resolutions. 
%It introduces a folding block that allows the decoder to generate consistent local geometry in the generated point cloud.
Reinforcement learning has also been explored with GANs trained for point cloud generation. The agent is trained to predict a good seed value for the adversarial reconstruction of incomplete point clouds \cite{sarmad2019rl}. The method uses an autoencoder trained on complete point clouds to generate the global feature vector (GFV) and a GAN that is trained to produce GFV. The pipeline uses GFV generated from an incomplete point cloud as a state and supplies it to an RL agent which the GAN uses to generate GFV close to the GFV of a complete point cloud. 

\subsection{Surface reconstruction}
% \subsubsection{Cross-Section Based Reconstruction}

One of the seminal works \cite{memari2008provably} proposes constructing 2D geometric shapes from 1D cross-sections. The method provides sampling conditions to guarantee the correct topology and closeness to the original shape for the Hausdorff distance. 
 One of the early works \cite{HUANG2002149} proposes a manifold mesh reconstruction method from unorganized points with arbitrary topology. The method proposed defines a two-step process for reliably reconstructing the geometric shape from an unorganized point cloud sampled from its surface.
 %In the first step, the method reconstructs the triangle mesh as a continuous manifold. In the next stage, operations are performed to obtain reliable curvature estimation for the reconstructed mesh surface.
 % Several other methods rely on the deformation of an existing mesh to match the required shape \cite{kanazawa2018learning,li2020online}. 
 
Early works took inspiration from medical imaging problems. A two-step process for the reconstruction of a surface from cross-sections has been proposed by first computing the arrangement for the cross-section within each cell and then reconstructing an approximation of the object. This is done by performing its intersection with the cell boundary and gluing the pieces back together to yield the surface \cite{boissonnat2007shape}. An algorithm for non-parallel cross-sections consisting of curve networks of arbitrary shape and topology has also been developed \cite{liu2008surface}. 
% The method is guaranteed to generate a closed surface interpolating the curve network on each cross-section. Variational methods for generating smooth surfaces have also been explored recently. One method focuses on the two goals of achieving smoothness and good data fitting, where the energy function is chosen to search for the best Hermite data  \cite{huang2019variational}.  For the problem is defined by using Duchon's energy and minimizing it. 
Several methods propose implicit field-based reconstruction.  One such method utilizes sign agnostic learning for geometric shapes \cite{atzmon2020sal}. This method uses a deep learning-based approach that allows learning of implicit shape representations directly from unsigned raw data like point clouds and triangle soups. The proposed unsigned distance loss family possesses plane reproduction property based on suitable initialization of the network weights.

 % \subsection{Image to Mesh}
 % Mesh reconstruction has also been performed from single RGB images via a two-stream network \cite{pan2019deep}. The proposed method uses an image encoder that generates a latent vector which is then passed into another sub-network responsible for deforming a spherical mesh. It further introduces topological changes by pruning faces on the basis of error driven by per point error from the points sampled from the surface.

  The surface reconstruction method has also been performed with topological constraints \cite{lazar2018robust}. This method relies on computing candidates for cell partitioning of ambient volume. The method is based on the calculation of a single surface patch per cell so that the connected manifold surface of some topology is obtained. 3D surface reconstruction from unorganized planar cross-sections using a split-merge approach using Hermite mean-value interpolation for triangular meshes has also been used \cite{sharma2017signed}. 
  % The cross-section boundaries as considered as generators for the signed distance function. Further, the proposed method uses a globally consistent and continuous function defined over the arrangement of cutting planes in order to approximate the signed distance field of the original object. 
  A divide-and-conquer optimization-based strategy can also be employed to perform topology-constrained reconstruction \cite{zou2015topology}. New methods like Orex \cite{sawdayee2022orex} leverage deep learning for cross-section to surface generation.
  % It follows a three-step process to generate a surface. In step one,  the space is partitioned into convex cells and  part of the final surface in each cell is computed, which is called as  tile. In the next stage, the tile topology is selected such that the global surface has the user-specified genus. In step three, the surfacing is performed, generating a smooth surface. 

%-------------------------------------------------------------------------

\section{Approach}
% \begin{figure*}[!htp]
%     \centering
%     % \includesvg[width=0.9\linewidth]{./images/Pipeline.svg}
%     \includegraphics[width=0.9\linewidth]{images/Pipeline_hres2.png}
%     %\includegraphics[scale=0.75]{images/gcn_pipeline.pdf}
%     %\def\svgwidth{\linewidth}
%     %\input{images/gcn_recons_pipeline.tex}    
% \caption{Overview of our reconstruction approach. Starting from a parametric representation a network is trained to generate a surface point cloud.} %we begin by generating cross-sections using mesh-plane intersections. Then we sample a subset of cross-sections randomly and convert these into parametric representation. This parametric representation is then used to train a graph neural network to generate an embedding from which a point cloud can be decoded.}
%     \label{fig:pipeline}
% \end{figure*}
  \begin{figure*}[!h]
    %\includegraphics[width=\textwidth]{banner_test/teaser_final_labelled.png}
    % \includesvg[width=0.9\linewidth]{images/GCN.svg}
    \centering
    \includegraphics[width=1.0\textwidth]{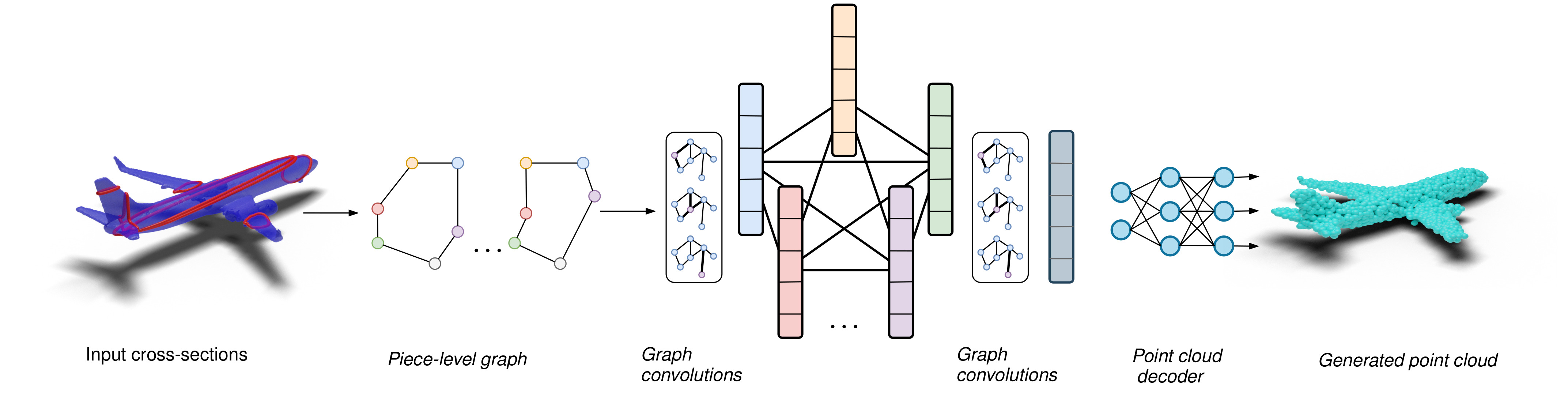}
    % \scriptsize \bf
    %     \includesvg[width=1.0\textwidth]{banner_test/gcn_blender_shadows_graph_only2.svg}
        % \includesvg[width=1.0\linewidth]{banner_test/gcn_blender_multi.svg}
    %\caption{We present a novel graph neural network-based approach for generating point clouds from parametric representation of cross-sections. The graph containing one-ring neighborhoods of pieces within a cross-section is passed through graph convolutions consisting of attention and further aggregation to obtain high-level representation for every cross-section. A new fully connected graph is then used to obtain the embedding from which the point cloud is recovered.}
    \caption{Overview of our reconstruction approach. Starting from a parametric representation of the given cross-sections, we train a network to generate a surface point cloud.}
    %\label{fig:teaser}
    \label{fig:pipeline}
  \end{figure*}
In this work, we develop an approach for shape reconstruction from a set of unorganized cross-sections. We design a deep neural network that learns the overall structure of various shapes and generates a point cloud representing the original object. 

Our approach can be defined as a three-step process. We first generate a large number of cross-sections from 3D models and sample them to create input cross-sectional data. Then surface points are sampled to generate a point cloud on which an autoencoder is trained to reconstruct the point cloud. In the final step, we use the encoded vector obtained from the autoencoder and train a Graph Neural Network on the parametric representation of input cross-sections to generate an embedding vector in a GAN-based setting to match the encoded vector for the same object. These embedding vectors can then be decoded to obtain the point cloud from the \emph{pre-trained} autoencoder network.

The cross-sections may be obtained as points sampled along a shape's boundary that can further be represented as a polyline. However, for complex cross-sections, we would want a representation that optimally captures the curvature-related information. Instead of sampling points, we convert an entire cross-section curve into its parametric representation. This allows us to reduce any loss of information that may occur due to sampling and further helps reduce the memory requirements needed to represent a large number of points in the network. Let's assume the density of points $\rho$ per unit length of a cross-section curve of length $l$. Depending on the sampling density $\rho$, the number of points in a curve can vary, and for better information capture we need a high $\rho$ value to capture the curvature accurately. We note that the parametric curve can be represented using a fixed number of coefficients from which any arbitrary density of points can be sampled. Our overall approach is shown in Figure ~\ref{fig:pipeline}.
The parametric curve fitting is further discussed in the supplementary.
% In this section, we first revisit the parametric representation of curves and understand how we use the parameters of a curve for representing the cross-sections. We then discuss in detail how the problem can be approached in a neural network-based setting and try to answer the nuances associated with the unique problem setting.

% \vspace{-10pt}

\subsection{Adaptive Splitting} \label{sec:ad_split}
It is important to ensure that a simpler piece (such as a straight line) is represented by fewer points so that more points can be assigned to a piece with many sharp turns. 
%Uniformly distributing points between all pieces does not ensure this hence,
We propose an adaptive splitting scheme for non-uniform distribution between pieces using the Douglas Peucker polyline simplification algorithm \cite{douglas1973algorithms} for finding a set of endpoints to generate the pieces within the curve. This helps to save more points for complex curves and uses fewer points for simpler curves further retaining more information than a uniform splitting scheme. Douglas Peucker algorithm is run for multiple iterations till the final number of unique endpoints returned is more than $k$, we select the $k$ points with maximum absolute angle, where the angle varies between $-90$ and $+90$. Once we have obtained $k$ pieces, we fit piece-wise polynomials as further discussed in the supplementary. %\ref{sec:curvefit}.
%For the sake of simplicity of this initial experimentation we fix $k$ to 32, the method can be extended to a variable number of pieces as well.

\subsection{Training on parametric space}

\begin{figure}[!h]
    \centering
    \scriptsize
    \includegraphics[width=0.9\linewidth]{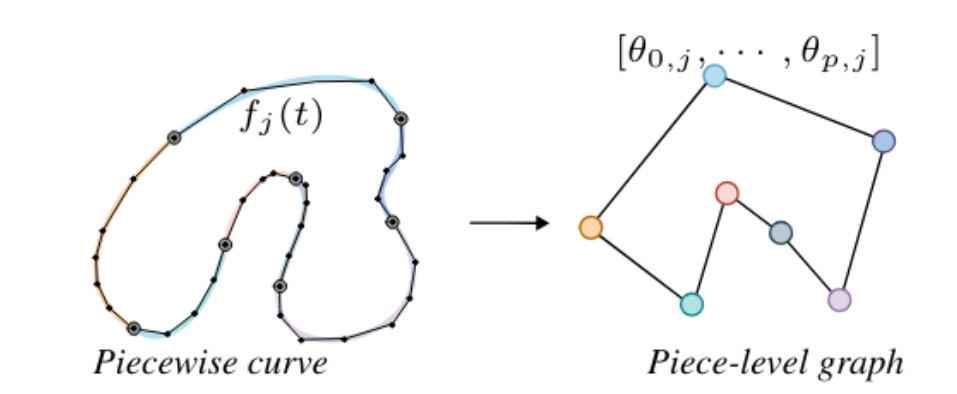}
    \caption{Converting a piecewise parametric representation of a cross-section (left) to a graph (right). The nodes in the graph are matrices of coefficients of the parametric functions.}
    \label{fig:parametric_graph}
\end{figure}

We take the ShapeNet dataset \cite{chang2015shapenet} and use the manifold meshes. The input cross-sections are generated using mesh-plane intersection 
%provided by Trimesh \cite{trimesh} 
and converted to parametric representation. Further in the text, \textit{cross-sections} shall refer to the parametric representation of cross-sections. We sample surface points from the meshes; thus, each set of cross-sections and the corresponding point clouds form the input and the corresponding ground truth for the network. In order to use parameters with a neural network there are certain properties that the operations on the parametric representation must possess. Each piece of a cross-section is represented as a tensor in $\mathbb{R}^{6\times 3}$ of coefficients of the parametric representation $f_j(t)$ of degree 5 %($p=5$)
in $\mathbb{R}^3$. See Figure~\ref{fig:parametric_graph} for our parametric curve representation and its corresponding graph. %Further, we propose that any operation performed on these coefficients must hold permutation invariance in both piece level and cross-section level operations. 

\subsubsection{Permutation Invariance and Neighborhoods}
We represent the coefficients of the parametric representation as a vector for the neural network to act on. Thus, the cross-sections are represented as tensors containing the vector for each parametric piece. Further, the cross-sections contain neighborhood information in the form of adjacency of the pieces.  

\par Therefore, the operations that we perform on the cross-sections must be permutation invariant since any combination of cross-sections represents the same object.
%%%
% \begin{equation}
%     f([x_i, x_j, x_k]) = const \; \forall \; combinations\;of\; i,j,k
% \end{equation}
% where $f$ is the network and $x_i, x_j, x_k$ are input cross-sections. Furthermore, within the cross-sections this property should hold at the level of pieces.
Given a set of $m$ parameterized cross-sections where each cross-section is partitioned into $k$ pieces with the coefficient matrix $\Theta_l$ of the parametric functions for the $l^{th}$ piece, the full set of stacked coefficients for the entire set of cross-sections are represented as the tensor
    $\mathcal{C} = \begin{bmatrix}
    \Theta_1, \Theta_2, \cdots, \Theta_m
    \end{bmatrix}^\intercal$ 
of size $m\times (p+1)k\times 3$.
Any permutation of rows of $\mathcal{C}$ still represents the same set of cross-sections (that is to say that the cross-sections can come in any order) and any circular permutation of these pieces represents the same cross-section. Therefore, any operation performed on $\mathcal{C}$ should ideally yield the same result irrespective of the ordering of its rows and any circular permutation within each row.
%Similarly, at the level of each cross-section, we have $k$ pieces spliced together and any circular permutation of these pieces represents the same cross-section.
%Therefore, we need another level of permutation invariance for the representation at the level of each cross-section. 
Within a neural network, representations are created using matrix multiplications, and different orders of the rows and columns of $\mathcal{C}$ would produce different results since,
% \vspace{-5pt}
\begin{equation}
    W^\intercal \mathcal{C} \neq W^\intercal S'(\mathcal{C}),\nonumber
\end{equation}
% \vspace{-5pt}
where $W$ is a weight matrix and $S'$ is a shuffle operation. Therefore we do away with this matrix-based representation. We create a graph-based representation using the piecewise parametric representation. We note that each cross-section has some adjacency information since the pieces of a cross-section are arranged in linear order along the contour. In order to use the neighborhood properties, we propose a graph-based representation, where each node is represented as the matrix of coefficients of a piece of the parametric curve and each edge denotes the adjacency. The graph-based representation allows our approach to take into account the desired permutation invariance while enabling us to use the additional adjacency information as needed.
%We further use a 2-level attention mechanism while changing the adjacency for piece level and cross-section level learning. 
Therefore, our final representation %is thus in terms of
uses coefficients of the pieces where the adjacency matrix stores the piece-level relations. 

\begin{figure}[!htp]
    \centering
    
    % \includesvg[width=1.\linewidth]{images/Architecture_new.svg}
    % \includegraphics[width=0.8\linewidth]{images/architecture.png}
    \includegraphics[width=0.9\linewidth]{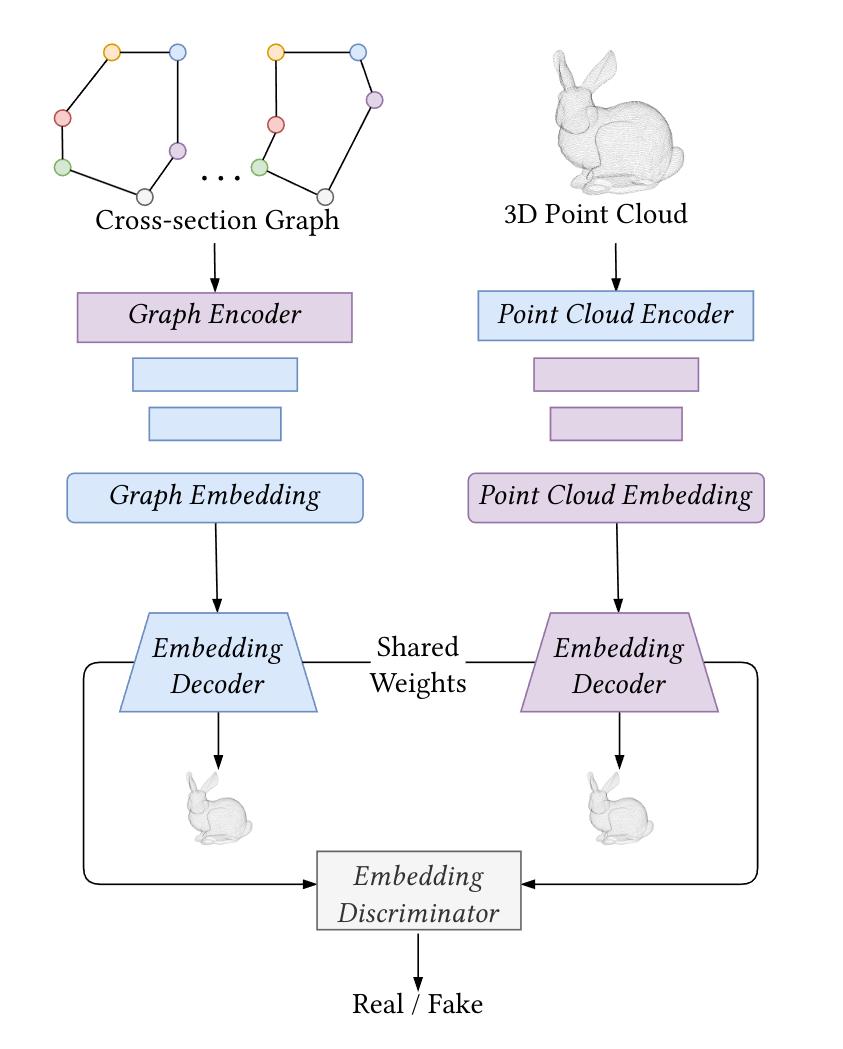}
    \caption{During training, the graph embedding decoder tries to generate an embedding that is similar to the point cloud embedding generated from the pre-trained encoder. This representation is then used by the decoder to generate the point cloud of a relevant shape.}
    \label{fig:train}
\end{figure}
\vspace{-7pt}
\subsubsection{Learning Point Cloud Representation}

%Graph neural networks suffer from the problem of over-smoothing due to repetitive aggregation operations.

%In our experiments, we initially experimented with a learnable decoder that was trained jointly with the graph encoder; however, that did not yield very detailed results.
% further leading to a mode collapse in the GAN based setting.
%Therefore, to circumvent this problem we found that a pre-trained model is much more effective. 

We train a point cloud auto-encoder on the ground truth point cloud generated by sampling 2048 points from the manifold meshes and then use the encoder embedding from this as the ground truth embedding similar to \cite{sarmad2019rl}, whereby a GAN is used to generate an embedding similar to that of a pre-trained point cloud auto-encoder which is very stable for training while allowing for stochasticity. Thus, the objective of the graph encoder is to learn the embedding from the cross-sections to produce a similar point cloud from the pre-trained decoder, as shown in Figure ~\ref{fig:train}. 
%Thus, the generator tries to produce an embedding that is similar to the embedding generated by the point cloud encoder for a given shape.
\vspace{-5pt}
\subsubsection{Cross-Section Attention}
Attention mechanism \cite{vaswani2017attention} allows a network to focus on different features and enables better learning of the network. %Taking inspiration from this, we introduce two new kinds of attention in our network for learning shapes.
Taking inspiration from this, we introduce attention at two levels in our network for learning shapes.
We use two levels of cross-section attention mechanism, which we call \emph{global attention}, and a piece-wise attention mechanism for focusing on local information. Each cross-section contains different amounts of information pertaining to the geometric shape of the object.
%Some cross-sections carry more salient features about the entire object than others. 
Similarly, within a cross-section, some pieces contain more information pertaining to the local regions, such as regions of high curvature. In order to focus on such regions, we introduce \emph{local attention}, which attends to each piece within a cross-section.
%The input to our network has the dimensionality of $B \times C \times Q \times P \times 3$ where $B$ is the batch size, $C$ is the number of cross-sections, $Q$ is number of pieces per cross-section and $P$ is the number of points per cross-section. Thus, global attention is defined by computing attention maps across $C$ while the local attention computed over the pieces $Q$.
The \emph{global} and \emph{local} attention are computed using Graph Attention \cite{velivckovic2017graph}. The normalized attention coefficient at the graph level can be expressed as $ \alpha_{i,j} = softmax(e_{i,j})$ where $\alpha_{i,j}$ are the normalized attention coefficients for node $i$ in the graph, $j\in\mathcal{N}_i$ where $\mathcal{N}_i$ is the neighbourhood of node $i$ and $e_{i,j}$ is the attention coefficient. The attention coefficient is calculated using the same method as described in \cite{velivckovic2017graph}.

First, attention is computed locally over the pieces of each cross-section, which we then aggregate into a single vector to represent each cross-section node. Finally, we apply the cross-section level attention for which we create a new adjacency matrix representing a complete graph. Since, at the cross-section level, there is no strict adjacency, representations for each cross-section must be learnable. We let the network perform attention on the complete graph giving it complete flexibility to attend to any cross-section. We still need to maintain the graph-level representation at this stage since we still require permutation invariance at this stage. 

In our implementation, in order to restrict the attention to piece-level and cross-section levels, we explicitly pass the piece-level adjacency matrix during initial graph convolutions; this restricts the neighborhood of the nodes to attend within cross-sections, after which we aggregate the piece-level information and later replace the adjacency matrix with a complete graph adjacency. 
\noindent
% \begin{minipage}{0.5\textwidth}

\begin{figure*}[!htp]
    \centering
    \setlength\tabcolsep{2pt} % default value: 6pt   
    \setlength{\abovecaptionskip}{1pt} 
    \setlength{\belowcaptionskip}{1pt} 
    % \begin{tabular}{cc}
    
    \scalebox{0.7}{
    \begin{tabular}{cccccccc}
    & \multicolumn{2}{c}{\textbf{Aircraft}} & \multicolumn{2}{c}{\textbf{Chair}} & \multicolumn{2}{c}{\textbf{Sofa}} 
 
 \\\cmidrule(lr){2-3} \cmidrule(lr){4-5} \cmidrule(lr){6-7} 
    & Ground Truth & Predicted & Ground Truth & Predicted & Ground Truth  & Predicted 
    \\
    \rotatebox{90}{{10 cross-sections}} &
    \includegraphics[width=0.18\linewidth,trim={10cm 3cm 10cm 0cm},clip]{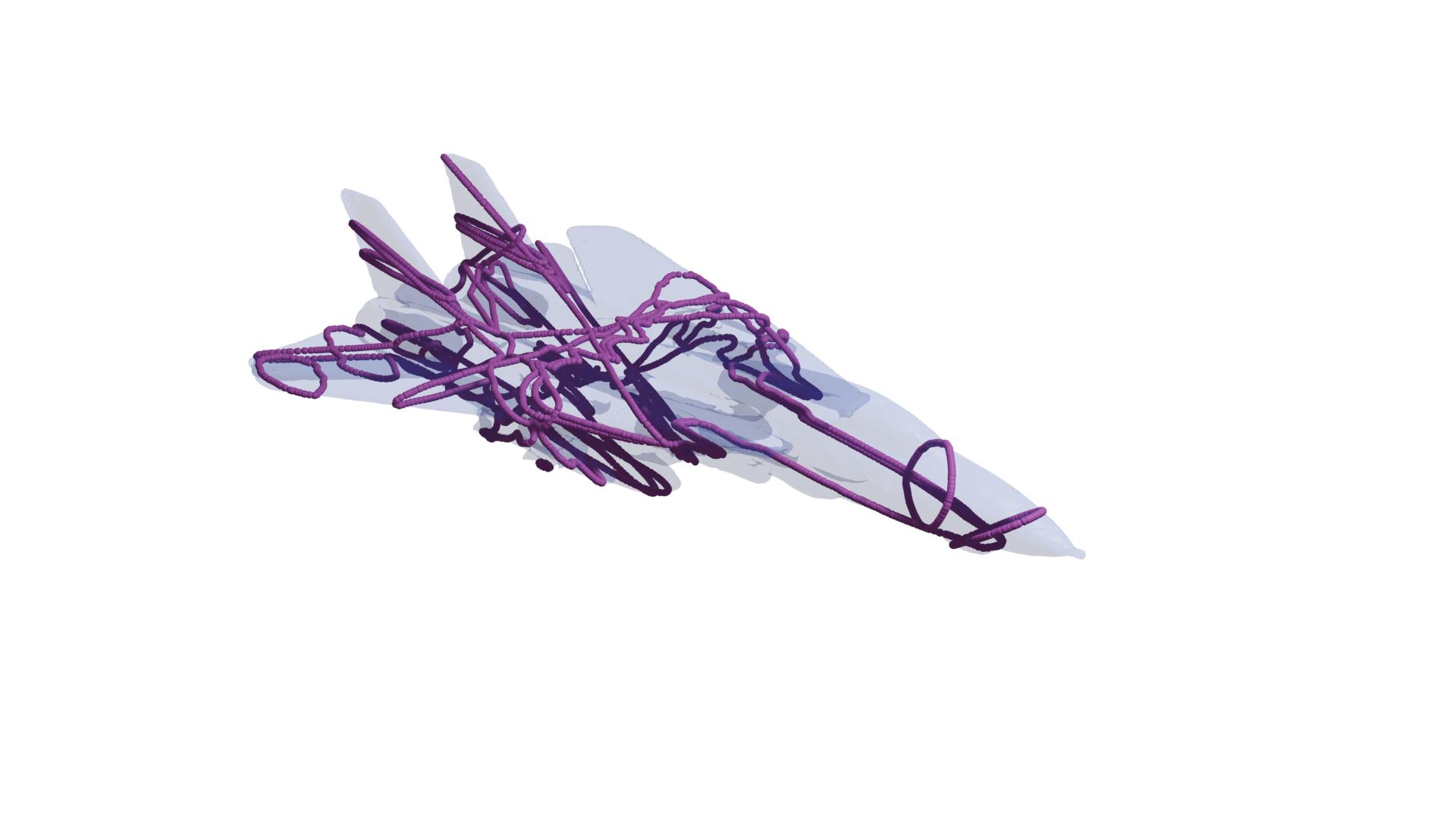}
    &
    \includegraphics[width=0.18\linewidth,trim={10cm 3cm 10cm 0cm},clip]{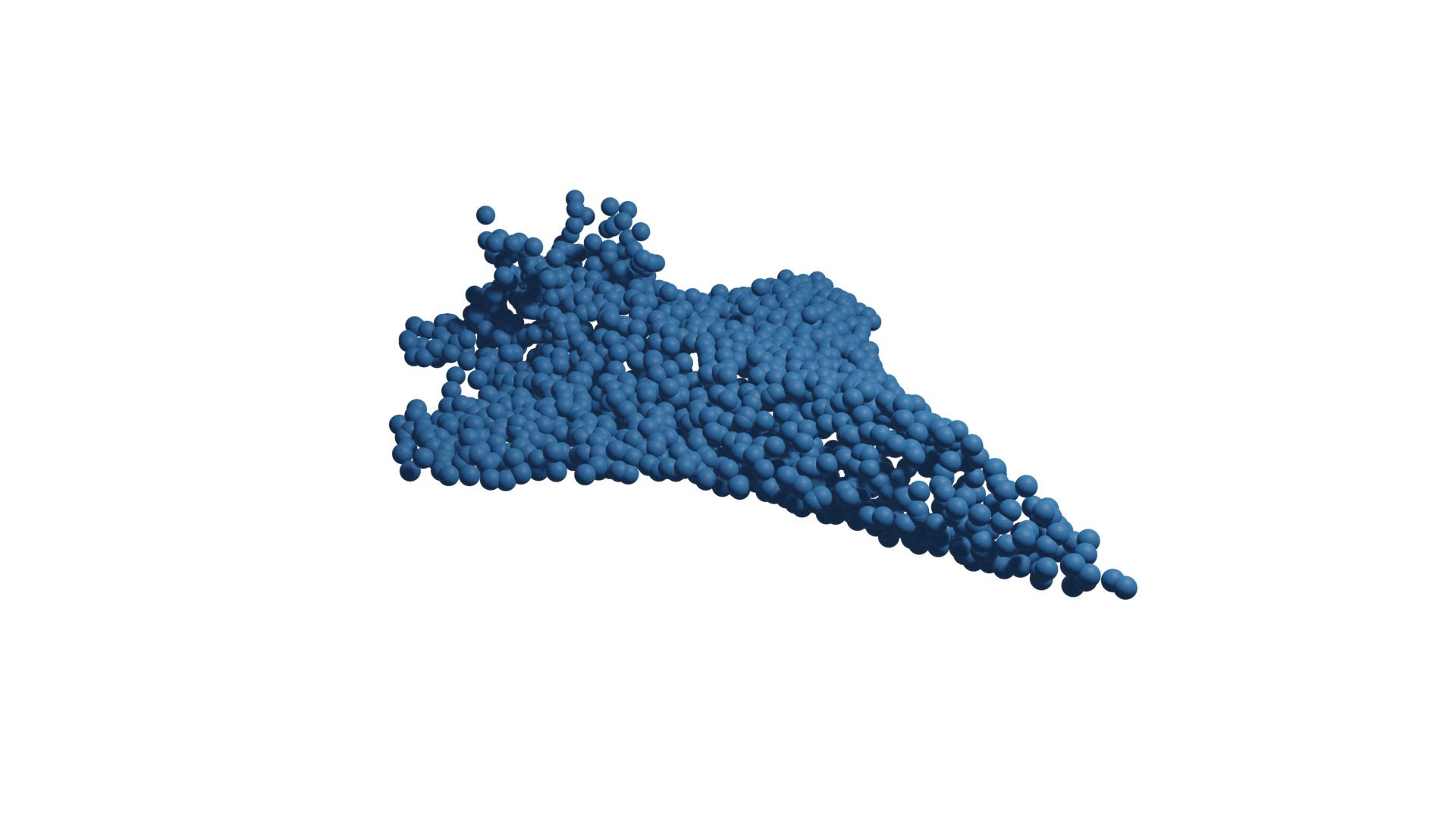}
    
    &
  
     \includegraphics[width=0.18\linewidth,trim={15cm 0cm 15cm 0cm},clip]{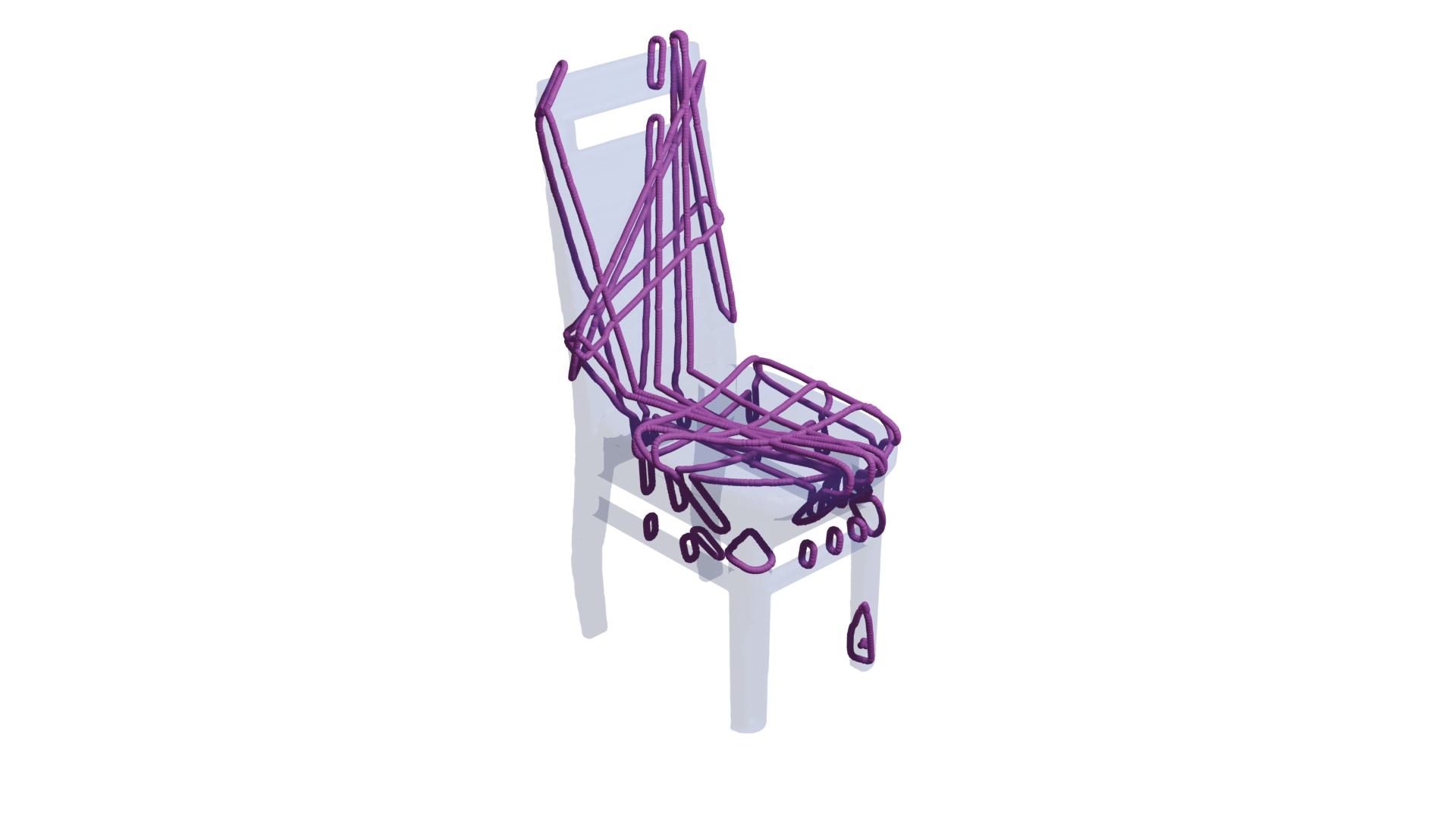} 
    &
    \includegraphics[width=0.18\linewidth,trim={15cm 0cm 15cm 0cm},clip]{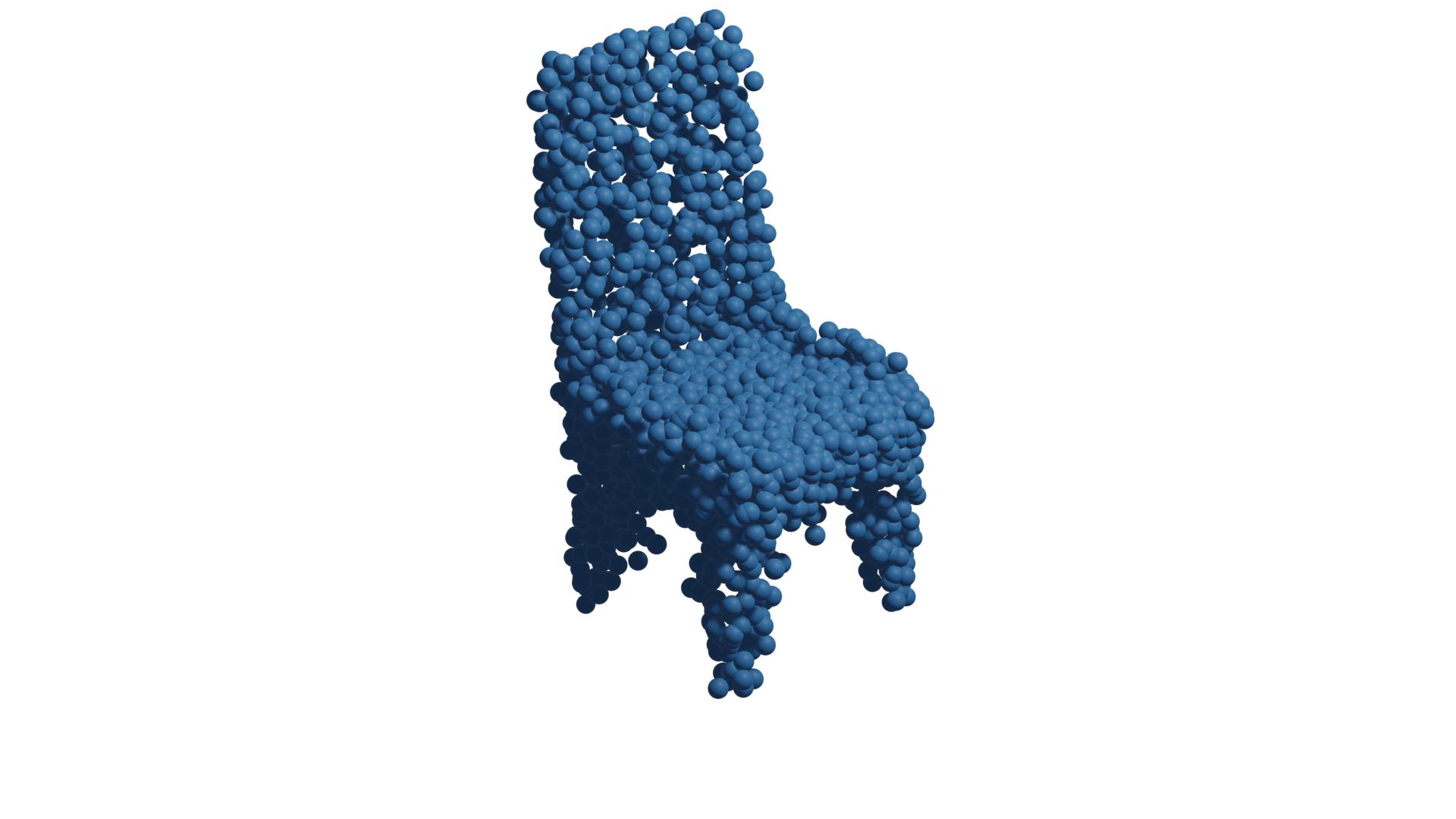} 
    
    & 
    
    \includegraphics[width=0.18\linewidth,trim={15cm 0cm 15cm 0cm},clip]{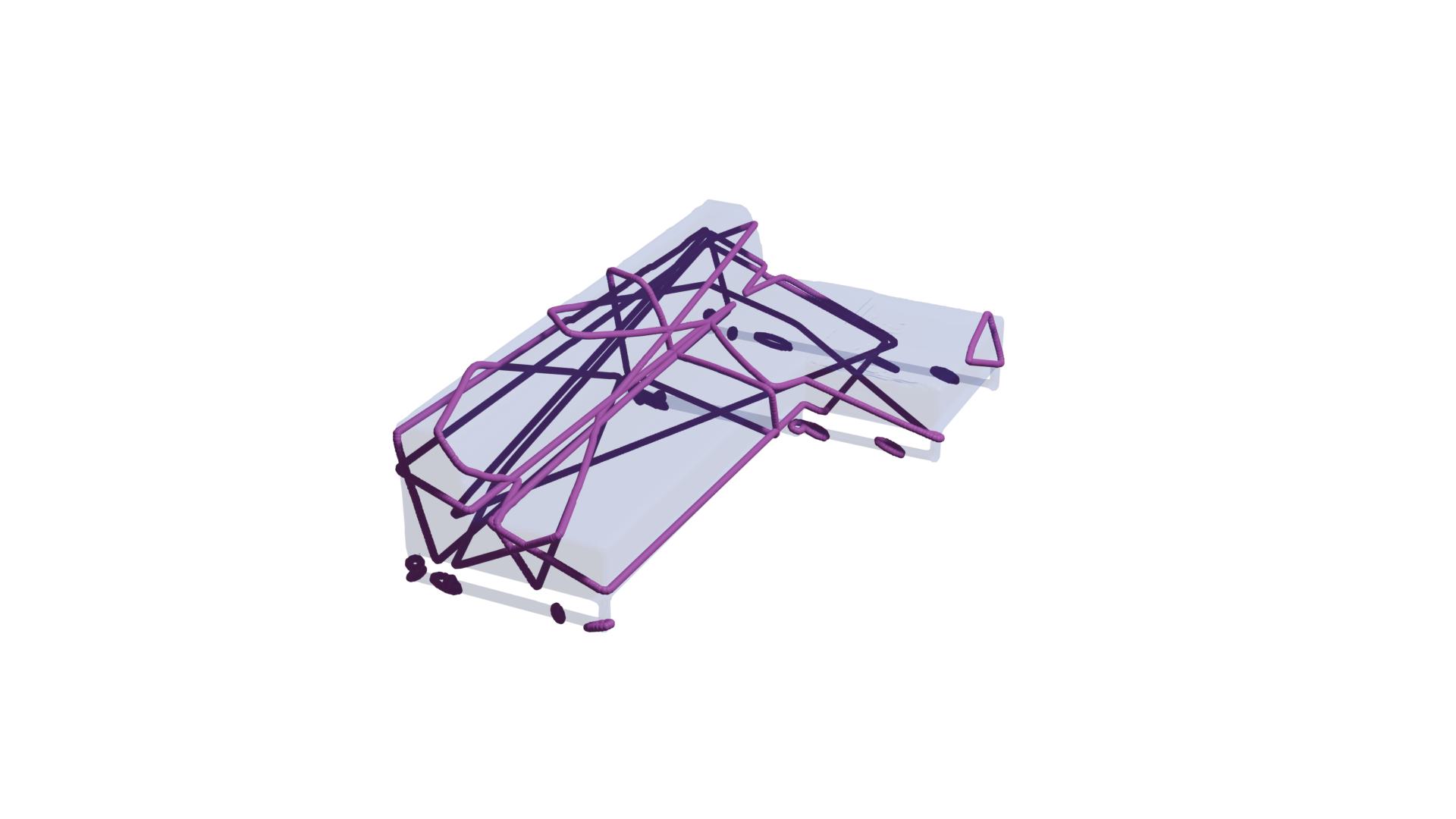} & 
    \includegraphics[width=0.18\linewidth,trim={15cm 0cm 15cm 0cm},clip]{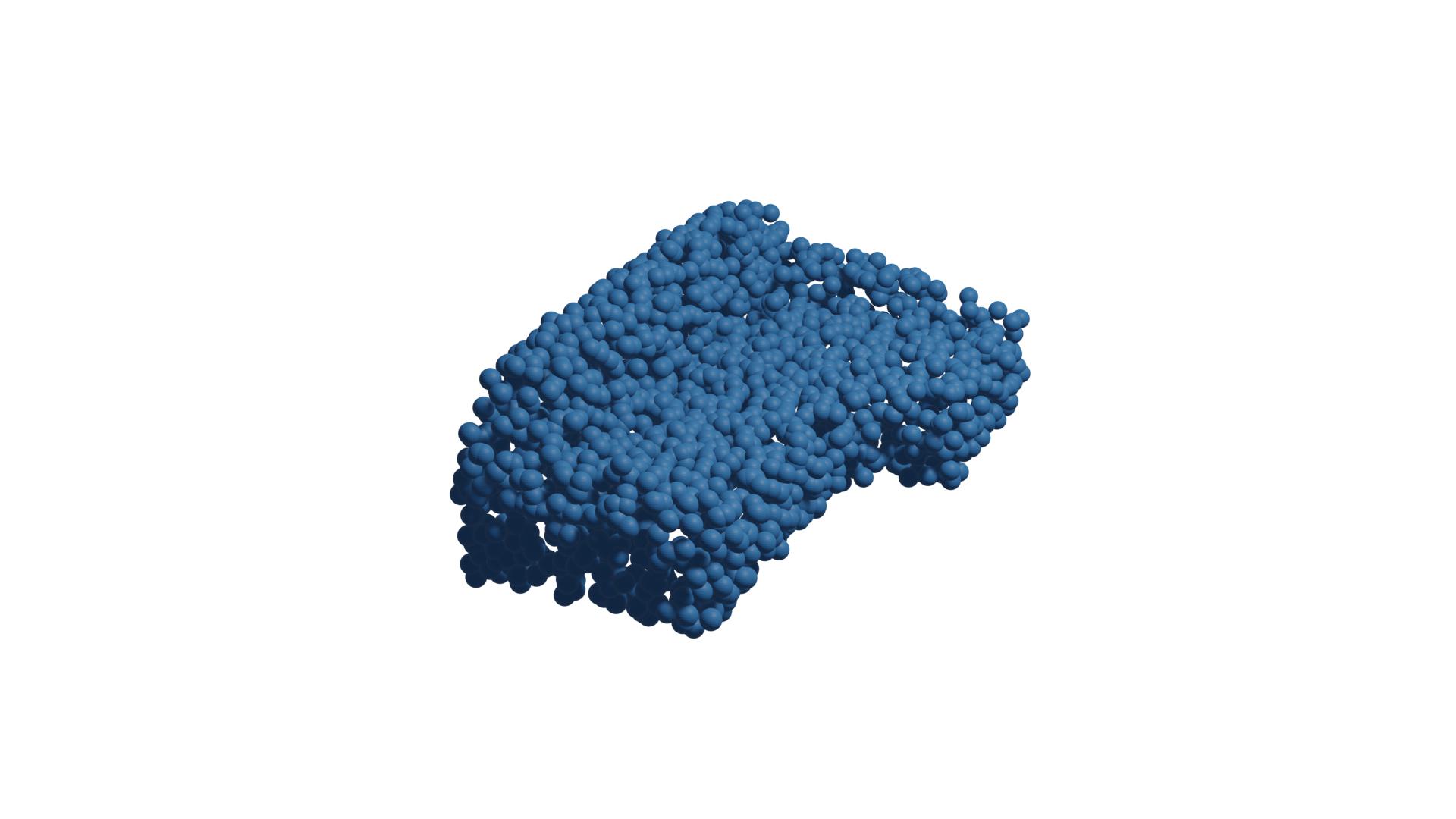}
    \\
    
        \rotatebox{90}{{5 cross-sections}} &

    \includegraphics[width=0.18\linewidth,trim={10cm 3cm 10cm 0cm},clip]{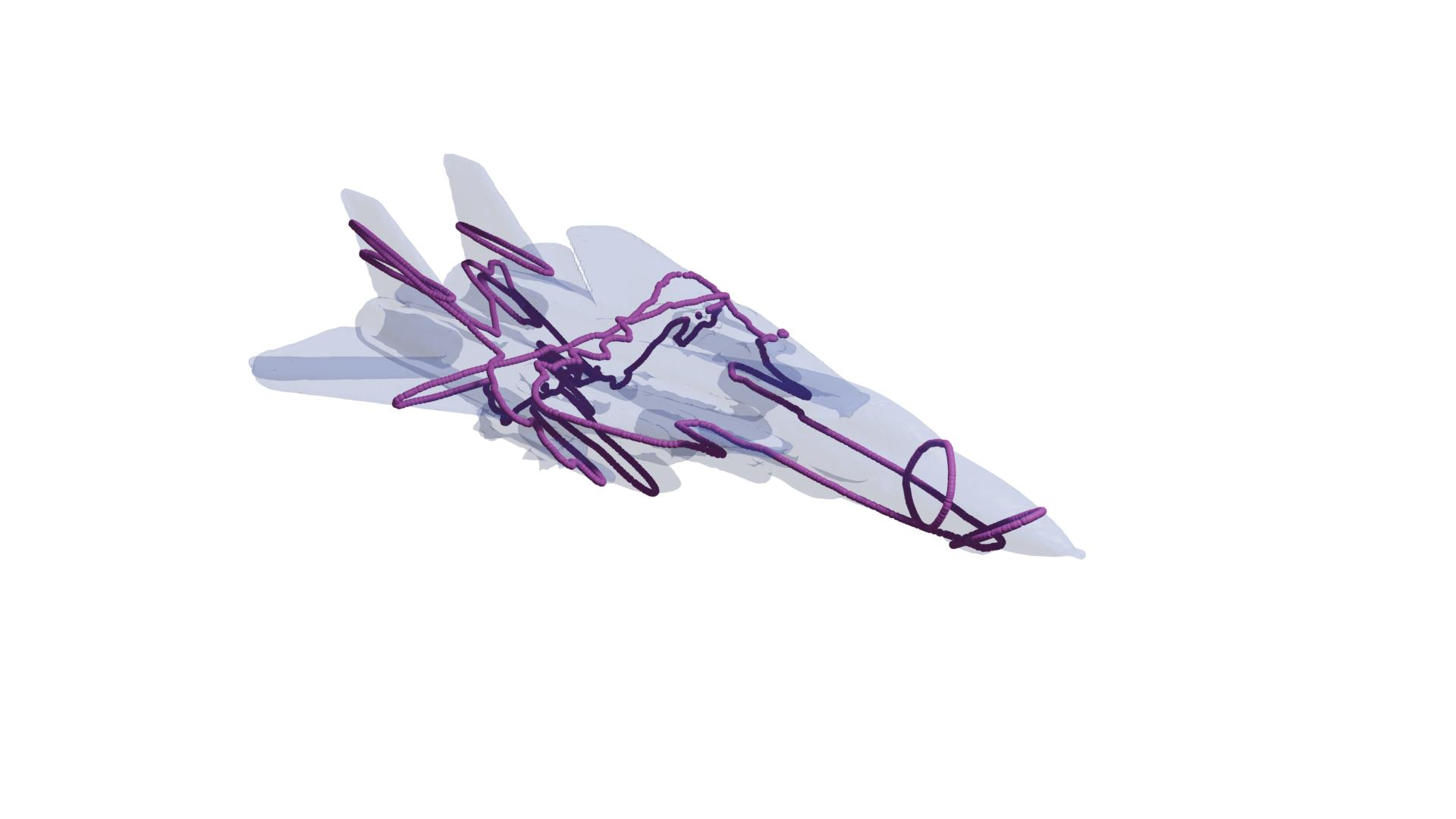}
    &
    \includegraphics[width=0.18\linewidth,trim={10cm 3cm 10cm 0cm},clip]{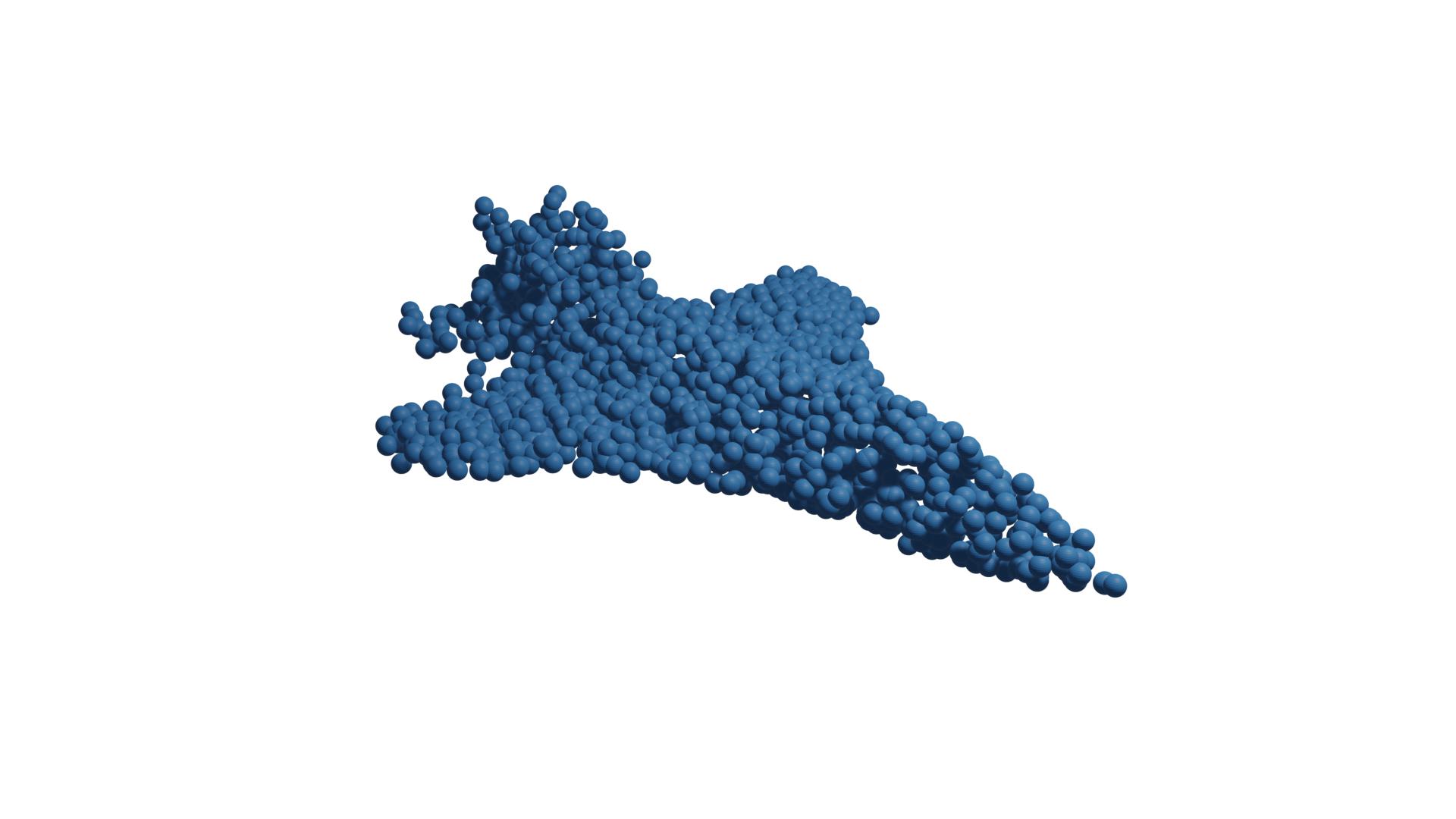}
    
    &

     \includegraphics[width=0.18\linewidth,trim={15cm 0cm 15cm 0cm},clip]{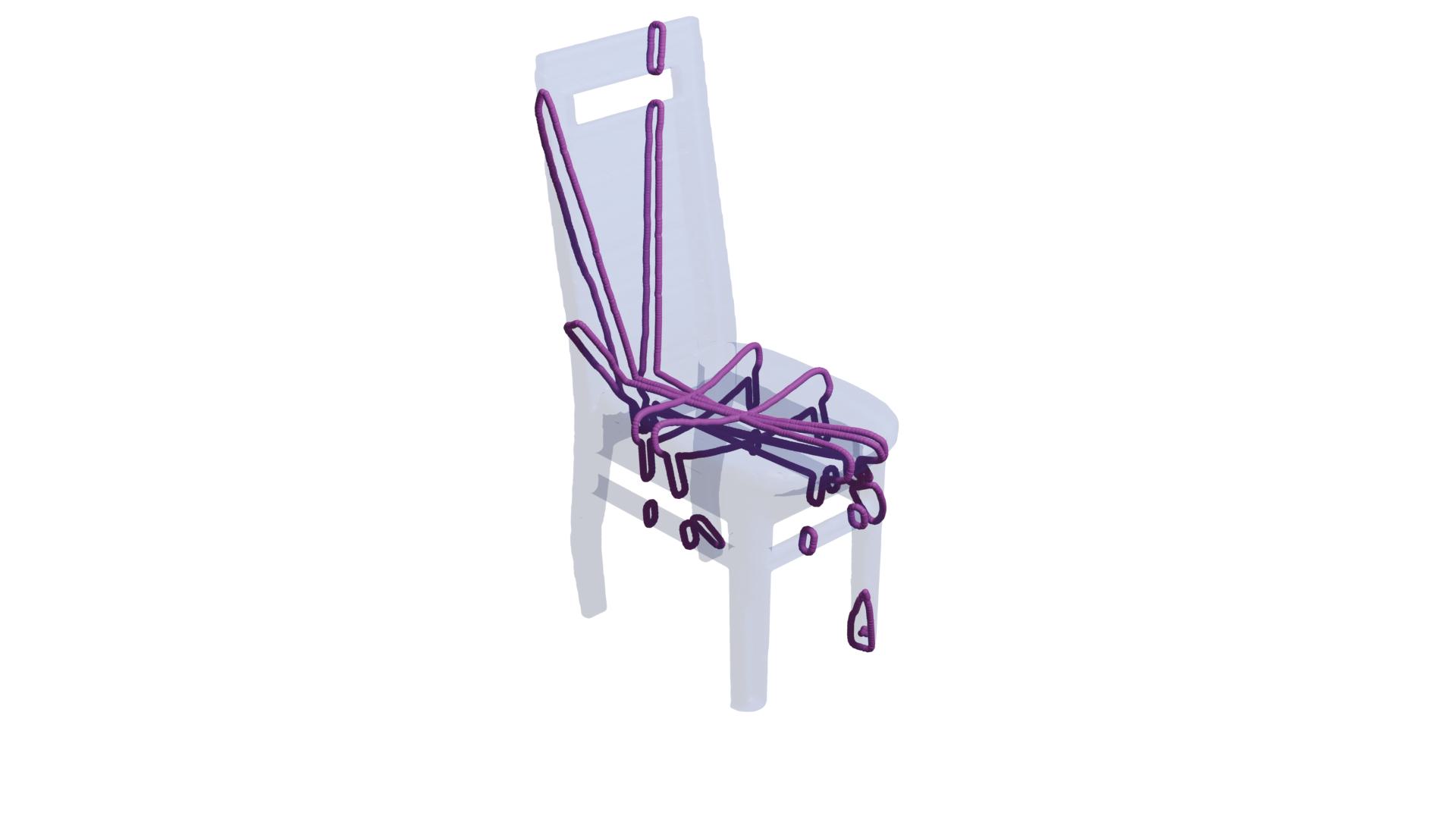} &
    
    \includegraphics[width=0.18\linewidth,trim={15cm 0cm 15cm 0cm},clip]{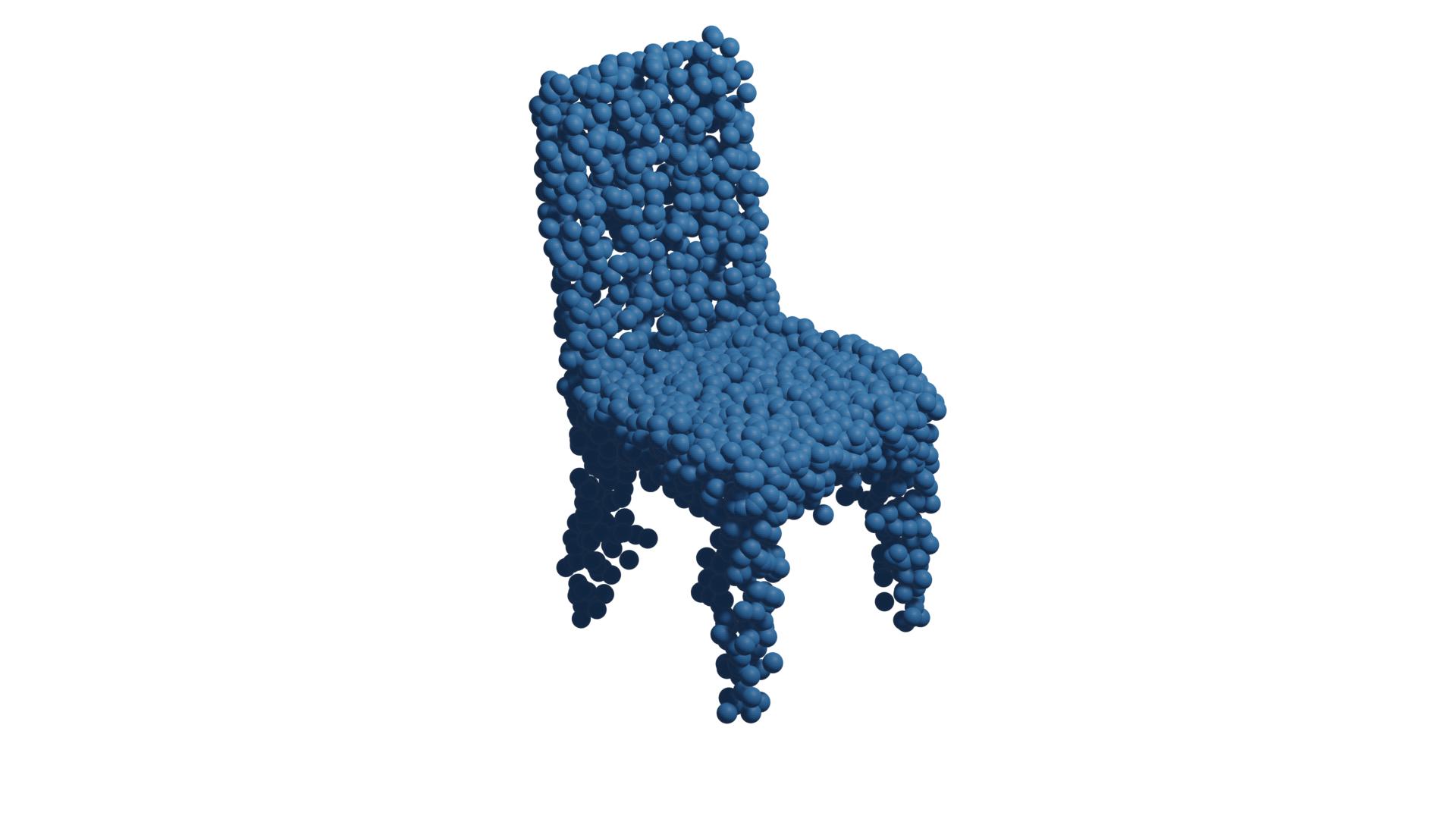}

    &

     \includegraphics[width=0.18\linewidth,trim={15cm 0cm 15cm 0cm},clip]{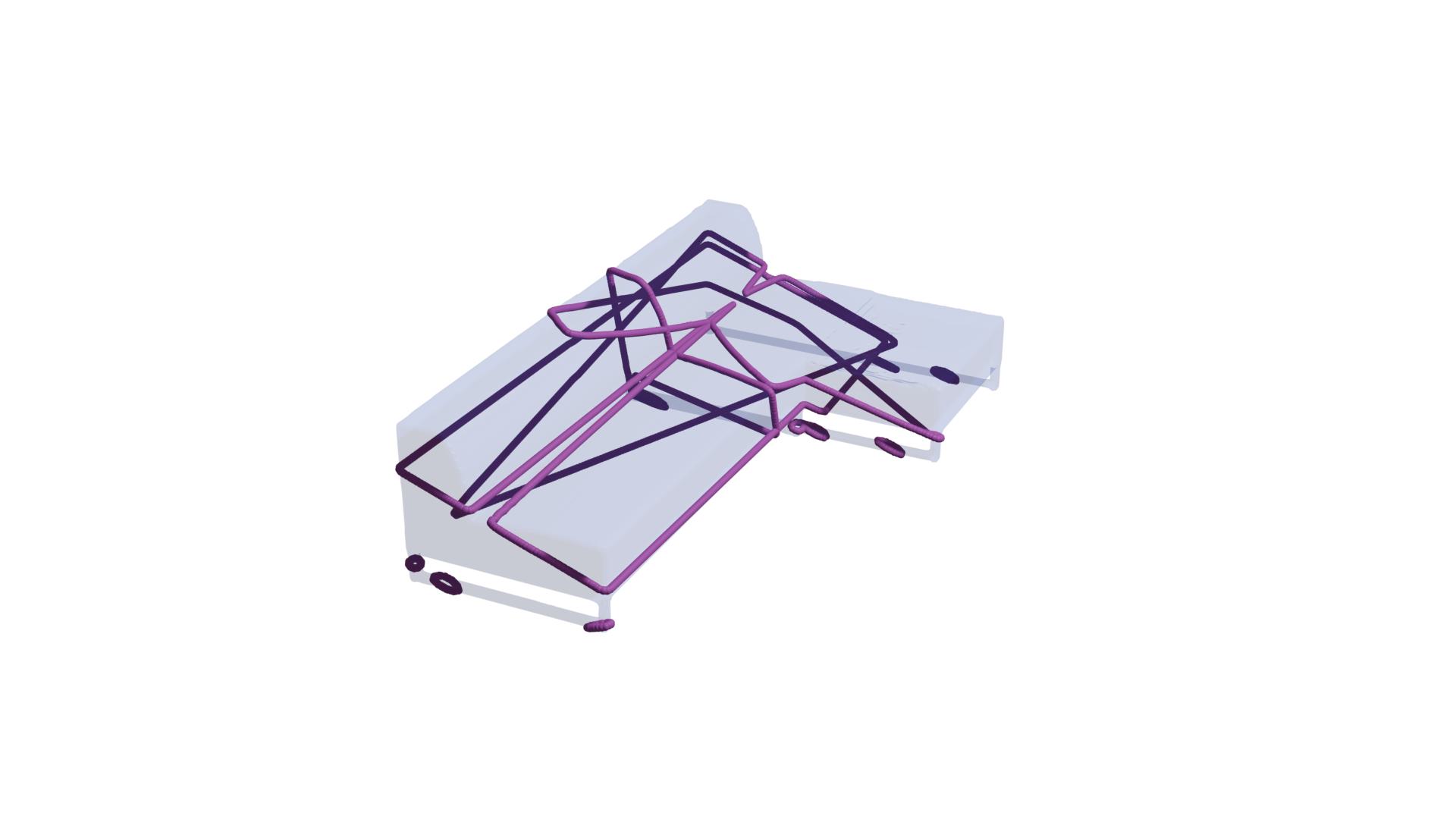} & 
    \includegraphics[width=0.18\linewidth,trim={15cm 0cm 15cm 0cm},clip]{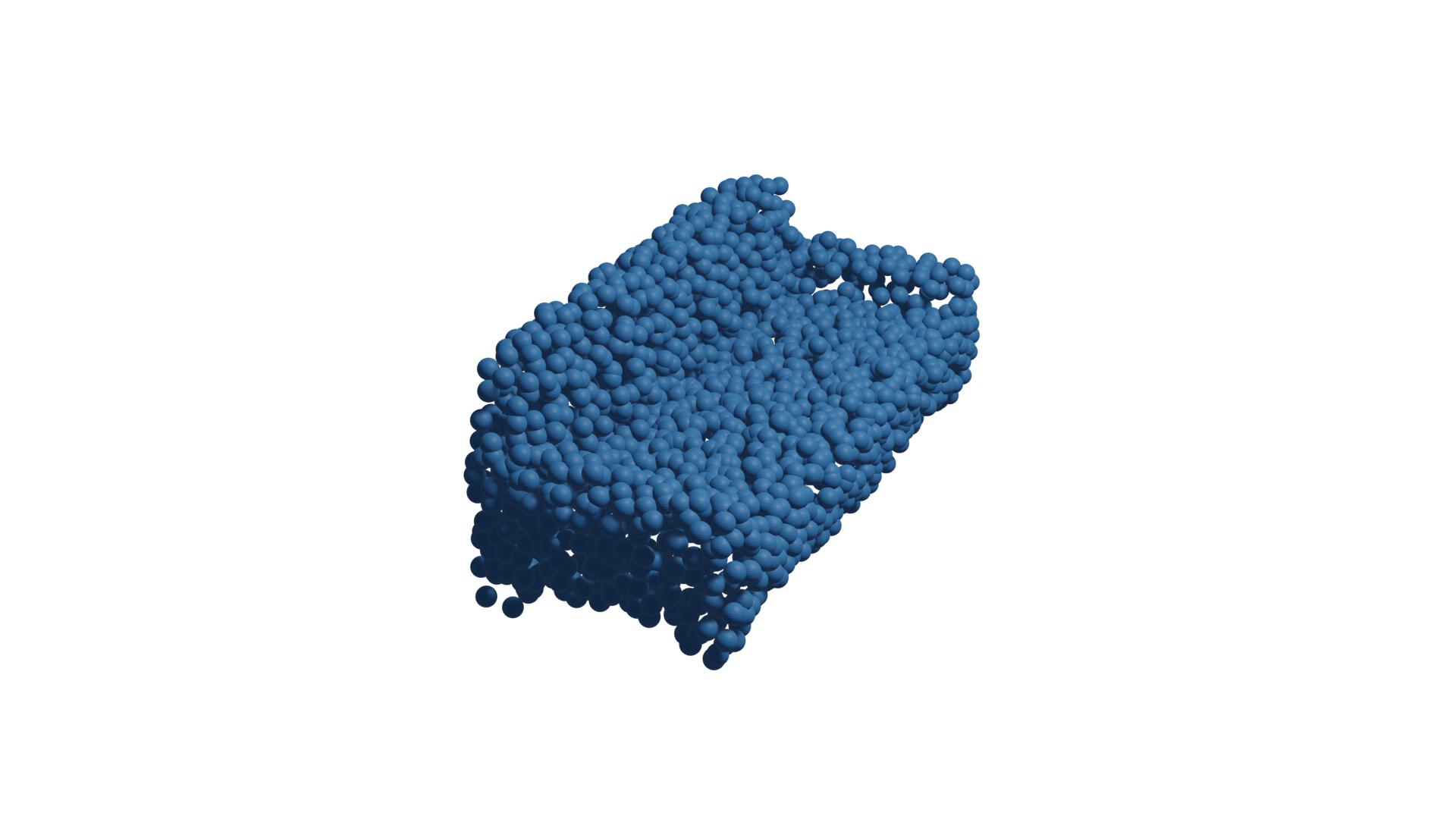}

    \\
    
        \rotatebox{90}{{2 cross-sections}} &

    \includegraphics[width=0.18\linewidth,trim={10cm 3cm 10cm 0cm},clip]{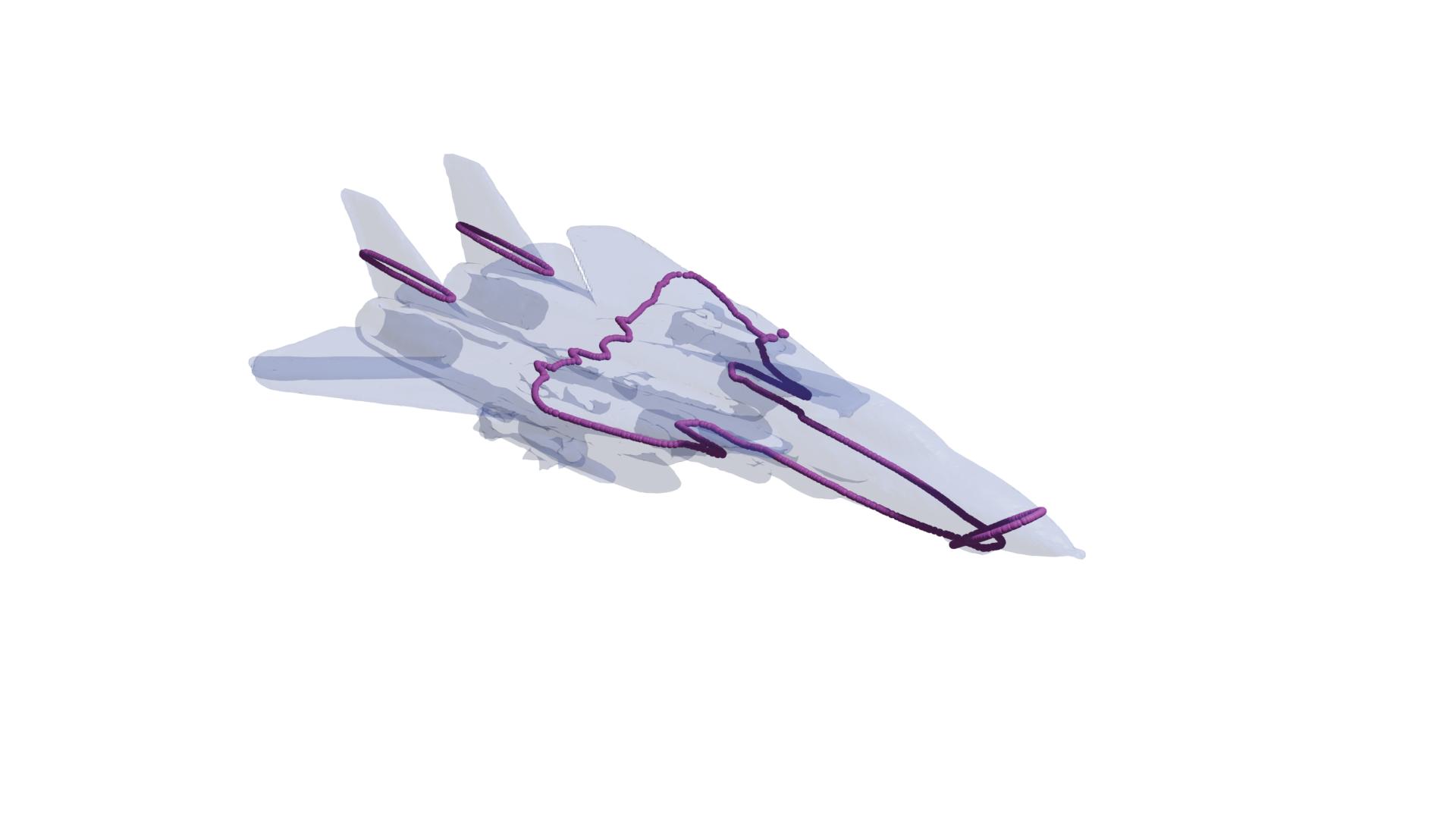}
        &
      \includegraphics[width=0.18\linewidth,trim={10cm 3cm 10cm 0cm},clip]{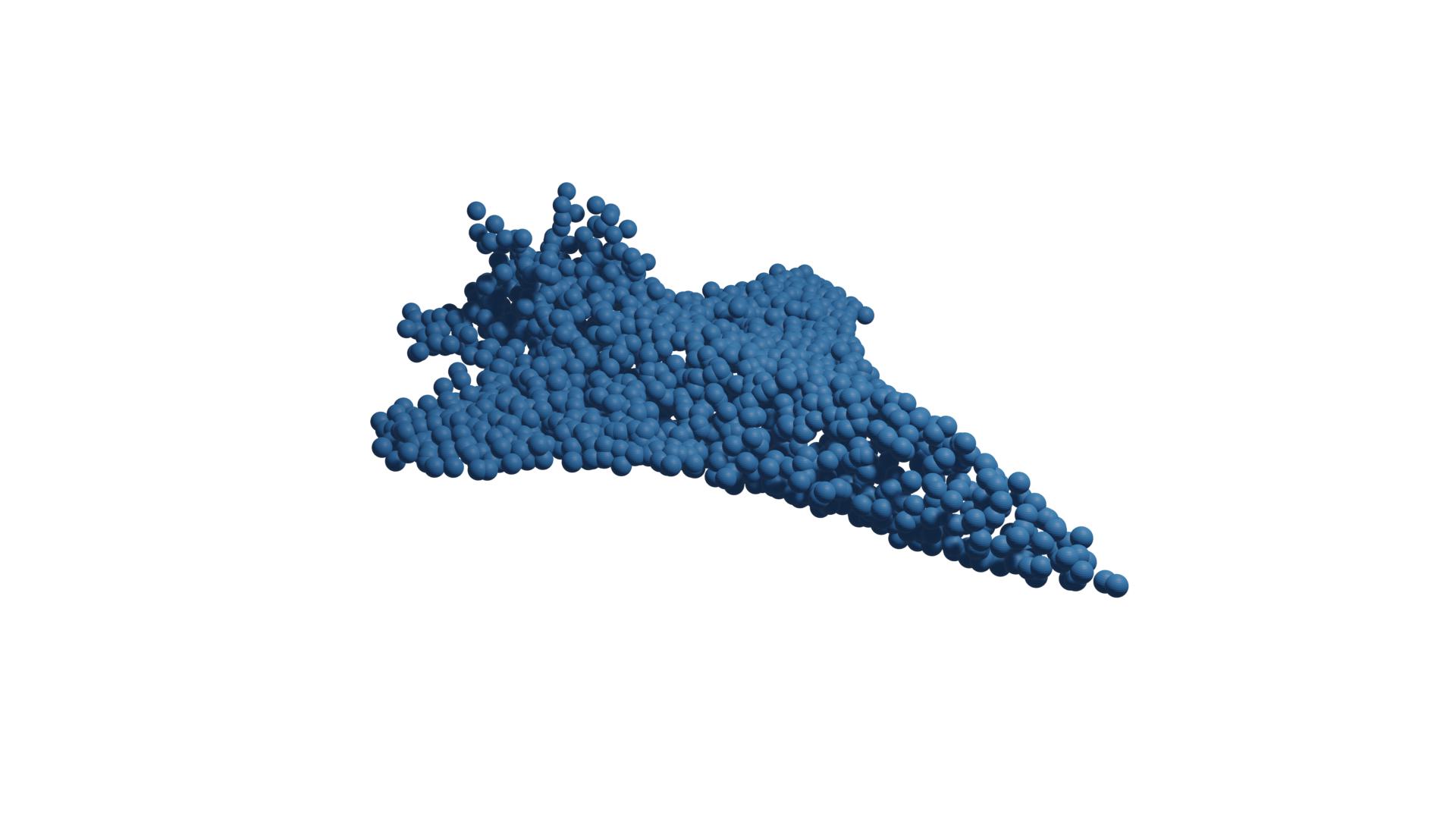}
    
    &
    
    \includegraphics[width=0.18\linewidth,trim={15cm 0cm 15cm 0cm},clip]{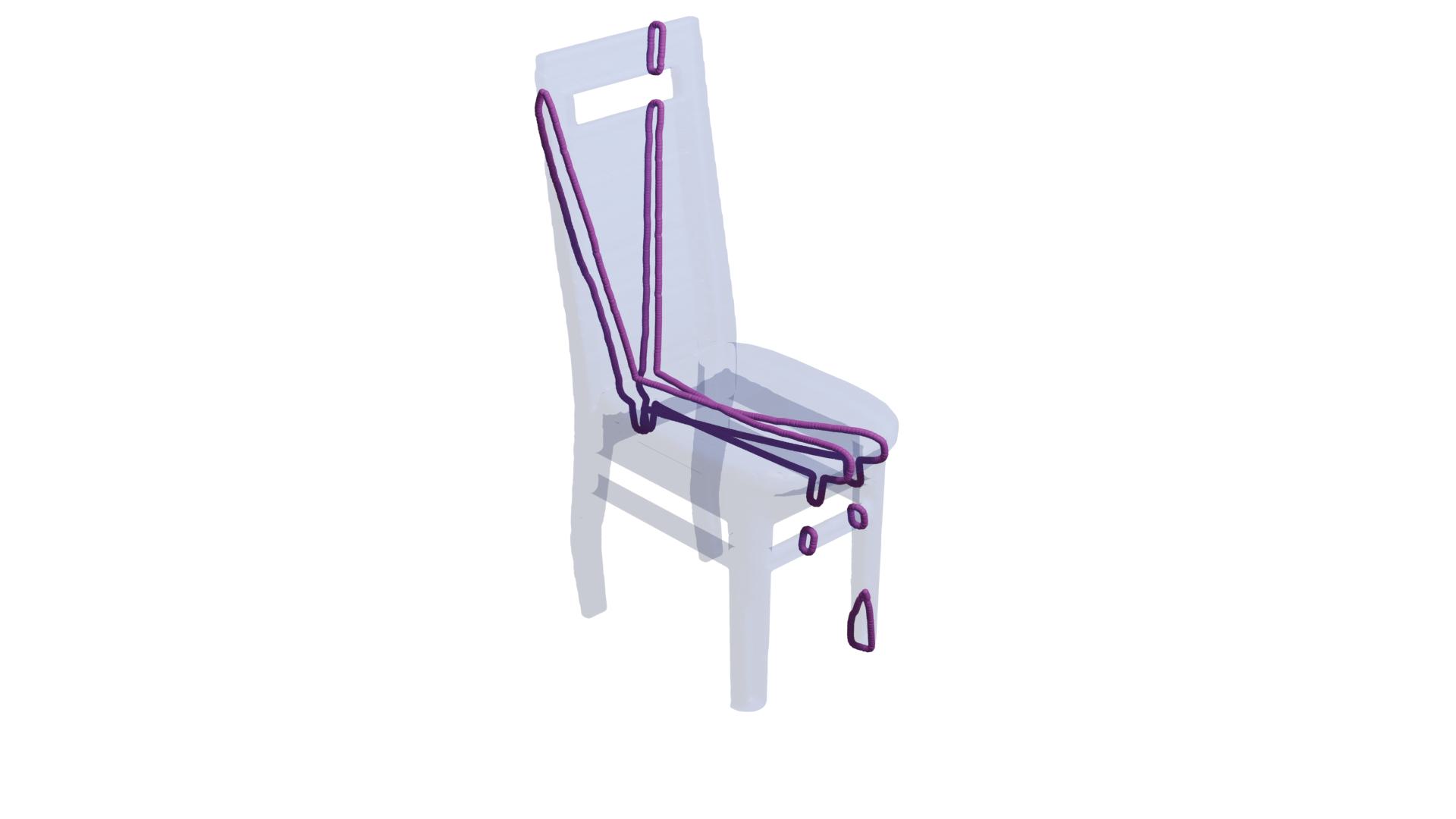} &

    \includegraphics[width=0.18\linewidth,trim={15cm 0cm 15cm 0cm},clip]{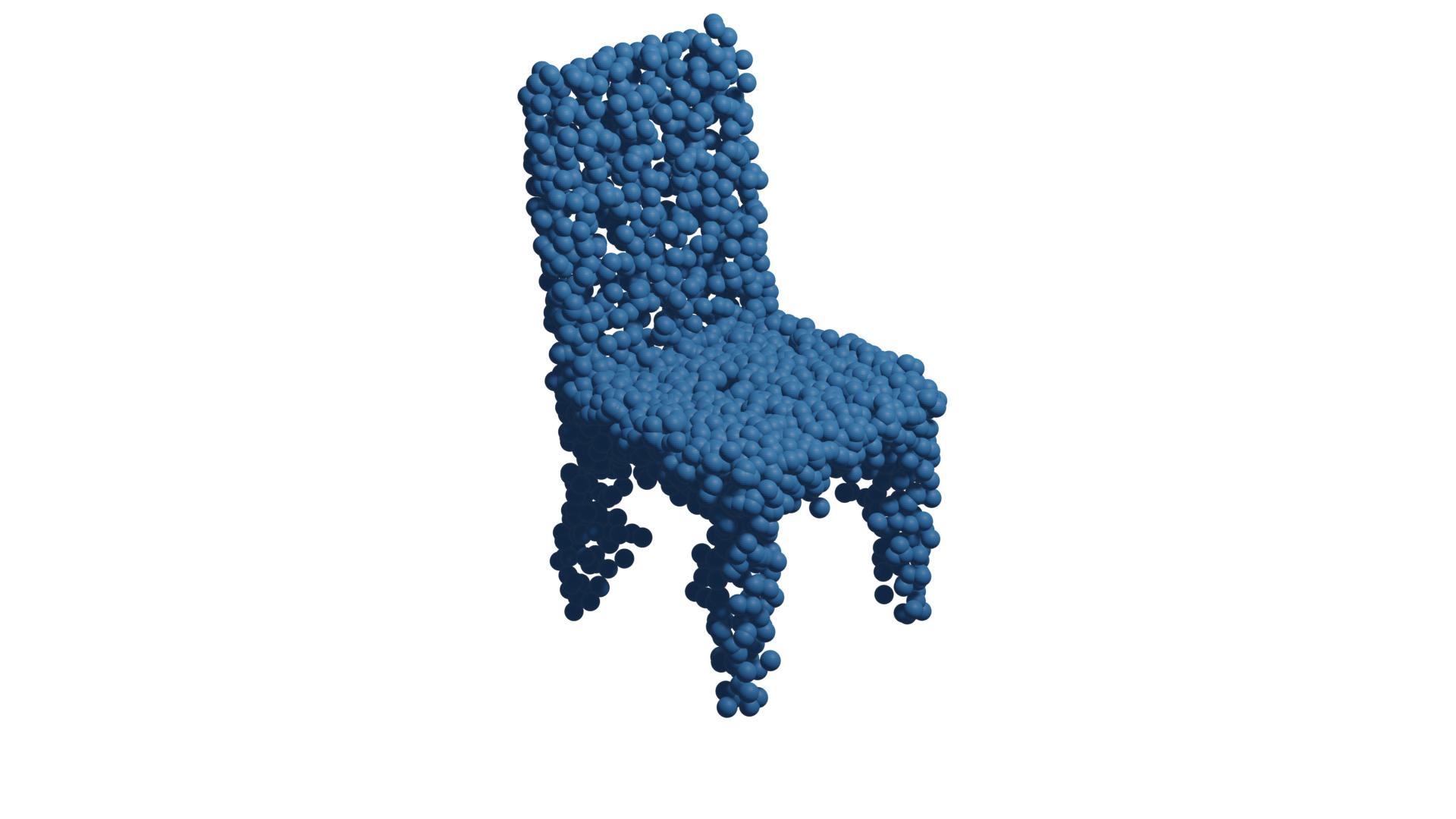}
    & 

     \includegraphics[width=0.18\linewidth,trim={15cm 0cm 15cm 0cm},clip]{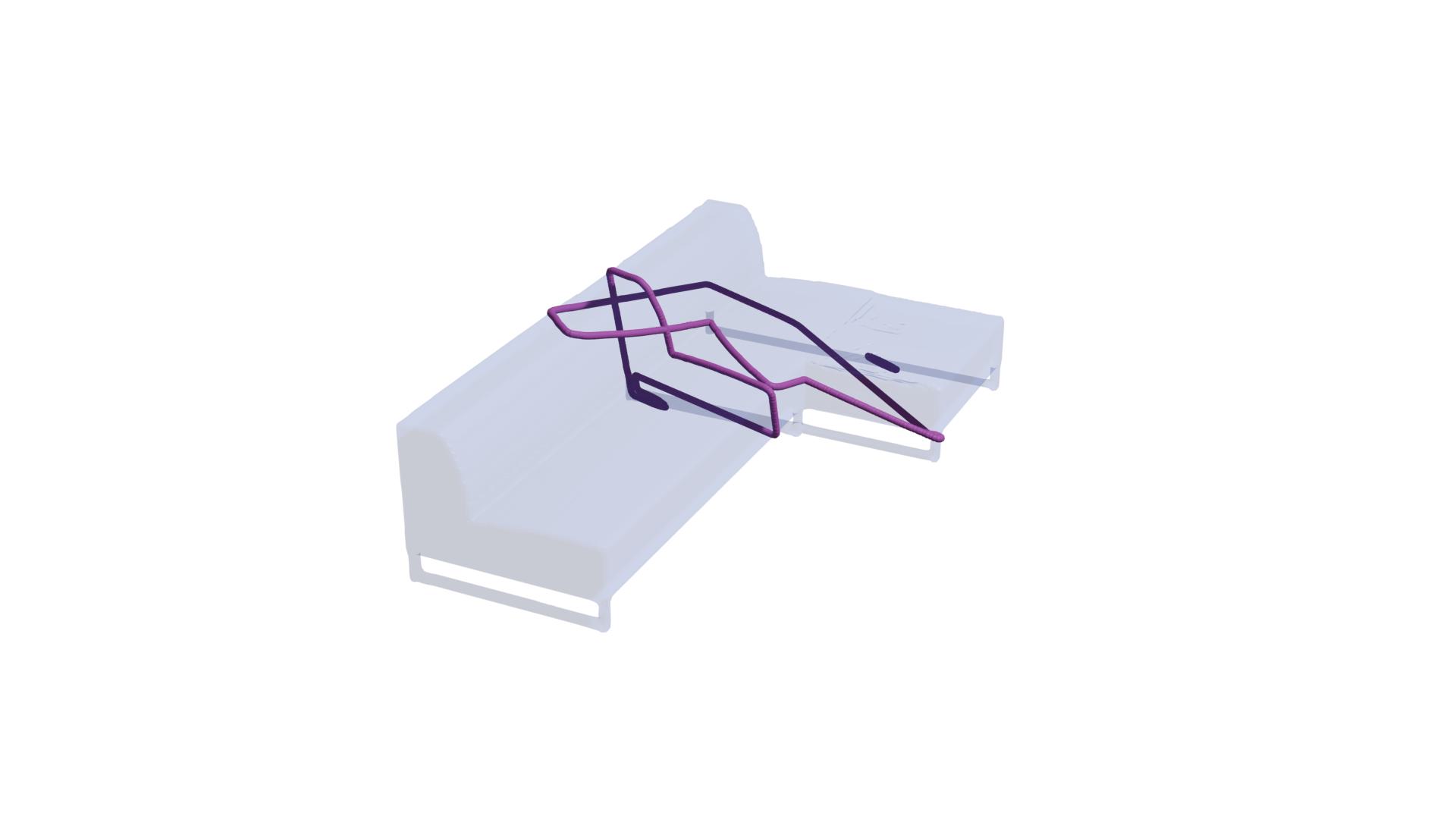} & 
    \includegraphics[width=0.18\linewidth,trim={15cm 0cm 15cm 0cm},clip]{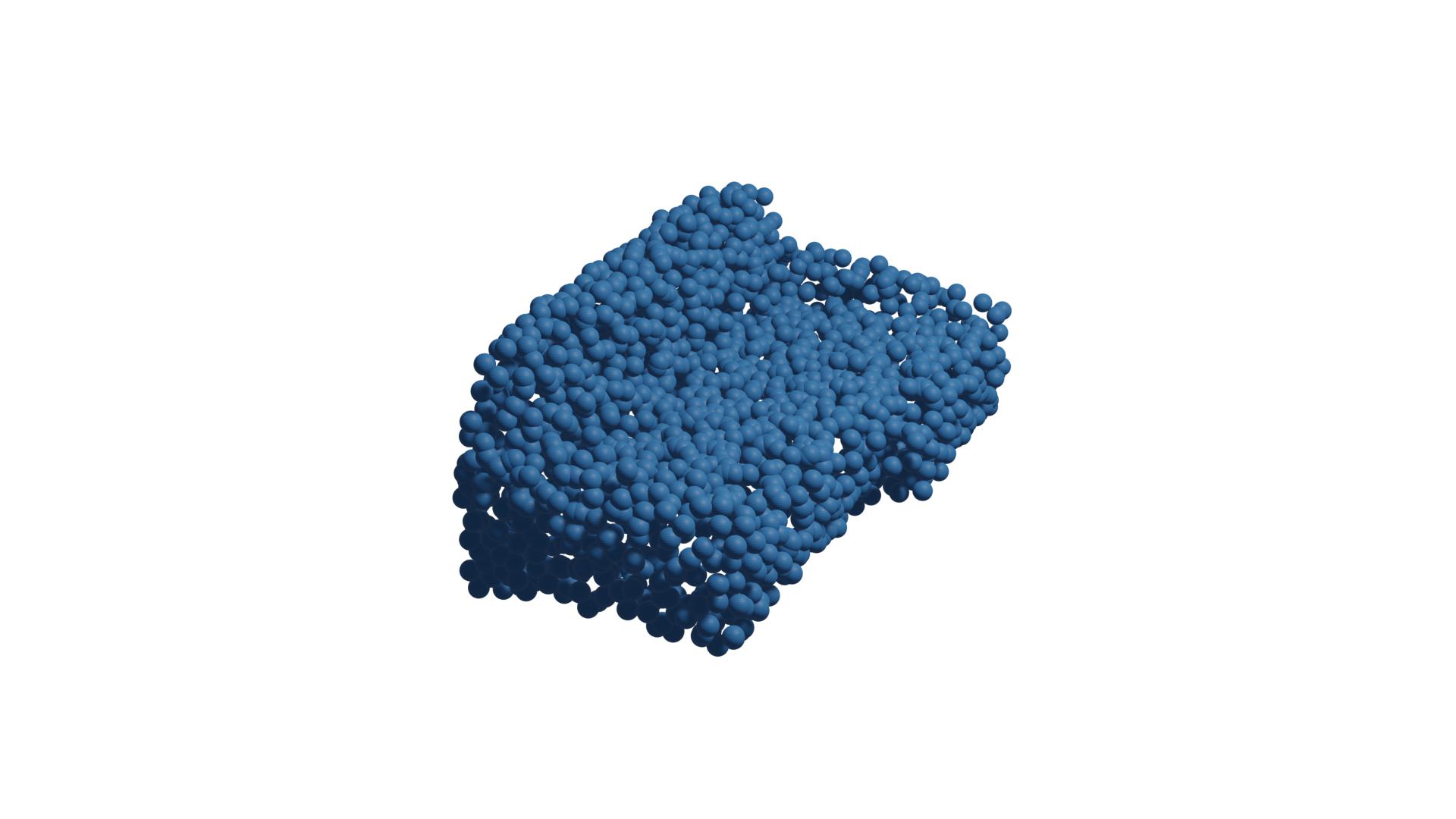}

    \end{tabular}}  
    % \end{tabular}
    \caption{ (Left) Comparison of reconstruction quality with an increasing number of cross-sections. Input to the network is the set of cross-sections (red) belonging to the ground truth mesh(blue). %(Right) Additional generated samples.
    \label{fig:comparison}
    }

\end{figure*}
\subsubsection{Adapting for Variable Cross-sections}

Since the network takes the input in the form of parametric cross-sections, where each cross-section consists of piecewise $C_1$ parametric curves, the parameters of the network become fixed during training if MLPs are used, prohibiting any changes in the number of cross-sections or pieces provided. In order to adapt to the variable nature of our data, we are further motivated to use the graph-based representation by allowing piece-level aggregation and cross-section level aggregation, which allows for a variable number of cross-sections to be provided to the network. Furthermore, we cannot use 1D-convolutions or 2-D convolutions directly in the parametric space because convolutions are not well defined on coefficient spaces. 
%Since the parameters across pieces belong to the same cross-sections and different permutations of cross-sections can belong to the same object; thus, any permutations of cross-sections and pieces should yield the same object. Graph convolutions can potentially handle a variable number of nodes and allow for permutation invariance giving the highest flexibility making them most suitable for our problem setting.

We use graph convolutions in both the generator and discriminator. The discriminator is conditioned using the input graph parameters and predicts whether the generated embedding vector is real or fake, the input graph is converted to a graph-level embedding using successive graph convolutions \cite{kipf2016semi} and aggregation. Then the embedding vector is concatenated with the generated embedding and passed to subsequent layers. While the generator consisted of SAGEConv \cite{hamilton2017inductive} followed by DiffNorm \cite{zhou2020towards} to prevent over-smoothing and allow for deeper network and Graph Attention Convolutions \cite{velivckovic2017graph} followed by aggregation and fully connected layers to generate graph embedding. In order to allow for stochasticity in the generated outputs like in a general GAN setting, we append a noise to the parameter vector of each piece.

\subsubsection{Training Details}
Since the network takes the input in the form of parametric cross-sections, where each cross-section consists of piecewise $C_1$ parametric curves, the parameters of the network become fixed during training if MLPs are used, prohibiting any changes in the number of cross-sections or pieces provided. In order to adapt to the variable nature of our data, we are further motivated to use the graph-based representation by allowing piece-level aggregation and cross-section level aggregation, which allows for a variable number of cross-sections to be provided to the network. Furthermore, we cannot use 1D-convolutions or 2-D convolutions directly in the parametric space because convolutions are not well defined on coefficient spaces. 
%Since the parameters across pieces belong to the same cross-sections and different permutations of cross-sections can belong to the same object; thus, any permutations of cross-sections and pieces should yield the same object. Graph convolutions can potentially handle a variable number of nodes and allow for permutation invariance giving the highest flexibility making them most suitable for our problem setting.

We use graph convolutions in both the generator and discriminator. The discriminator is conditioned using the input graph parameters and predicts whether the generated embedding vector is real or fake, the input graph is converted to a graph-level embedding using successive graph convolutions \cite{kipf2016semi} and aggregation. Then the embedding vector is concatenated with the generated embedding and passed to subsequent layers. While the generator consisted of SAGEConv \cite{hamilton2017inductive} followed by DiffNorm \cite{zhou2020towards} to prevent oversmoothing and allow for deeper network and Graph Attention Convolutions \cite{velivckovic2017graph} followed by aggregation and fully connected layers to generate graph embedding. In order to allow for stochasticity in the generated outputs like in a general GAN setting, we append a noise to the parameter vector of each piece.

\subsubsection{Training details}
Given a pre-trained autoencoder with encoder $En$ and decoder $De$ and a GCN-based generator-discriminator pair $\{G, D\}$ we pass a ground truth point cloud $P_{gt}$ containing 2048 points through the encoder to generate an embedding, $En(P_{gt})$. For a set of input parameterized cross-sections $C$, we create the piece-wise adjacency matrix $\mathbf{A}_{p}$ for each cross-section and a cross-section adjacency matrix $\mathbf{A}_{c}$.

The generator is trained to generate an embedding using the cross-section set $C$ and the two adjacency matrices for the point cloud. The generator loss is given by
\begin{align*}
    \mathcal{L}_G = &\log\left(1 - D\left(G\left(C, \mathbf{A}_{p}, \mathbf{A}_{c}\right),C, \mathbf{A}_{p}, \mathbf{A}_{c}\right)\right) + \\
    &\mathcal{L}_{ch}\left(De\left(G\left(C, \mathbf{A}_{p}, \mathbf{A}_{c}\right)\right), De\left(En\left(P_{gt}\right)\right)\right) + \\
    &\mathcal{L}_{mse}\left(G\left(C, \mathbf{A}_{p}, \mathbf{A}_{c}\right), En\left(P_{gt}\right)\right),
\end{align*}
where $\mathcal{L}_{ch}$ is the Chamfer loss between the point clouds generated using the embedding estimated by the generator and the embedding of the ground truth point cloud. $\mathcal{L}_{mse}$ is the mean squared error between the embedding estimated by the generator and the embedding of the ground truth point cloud. The discriminator loss can be formulated as
\begin{align*}
\mathcal{L}_D = &\left(1-\log\left(D\left(G\left(C, \mathbf{A}_{p}, \mathbf{A}_{c}\right), C, \mathbf{A}_{p}, \mathbf{A}_{c}\right)\right)\right) +\\ 
&\log\left(D\left(En\left(P_{gt}\right), C, \mathbf{A}_{p}, \mathbf{A}_{c}\right)\right),
\end{align*}
where the discriminator is conditioned on the input cross-section graph. The generator and discriminator are trained in an adversarial manner (see \cite{goodfellow2020generative}).

\section{Results and Discussion
}

We evaluate our approach on different classes of the ShapeNet dataset. We perform an experimental procedure similar to DeepSDF where we divide the models into \textit{known} shapes, i.e. shapes that were in the training set and testing set referred to as \textit{unknown} shapes. We test our method in both single-class and multi-class settings. We show some samples for single-class training as well in the supplementary however our key focus is on multi-class training and its analysis.
We perform the training in a multi-class setting. For the multiclass setting, we test on 4 classes - \textit{airplane} (4K models), \textit{chair} (6K models), \textit{lamp} (2K models),  and \textit{sofa} (3K models). Our implementation source code will be made available on Github. We do not perform any class balancing techniques and directly train on the ShapeNet dataset.
We use pytorch geometric \cite{Fey/Lenssen/2019} for this. We demonstrate the impact of these attentions via an ablative study in the supplementary.
%inference, analysis , discussion 
% \vspace{-10pt}
\subsection{Cross-section dependence}
We compare the mean Chamfer loss obtained across the different classes for different numbers of input cross-sections (5, 10, 11, 15, 20, and 25) provided as input in Table~\ref{tab:cross_analysis}. We observe results for the Chamfer distance obtained after training are shown in Table \ref{tab:cross_analysis}. We observe that the number of cross-sections provided as input has a vital control on the output of the generated point cloud surface, as can be seen from Table~\ref{tab:cross_analysis}.  We show the results of the proposed model trained on four classes: Airplane, Chair, Lamp, and Sofa with a different number of input parameterized cross-sections in Figure ~\ref{fig:comparison}.
The first column displays the ground truth mesh used to sample the ground truth point cloud with cross-sections in red, the second column shows the reconstruction with our method.
%, the third column displays the point cloud reconstruction of \textit{complete} point cloud used for training our network and the last column displays the input cross-sections. 
We also analyze the variation of cross-sections using Chamfer distance between the generated and ground truth point surfaces as the number of input cross-sections increases/decreases in the supplementary material. We observe a dip in the loss as the number of cross-sections is increased with the eventual flattening of the loss curve (for further discussion and figures, refer to supplementary material). 

\begin{table}
\centering
\caption{Per-class Chamfer Distance corresponding to the variation in the number of cross-sections (results for both undersampled and oversampled($>10$) cross-sections are shown for a model trained on all aforementioned classes.
\label{tab:cross_analysis}
}
\scalebox{0.8}{
\begin{tabular}{cccccc}
    \toprule
    \textbf{\# cross-} & \multicolumn{4}{c}{\textbf{Per-class Chamfer distance}} & \textbf{Mean} \\
    \cline{2-5}
    \textbf{sections} & \textbf{Airplane} & \textbf{Chair} & \textbf{Lamp} & \textbf{Sofa} & \\ \midrule
    2 & 0.4050 & 0.1765 & 2.7306 & 0.3770 & 0.9223  \\ \midrule
    5 & 0.0493  & 0.0872 & 0.2394 & 0.0772 & 0.1133 \\ \midrule
    10 & 0.0395  & 0.0829  & 0.0958 & 0.0728 & 0.0728 \\ \midrule
    11 & 0.0385  & 0.0824 & 0.0927 & 0.0724 & 0.0715 \\ \midrule
    15 & 0.0378  & 0.0813 & 0.0909 & 0.0715 & 0.0704 \\ \midrule
    20 & 0.0374  & 0.0807 & 0.0898 & 0.0709 & 0.0697 \\ \midrule
    25 & 0.0370 &  0.0803 & 0.0896 & 0.0704 & 0.0693 \\ \bottomrule
\end{tabular}}
\end{table}
We discuss these trends and perform the t-SNE of the embeddings and demonstrate how the distinguishing capabilities of the network improve further with increasing the number of cross-sections in the supplementary.
However, as in Figure ~\ref{fig:comparison}, despite the sharp reduction in the number of cross-sections, the network still generates a reliable general shape for the class and can distinguish between the classes of parametric forms. In some cases, the failure of reconstruction is much higher depending on the number of samples of a particular shape of the object the network sees and the information in the cross-sections supplied. For example, in Figure ~\ref{fig:fail_cases}, in an airplane object, the cross-sections do not contain sufficient information, leading to a completely different object being created, though it is noteworthy that the class of the object reconstructed does seem correct visually. 

%TODO: write observations based on table when it is added.

% \vspace{-10pt}
\subsection{Failure Cases}
\begin{figure}
\centering
        \scalebox{0.7}{
        \begin{tabular}{cc}
        \textbf{Ground Truth} & \textbf{Predicted} \\\midrule

        \includegraphics[width=0.4\linewidth,trim={10cm 3cm 10cm 0cm},clip]{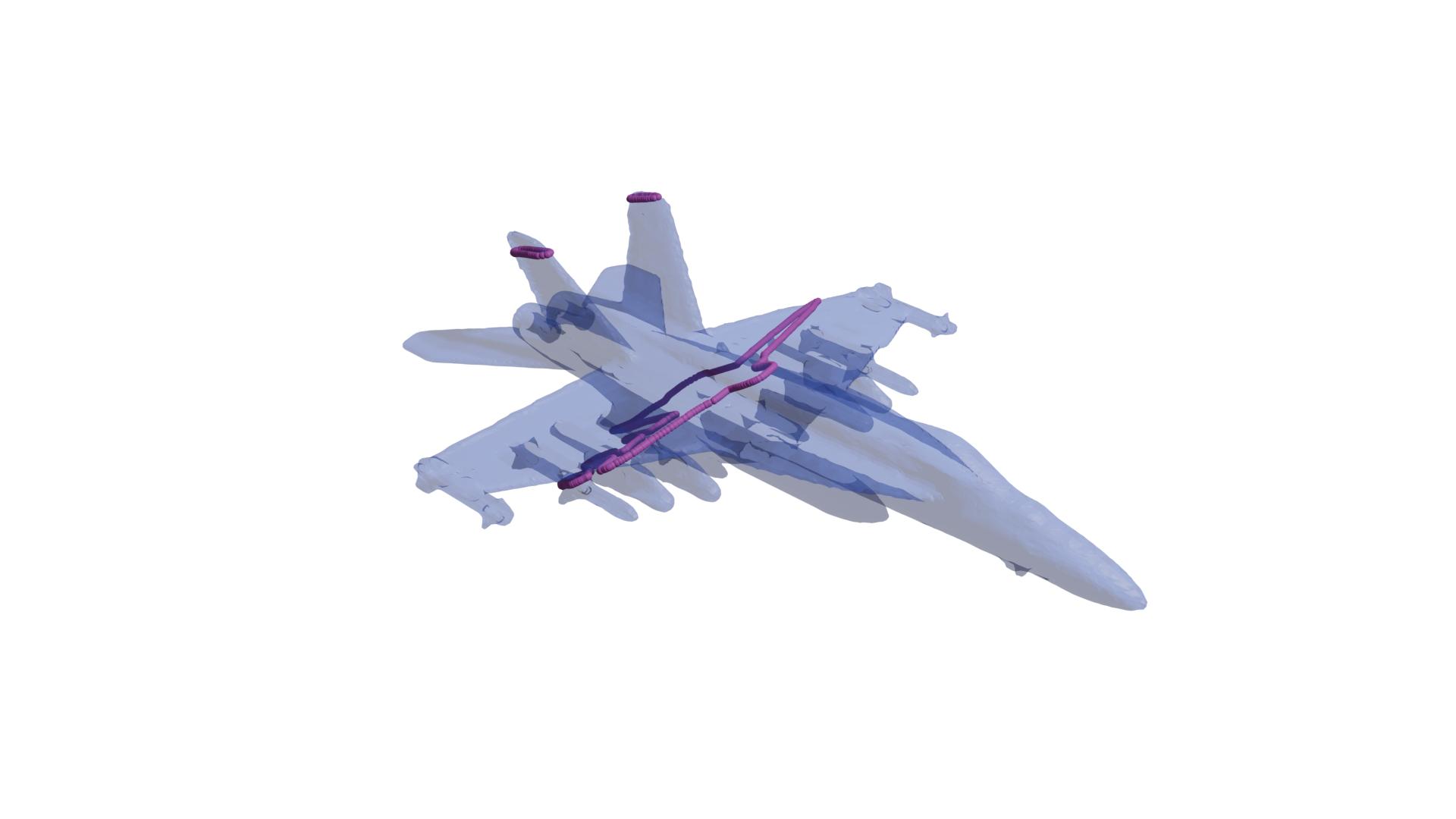}
        &

        \includegraphics[width=0.4\linewidth,trim={10cm 3cm 10cm 0cm},clip]{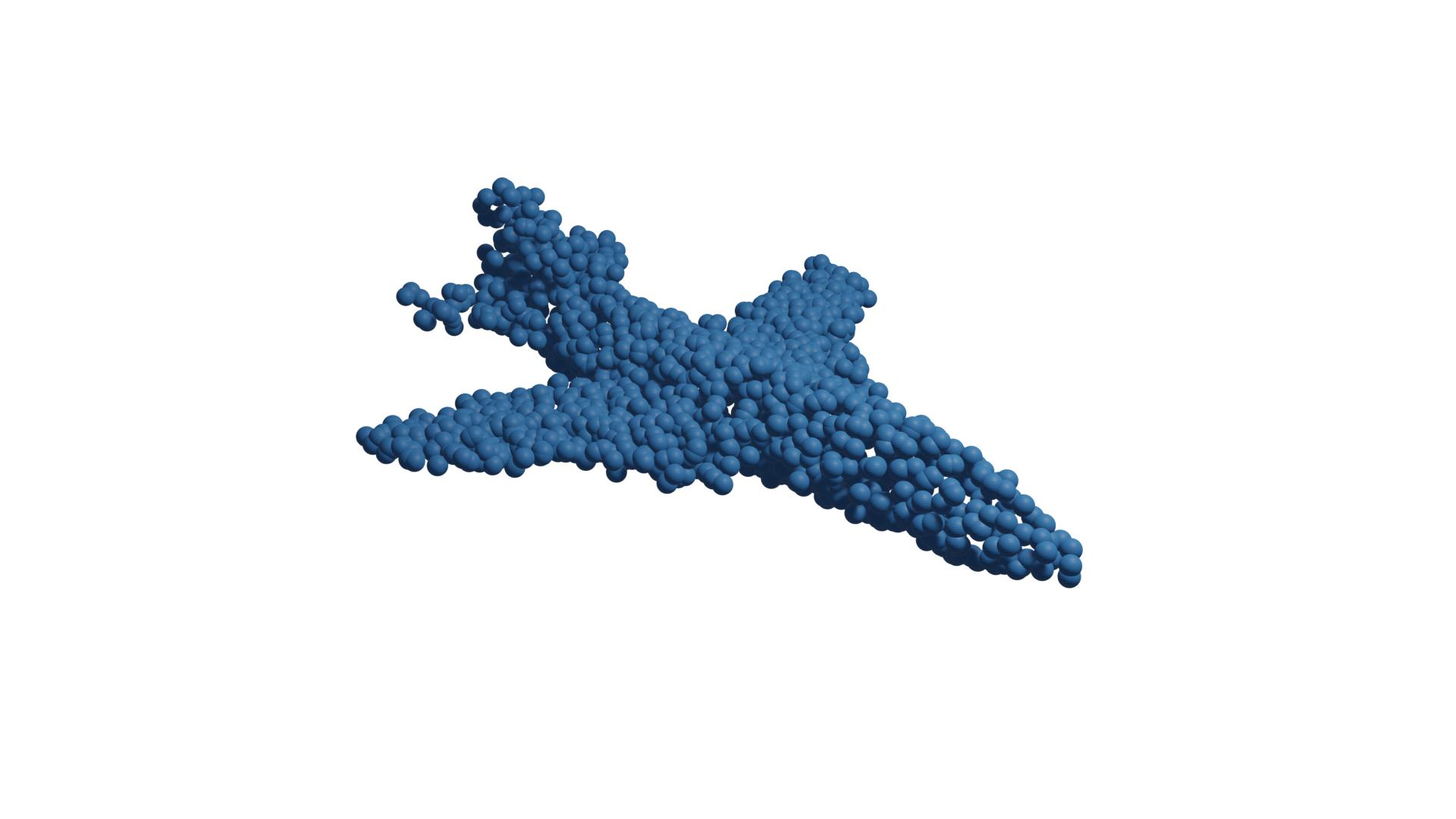}
        \\

         \includegraphics[width=0.4\linewidth,trim={10cm 3cm 10cm 0cm},clip]{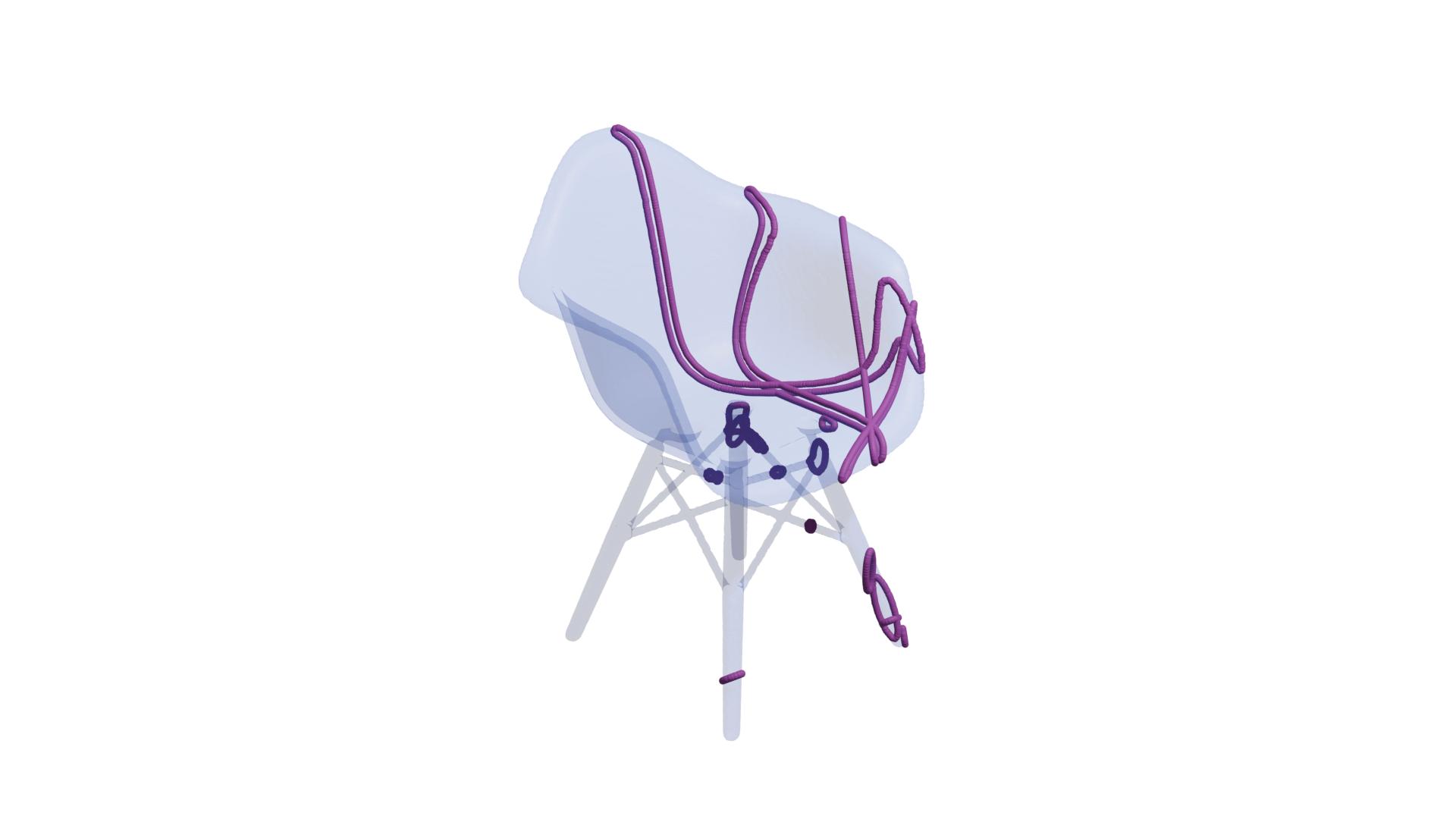}

        &

        \includegraphics[width=0.4\linewidth,trim={10cm 3cm 10cm 0cm},clip]{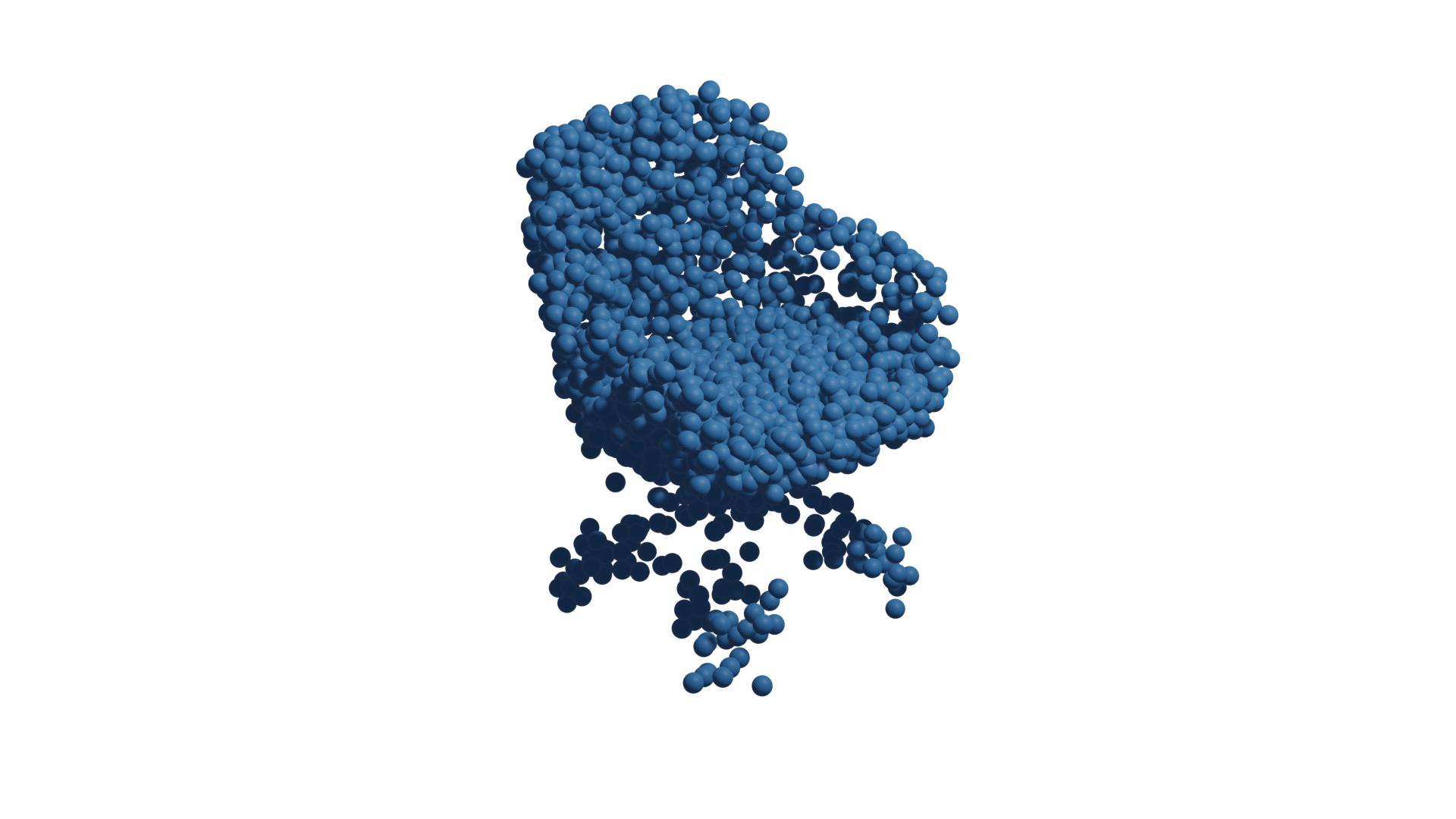}

    \end{tabular}}
    \caption{Failure cases resulting in incorrect shapes. Input to the network are the cross-sections (red) belonging to the ground truth mesh(white).}

    % \vspace{-4mm}
    \label{fig:fail_cases} 
\end{figure}

We observe that in the case of a failure, the network reconstructs a simple object of the class. %and exhibits a mode collapse sort of behaviour. A better training strategy such as WGAN could be used to circumvent the issue. 
However, despite it being a failure case, the class of object is still distinguishable by the network. Further, we also notice a deterioration in the samples containing holes, such as chairs and lamps. In Figure ~\ref{fig:fail_cases} we can see such samples, the reconstruction is not accurate in the case of airplanes. For example, in the case of the chair, the reconstruction does not accurately maintain the genus of the object for some samples.

% \vspace{-7pt}

\section{Comparisons}
\begin{figure*}[!htp]
    \centering
    
    \scalebox{0.75}{
    \begin{tabular}{cccc}
        \rotatebox{90}{\textbf{GT Mesh + Input }} &

        \includegraphics[width=0.27\linewidth,trim={10cm 3cm 10cm 3cm},clip]{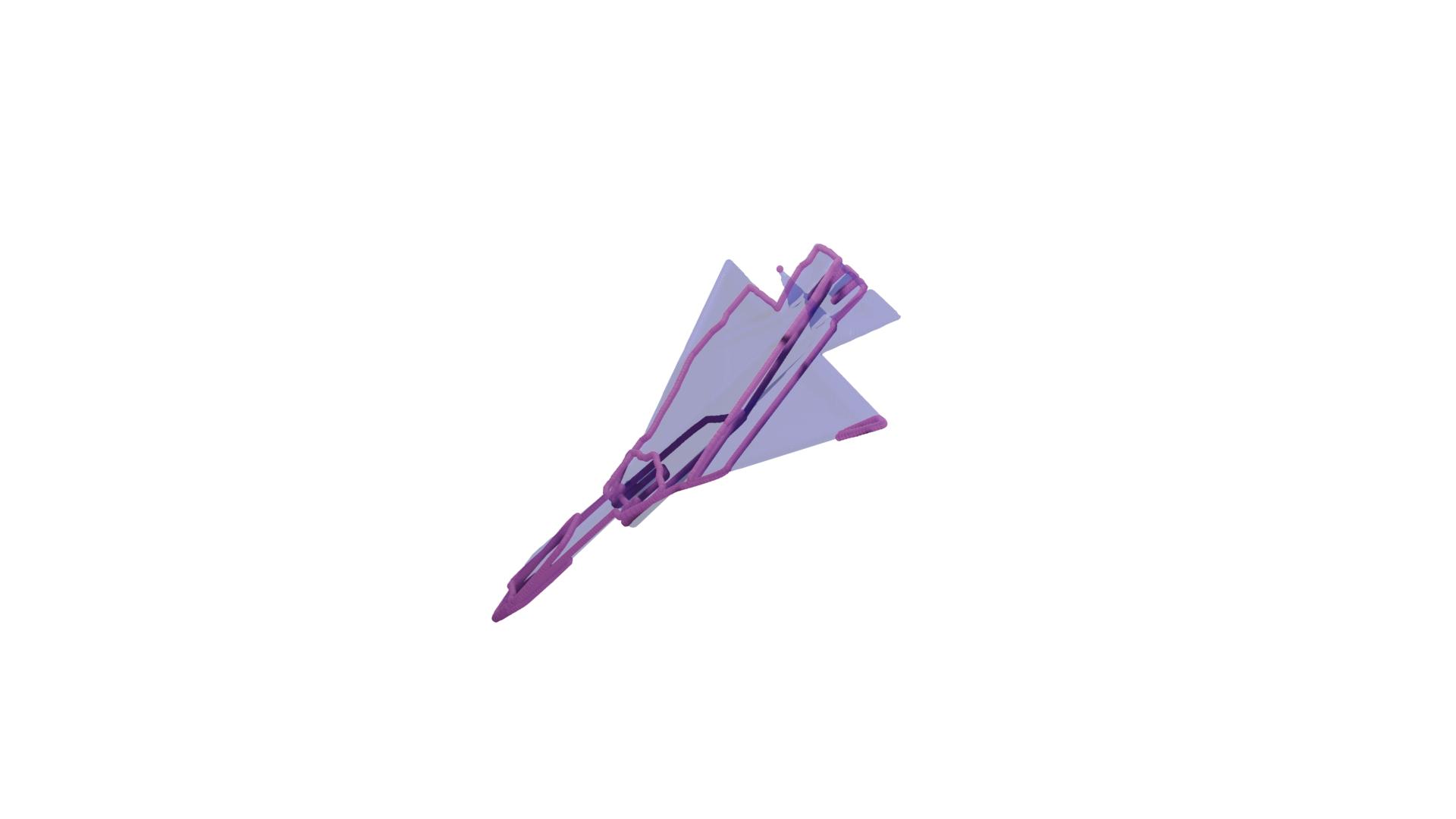}
        
        &
        \includegraphics[width=0.2\linewidth,trim={10cm 3cm 10cm 3cm},clip]{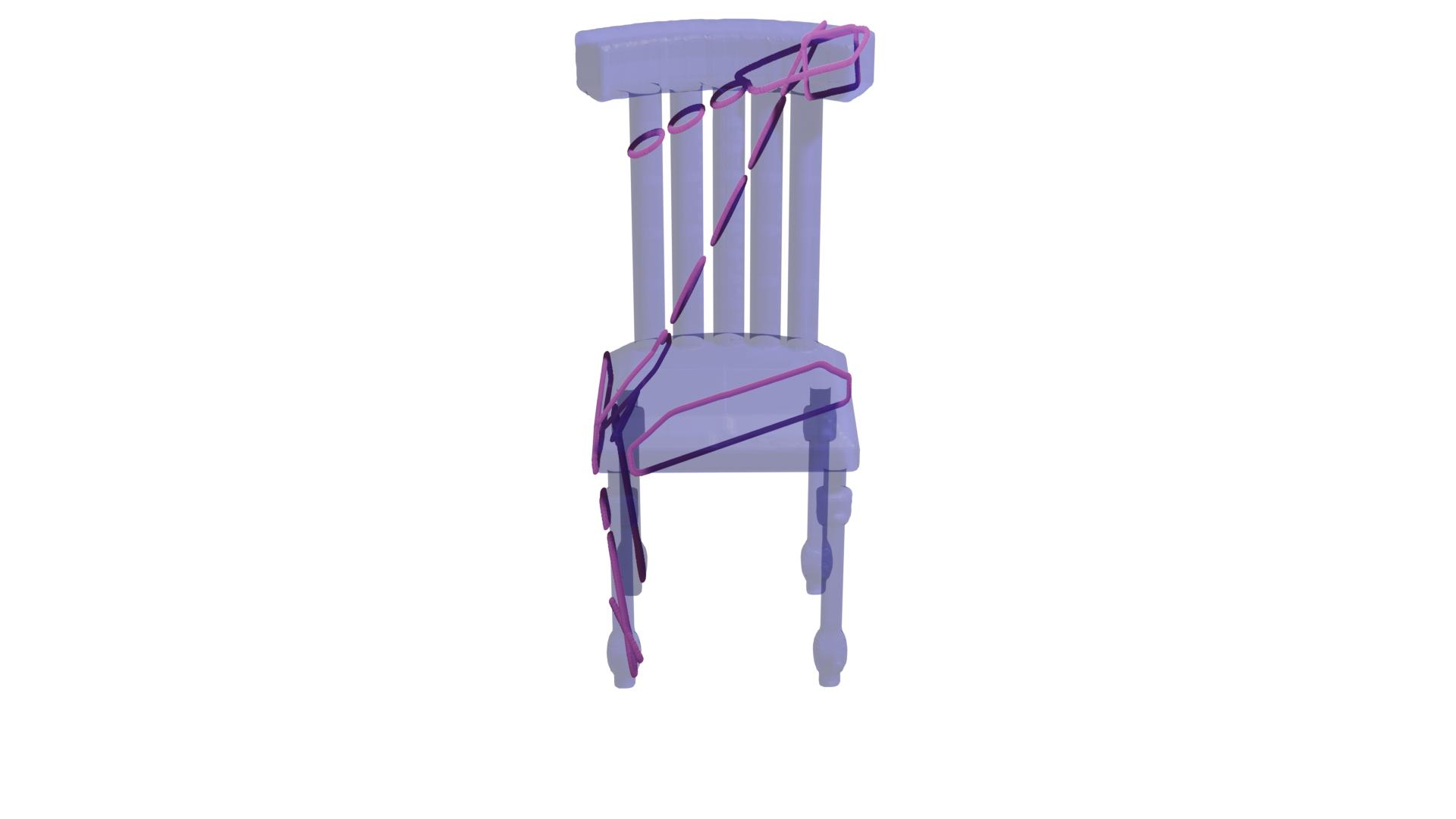}

    &
    
        \includegraphics[width=0.2\linewidth,trim={10cm 3cm 10cm 3cm},clip]{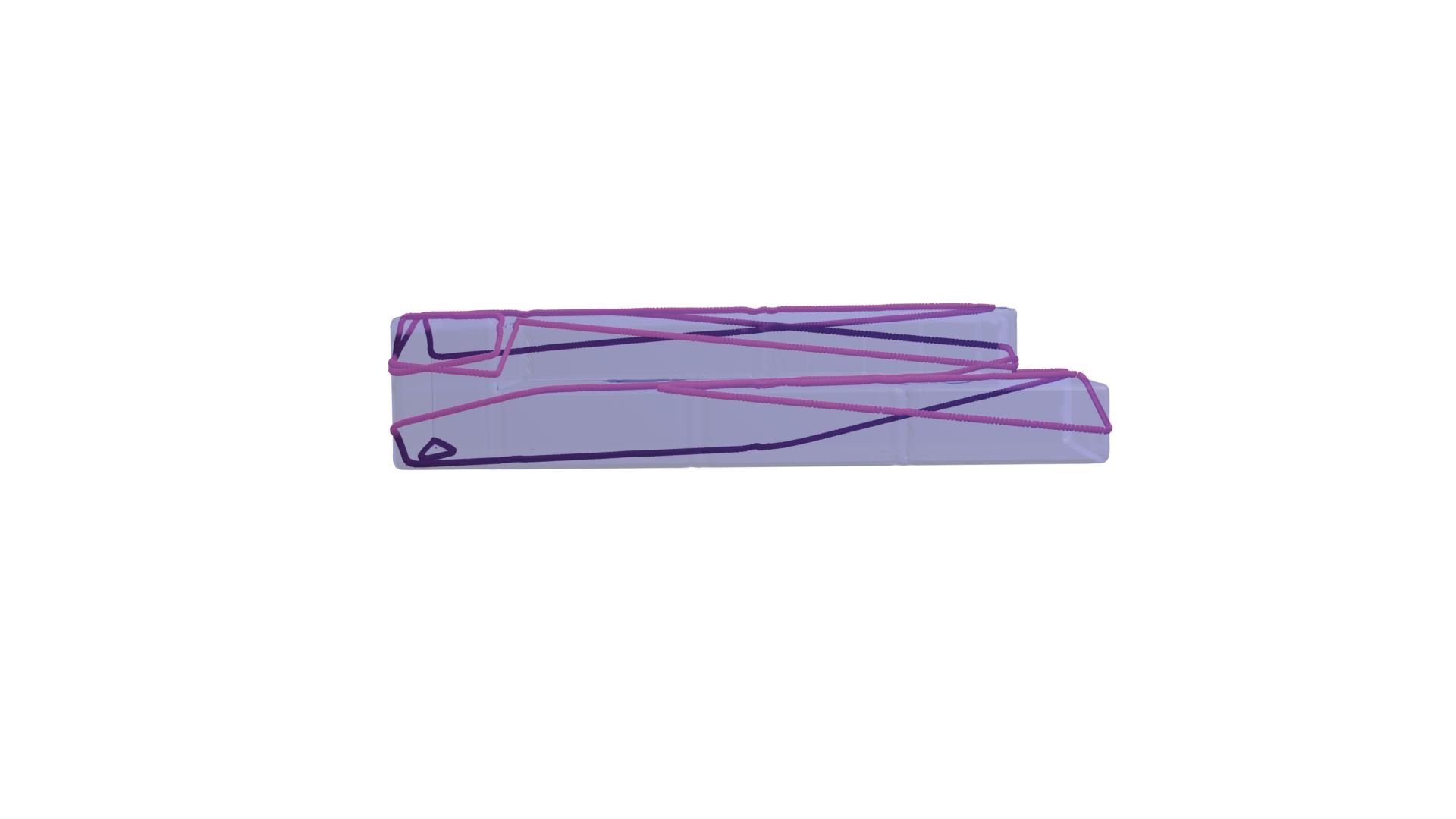}
    % \includegraphics[width=0.25\linewidth,trim={10cm 3cm 10cm 3cm},clip]{blender_shadows_all/comparisons/sofa_2_exp_adj.png}
    
    % \includegraphics[width=0.25\linewidth,trim={8cm 3cm 7cm 3cm},clip]{blender_shadows_all/comparisons/sofa_soft_shadow_exp_adj.png}

        % \includegraphics[width=0.15\textwidth,trim={0cm 5cm 0cm 0cm},clip]{images/comparisons/11_crop/gt_mesh_with_cross_lines_red00_crop.png} &
       
        % \includegraphics[height=0.15\textwidth,trim={5cm 0cm 5cm 0cm},clip]{inset_images/chair/gt_cross00_crop_2.png} 
        
        % &
     
        %  \includegraphics[height=0.08\textwidth, trim={5cm 2cm 4cm 5cm},clip]{inset_images/sofa/same_orientation/gt_crop.png}
         
        \\

\rotatebox{90}{\textbf{VIPSS} \cite{huang2019variational}} &
        \begin{tikzpicture}
        \node (n0)  {\includegraphics[width=0.3\textwidth]{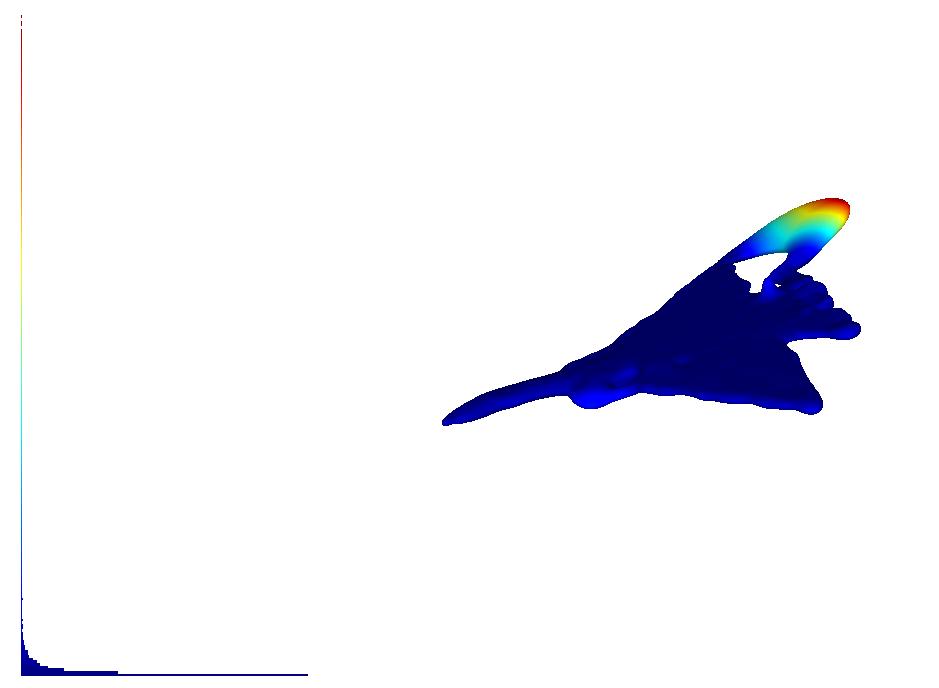}};
        \node[draw=black, below right=of n0.north west] at (-3,2.7)
            {\includegraphics[width=0.15\textwidth,trim={12cm 9cm 12cm 8cm}, clip]{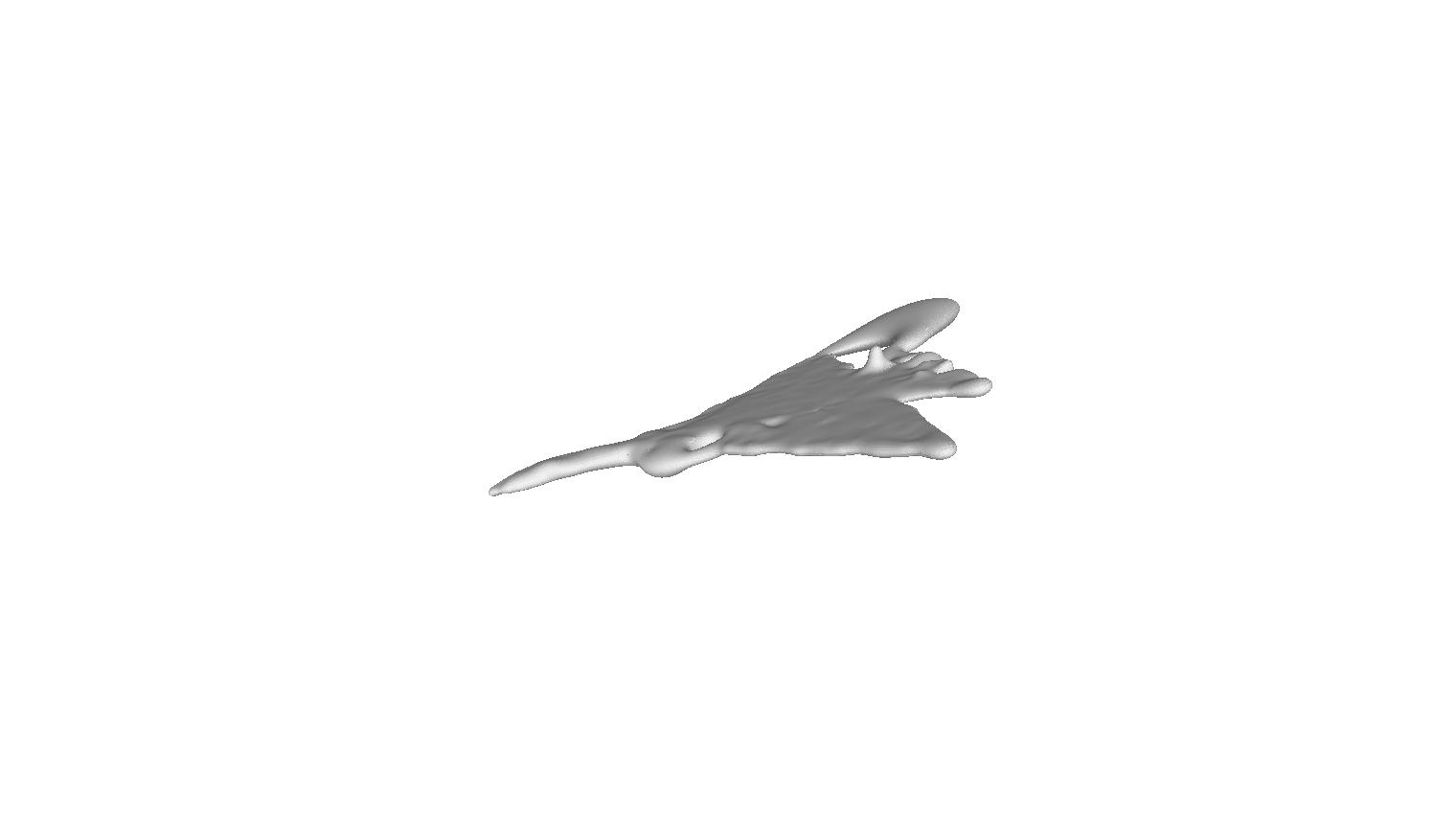}};
        \end{tikzpicture} &
        
        \begin{tikzpicture}
        \node (n0)  {\includegraphics[width=0.3\textwidth]{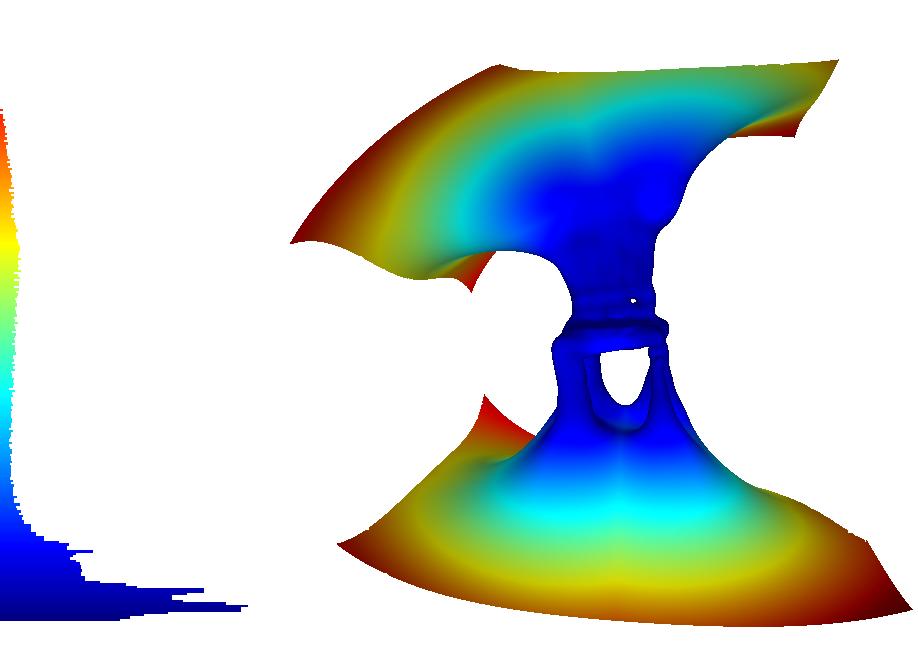}};
        \node[draw=black, below right=of n0.north west] at (-3.5,2.7)
            {\includegraphics[width=0.08\textwidth,trim={15cm 6cm 15cm 6cm}, clip]{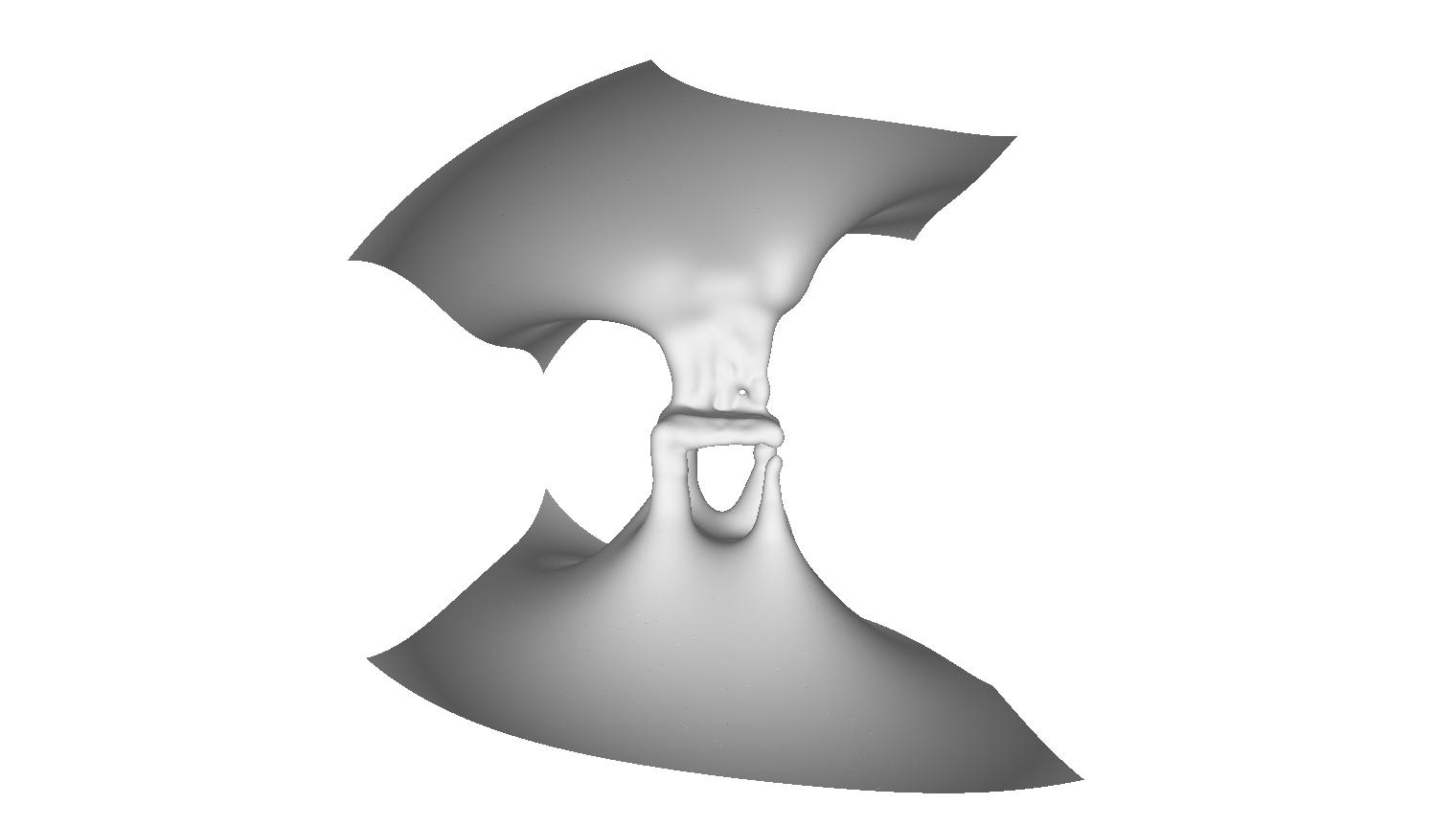}};
        \end{tikzpicture}

         &
         
          \begin{tikzpicture}
        % \node (n0)  {\includegraphics[width=0.3\textwidth]{inset_images/sofa/sofa_heatmap_vipss.png}};
        \node (n0)  {\includegraphics[width=0.3\textwidth]{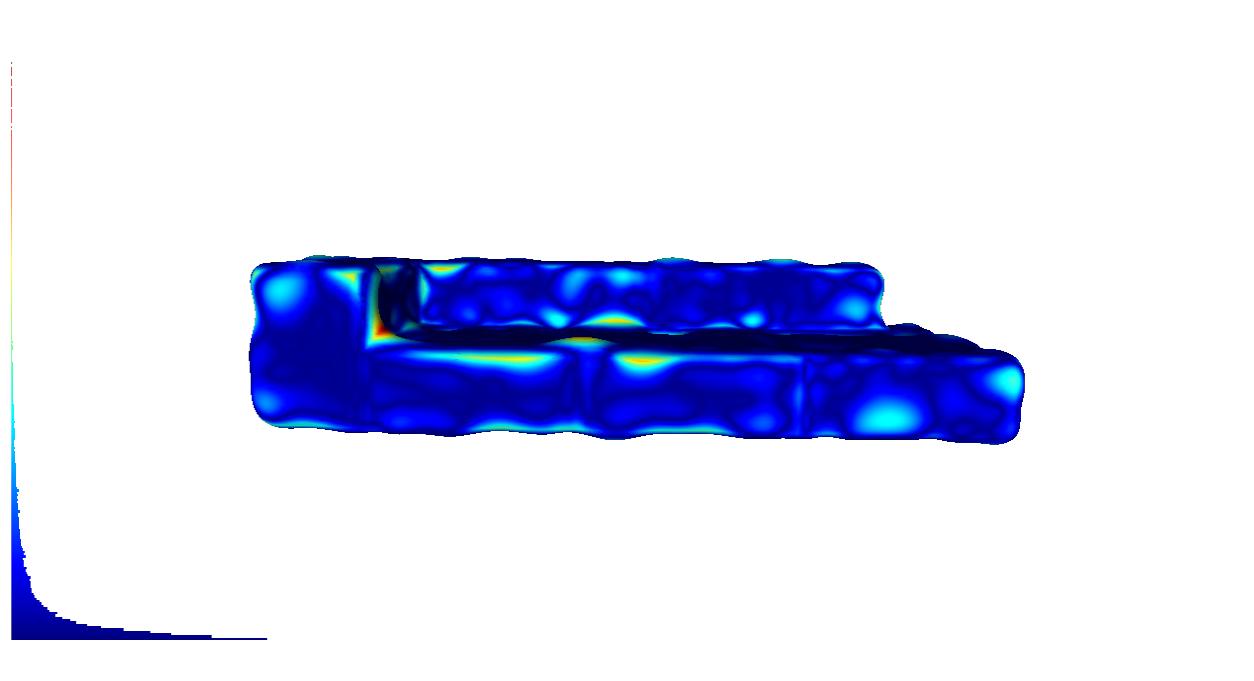}};
        \node[draw=black, below right=of n0.north west] at (-3,2.7)
            {\includegraphics[width=0.15\textwidth,trim={7cm 9cm 7cm 9cm}, clip]{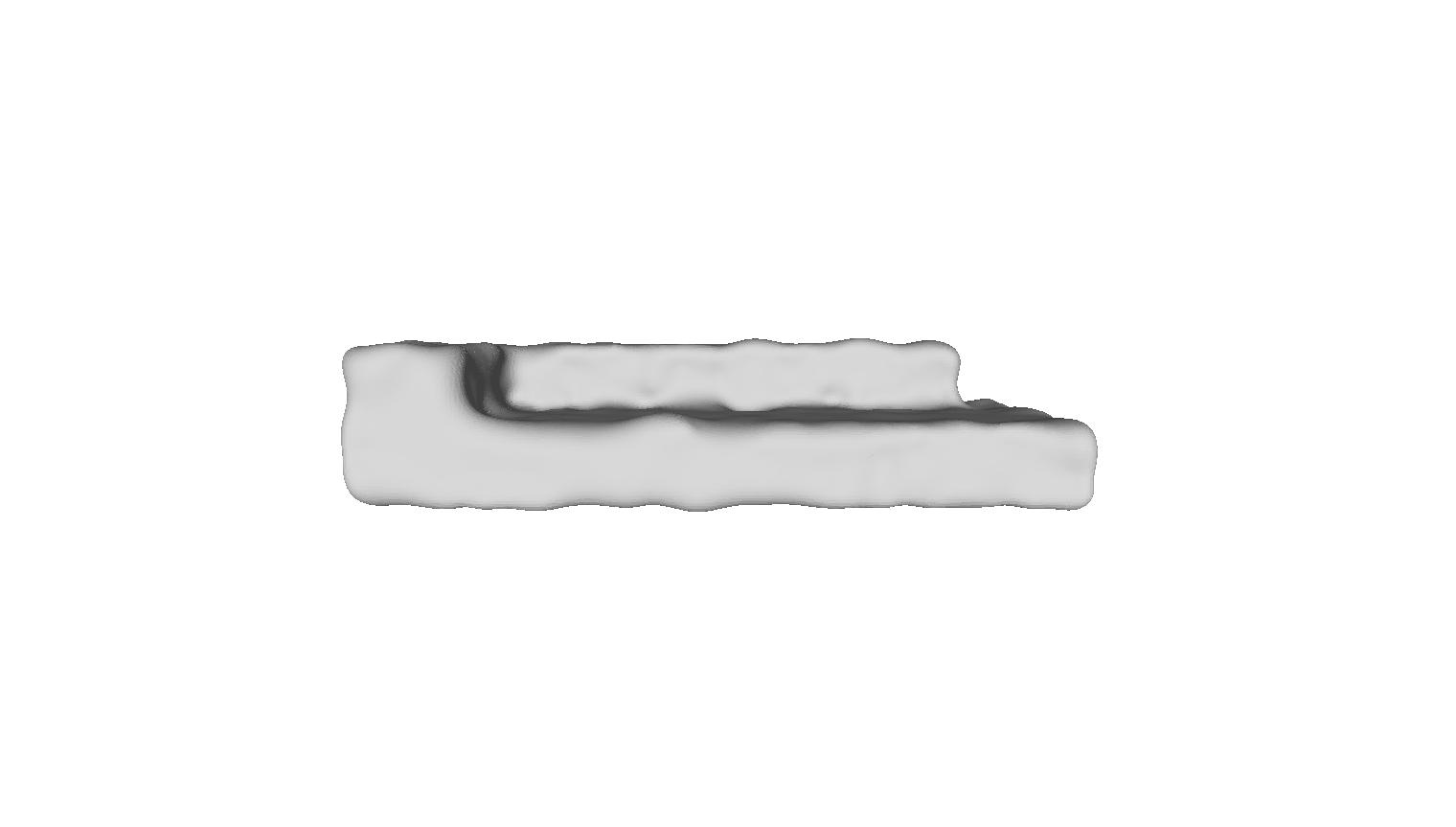}};
        \end{tikzpicture}

         \\
         & $d_H: 0.032574$ & $d_H: 0.056161$ & $d_H:  0.001533$
         %sofa $d_H: 0.001533$
         \\
         
         \rotatebox{90}{\textbf{Liu et. al} \cite{liu2008surface}} & 
        \begin{tikzpicture}
        \node (n0)  {\includegraphics[width=0.3\textwidth]{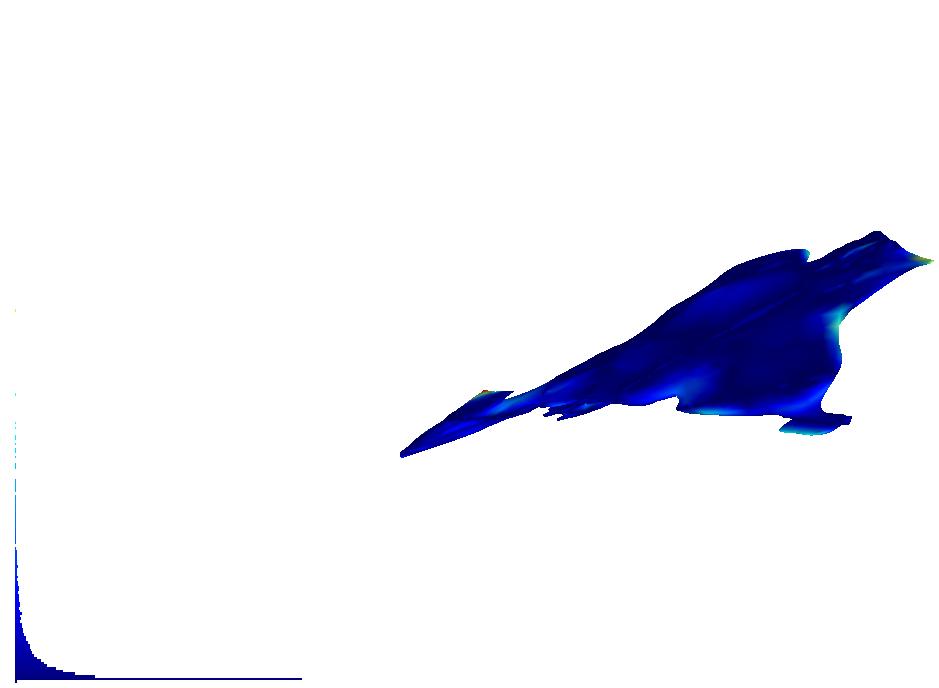}};
        \node[draw=black, below right=of n0.north west] at (-3,2.7)
            {\includegraphics[width=0.15\textwidth,trim={12cm 9cm 12cm 9cm}, clip]{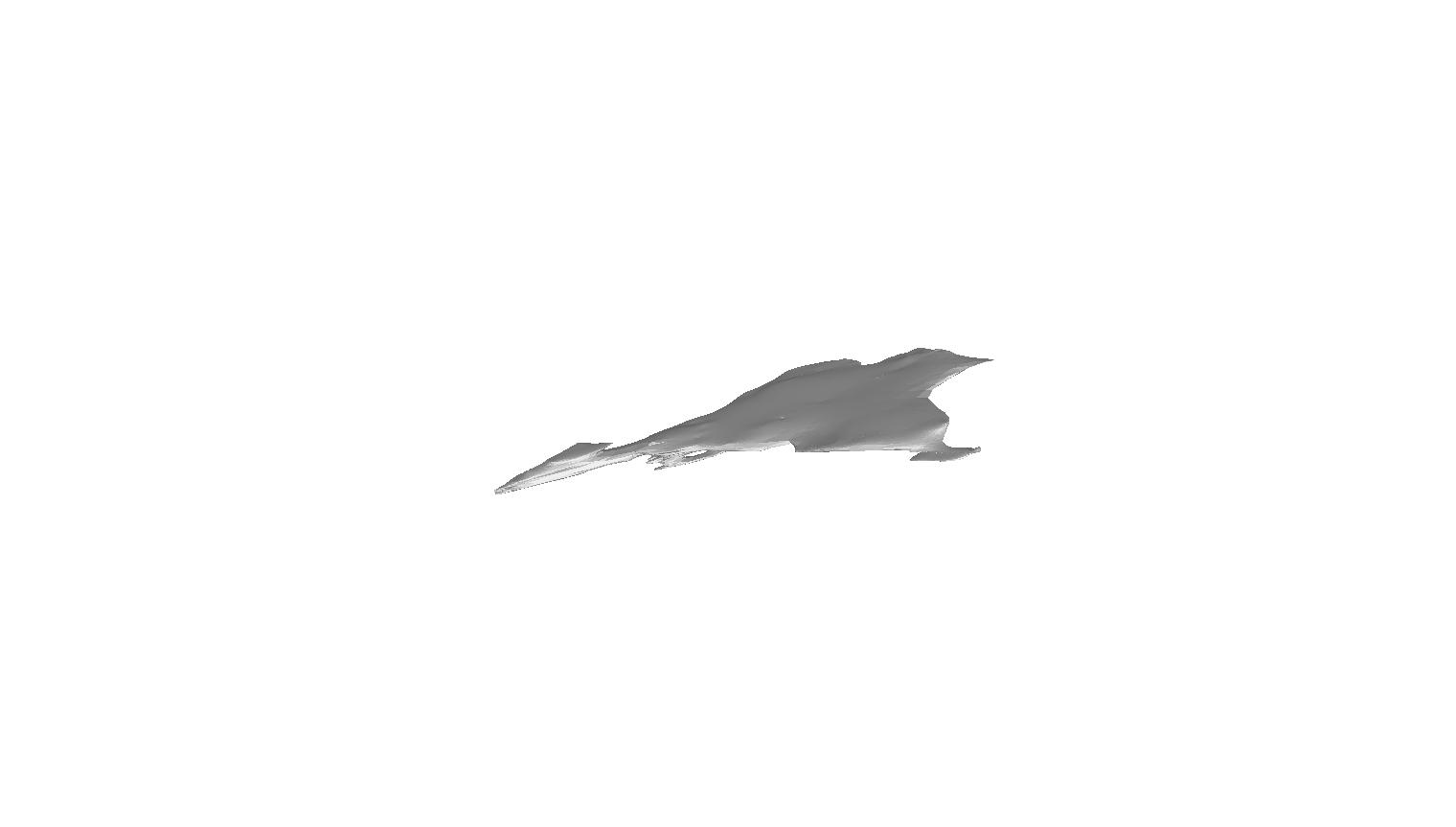}};
        \end{tikzpicture}

         &

          \begin{tikzpicture}
        \node (n0)  {\includegraphics[width=0.3\textwidth]{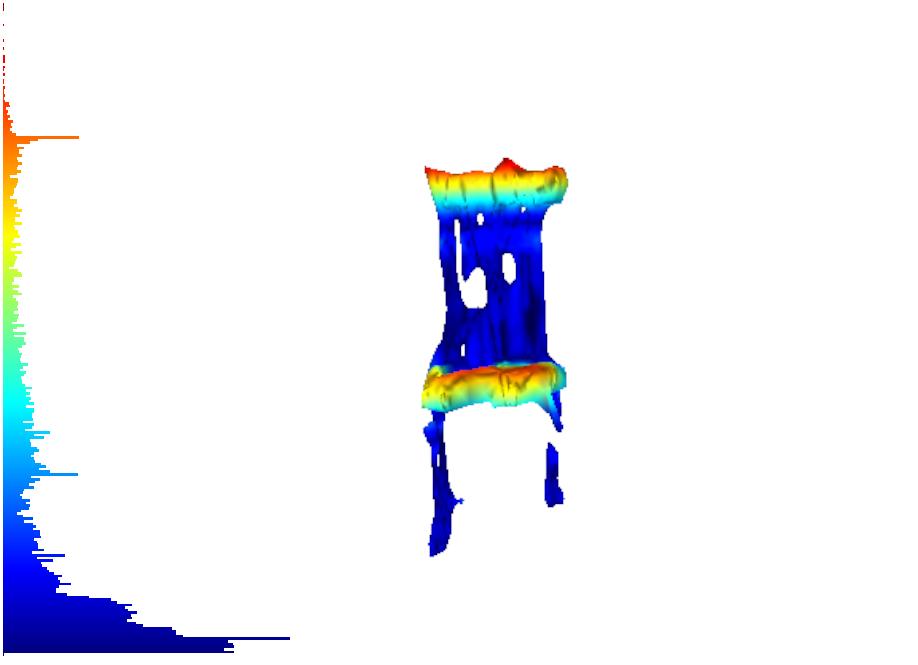}};
        \node[draw=black, below right=of n0.north west] at (-3,2.7)
            {\includegraphics[width=0.08\textwidth,trim={16cm 8.5cm 16cm 6cm}, clip]{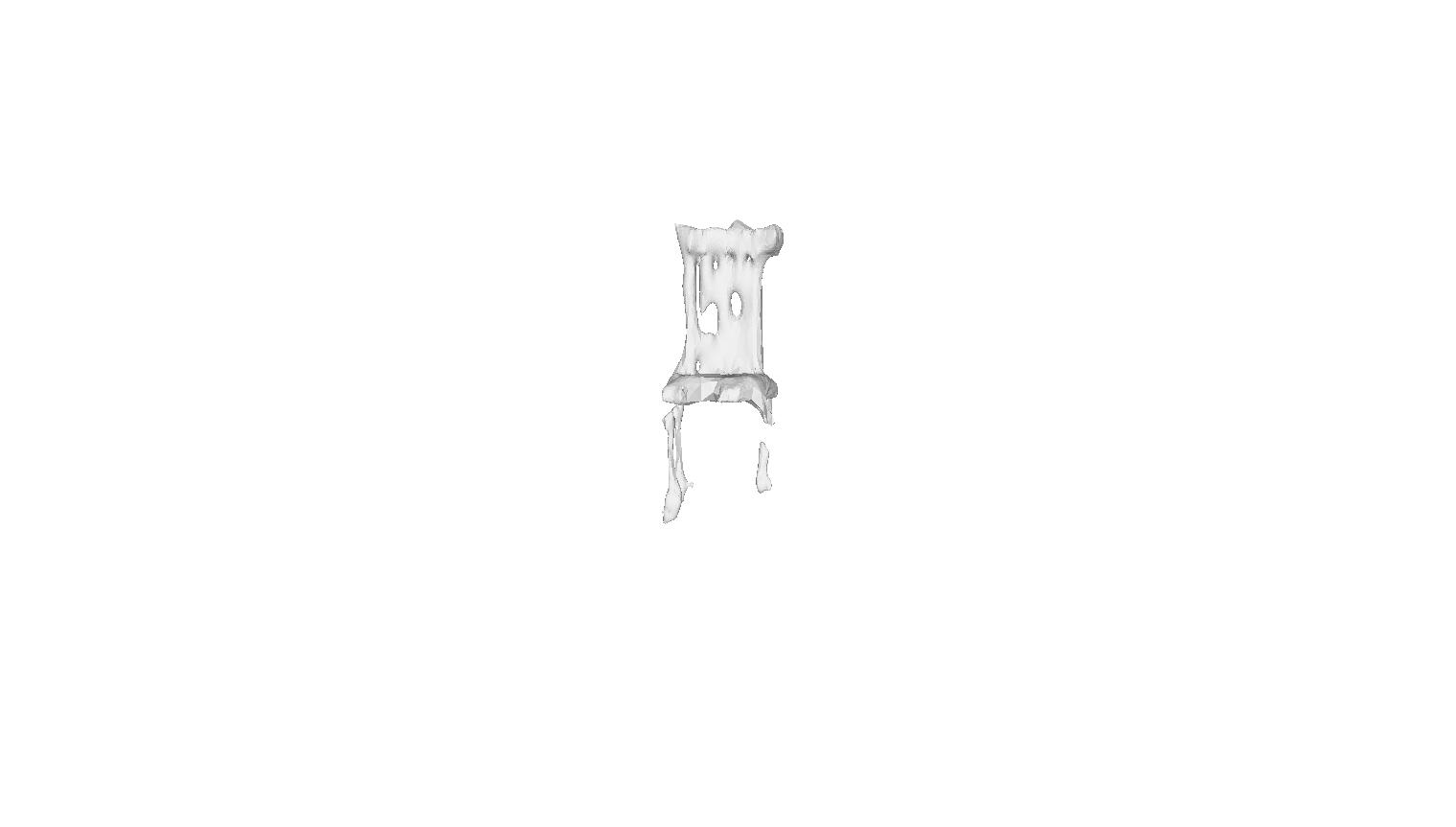}};
        \end{tikzpicture}

         &

          \begin{tikzpicture}
        % \node (n0)  {\includegraphics[width=0.3\textwidth]{inset_images/sofa/sofa_heatmap_liu.png}};
         \node (n0)  {\includegraphics[width=0.3\textwidth]{inset_images/sofa/same_orientation/liu_heatmap.jpg}};
        \node[draw=black, below right=of n0.north west] at (-3,2.7)
            {\includegraphics[width=0.15\textwidth,trim={8cm 9cm 7cm 9cm}, clip]{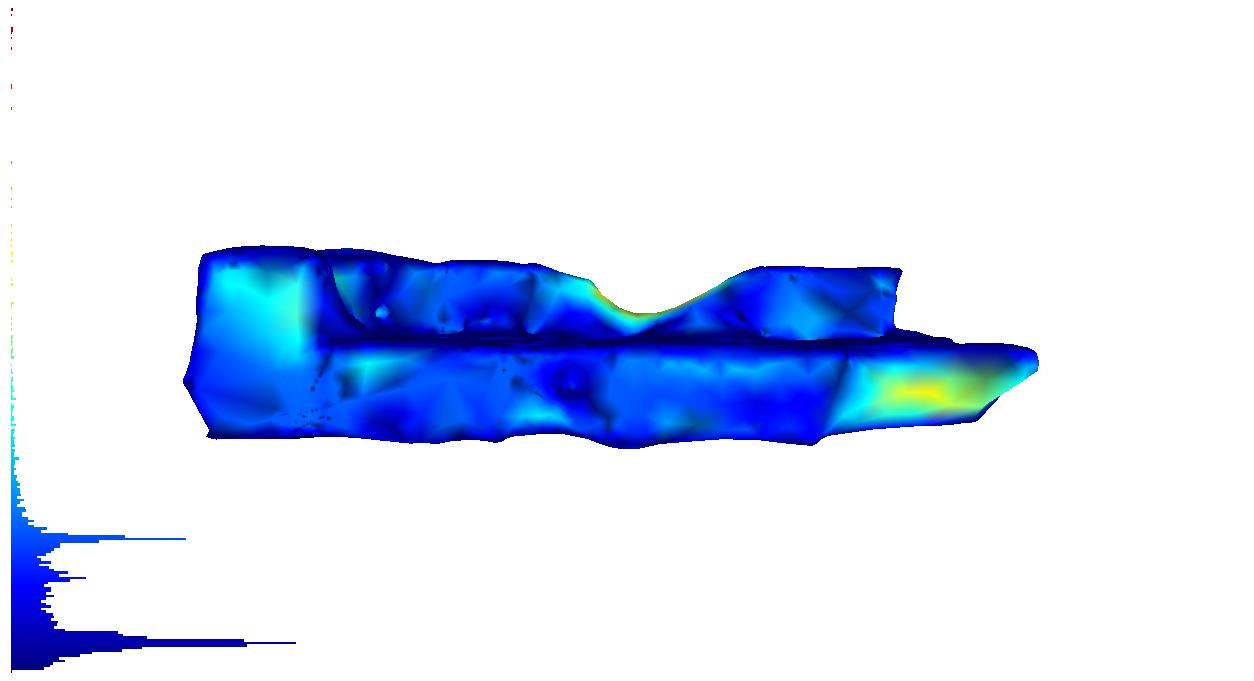}};
        \end{tikzpicture}
         \\
         & $d_H: 0.005765$ & $d_H: 0.039835$ & $d_H: 0.010035$\\
         
         \rotatebox{90}{\textbf{P2P Net} \cite{yin2018p2p}} &
         \begin{tikzpicture}
        \node (n0)  {\includegraphics[width=0.3\textwidth]{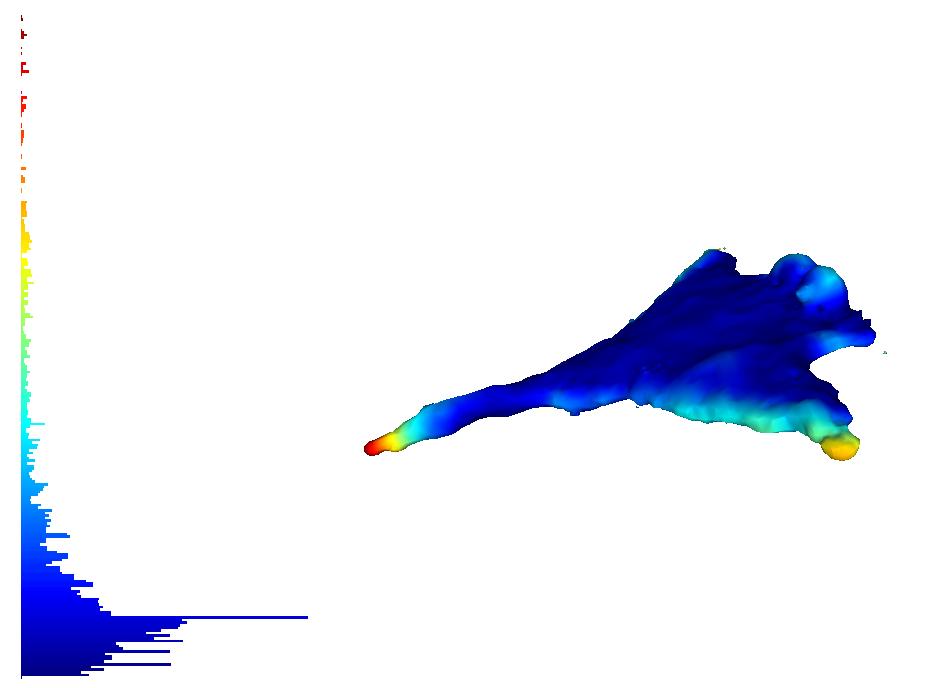}};
        \node[draw=black, below right=of n0.north west] at (-3,2.7)
            {\includegraphics[width=0.15\textwidth,trim={10cm 8cm 12cm 8cm}, clip]{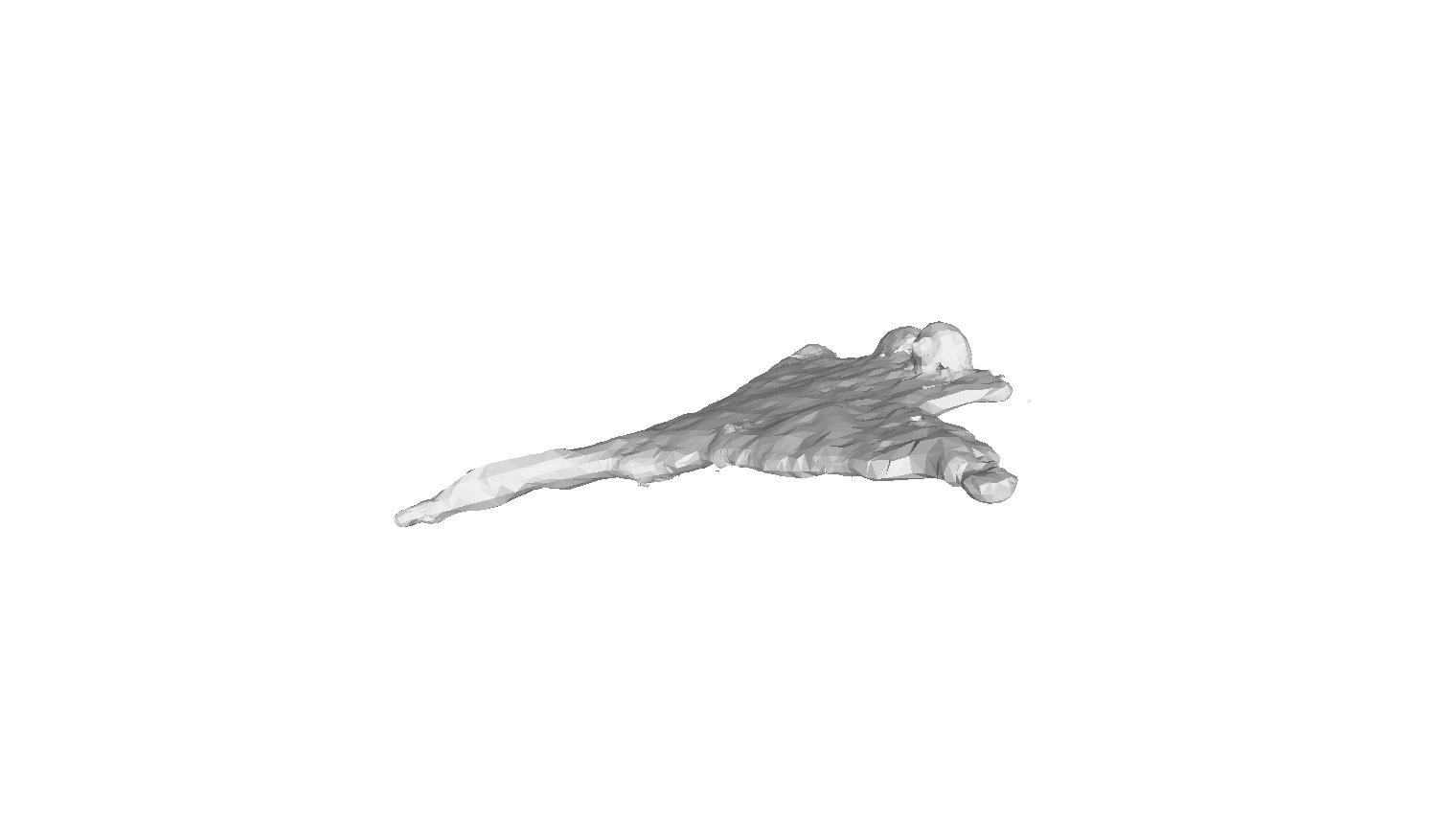}};
        \end{tikzpicture}
         &

         \begin{tikzpicture}
        \node (n0)  {\includegraphics[width=0.3\textwidth]{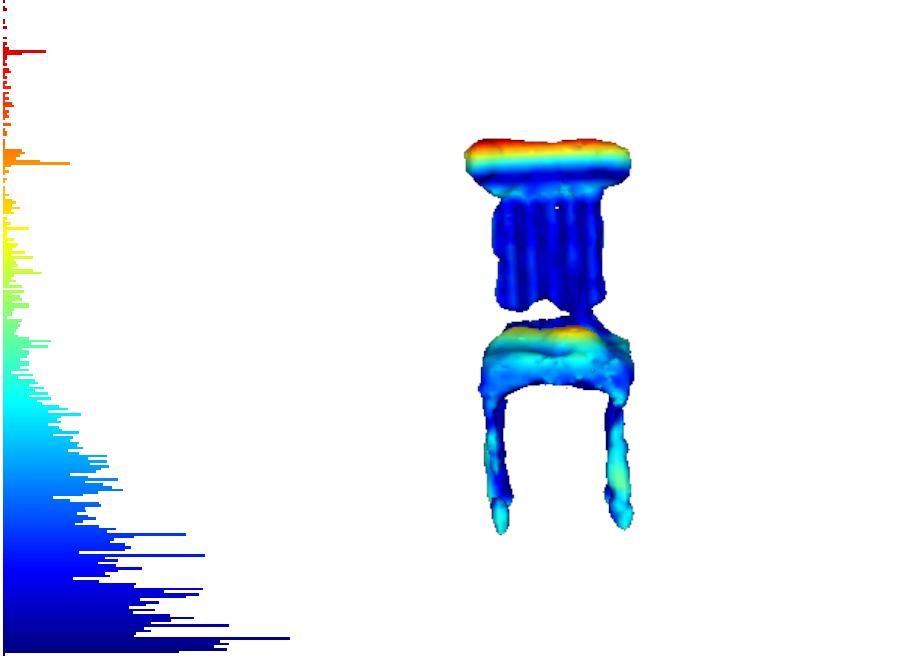}};
        \node[draw=black, below right=of n0.north west] at (-3,2.7)
            {\includegraphics[width=0.08\textwidth,trim={15cm 7cm 15cm 6cm}, clip]{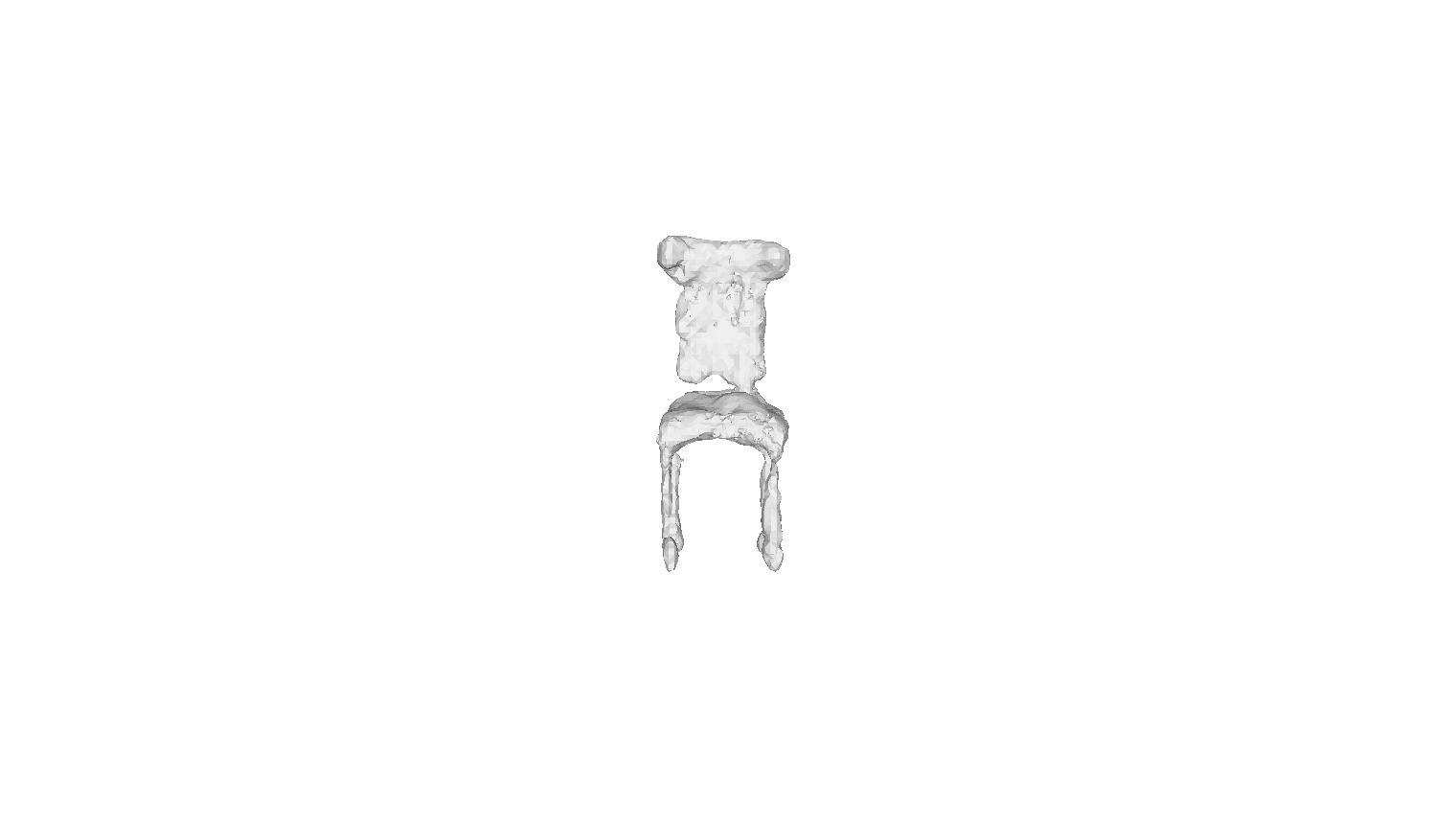}};
        \end{tikzpicture}

        & 

          \begin{tikzpicture}
        % \node (n0)  {\includegraphics[width=0.3\textwidth]{inset_images/sofa/sofa_heatmap_p2p.png}};
        \node (n0)  {\includegraphics[width=0.3\textwidth]{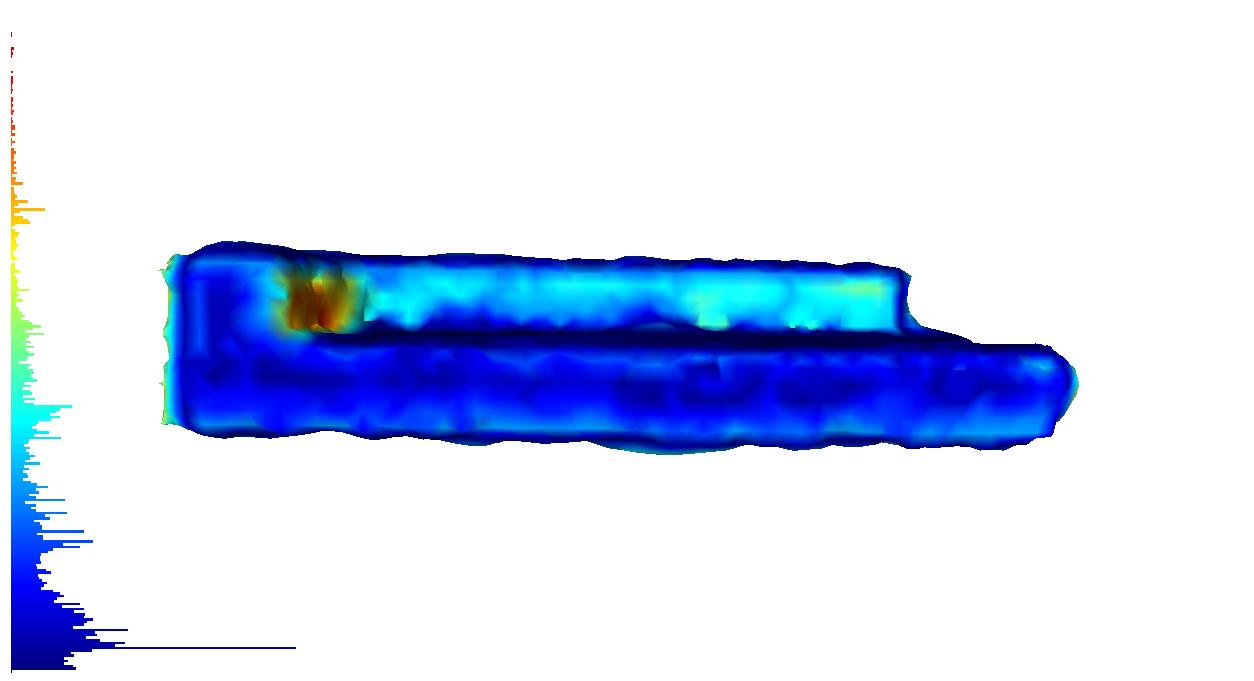}};
        \node[draw=black, below right=of n0.north west] at (-3,2.7)
            {\includegraphics[width=0.15\textwidth,trim={7cm 9cm 7cm 9cm}, clip]{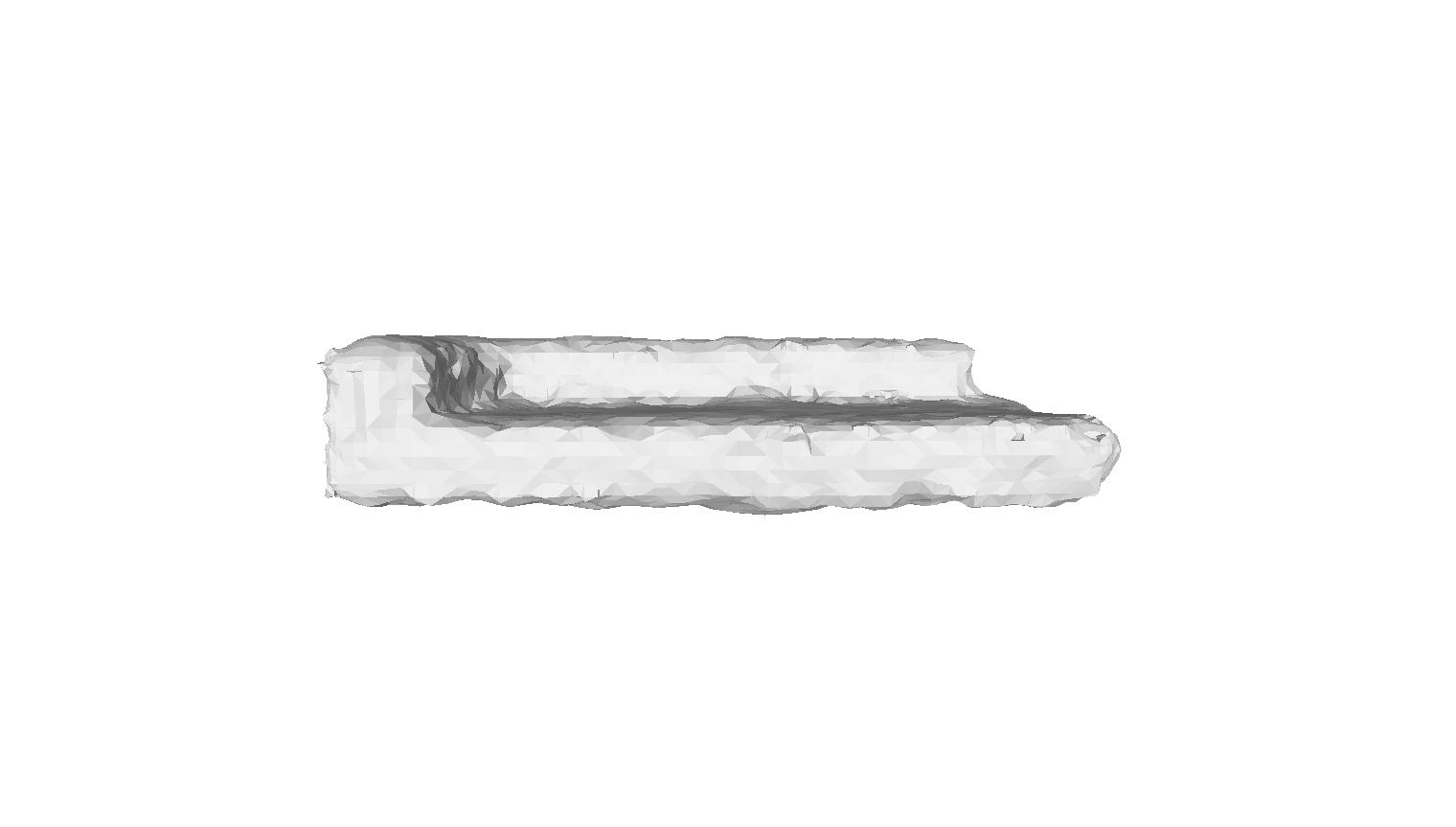}};
        \end{tikzpicture}
         \\
         & $d_H: 0.023185$ & $d_H: 0.032854$ & $d_H: 0.012473$\\

     \rotatebox{90}{\textbf{ORex} \cite{sawdayee2022orex}} & 

        \begin{tikzpicture}
        \node (n0)  {\includegraphics[width=0.35\textwidth]{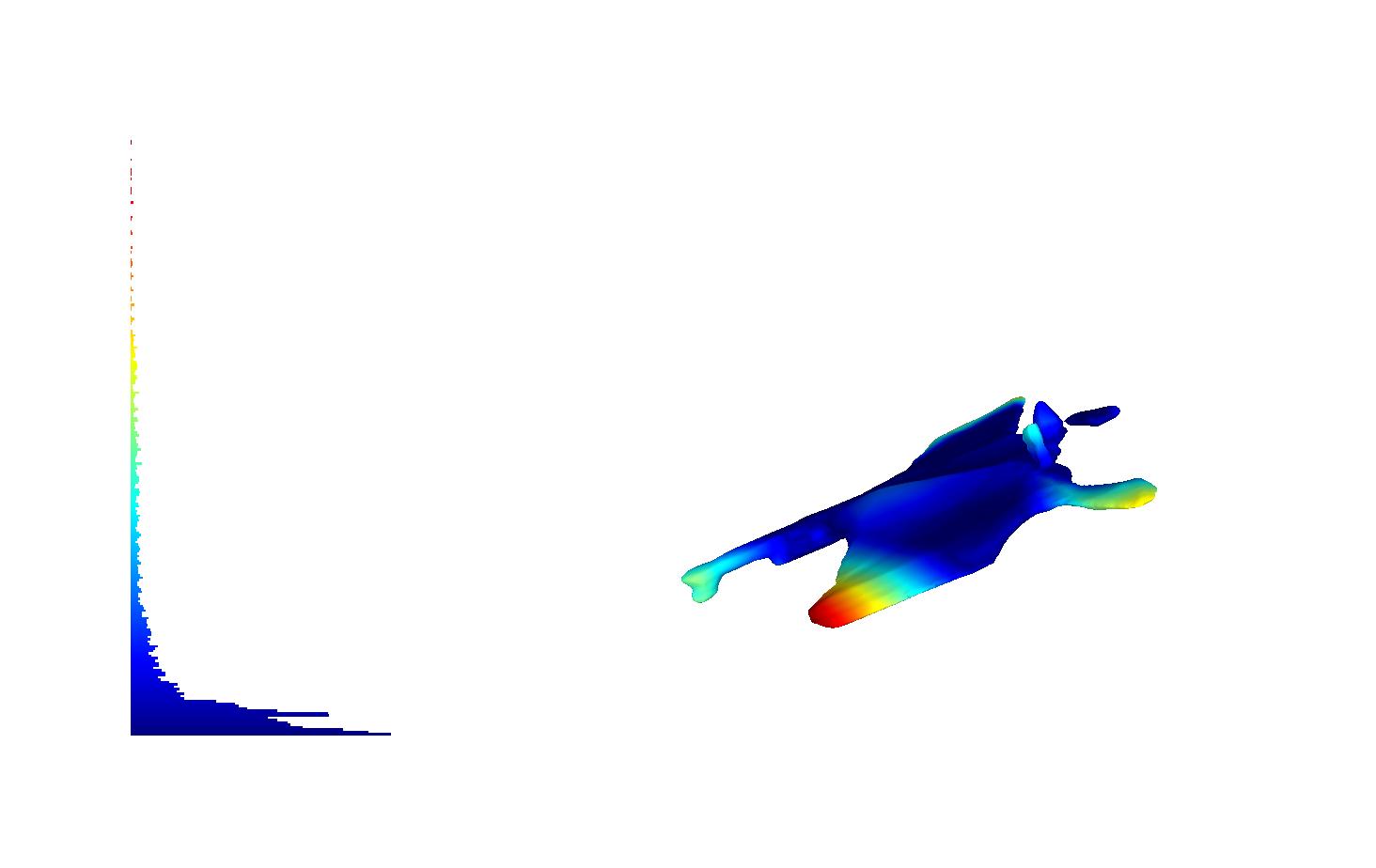}};
        \node[draw=black, below right=of n0.north west] at (-3,2.7)
            {\includegraphics[width=0.15\textwidth,trim={10cm 8cm 10cm 8cm}, clip]{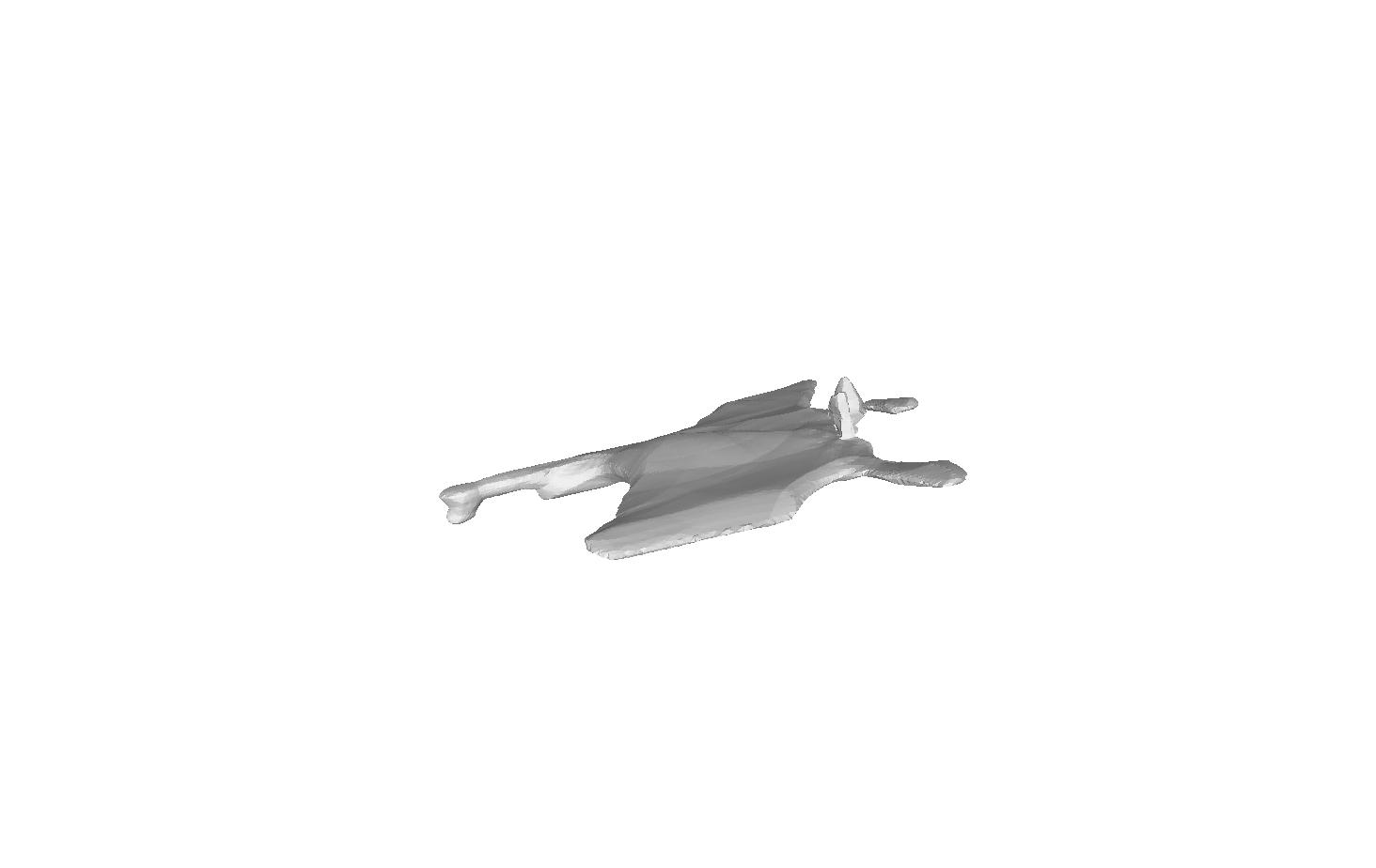}};
        \end{tikzpicture}
        
        &
         \begin{tikzpicture}
        \node (n0)  {\includegraphics[width=0.32\textwidth]{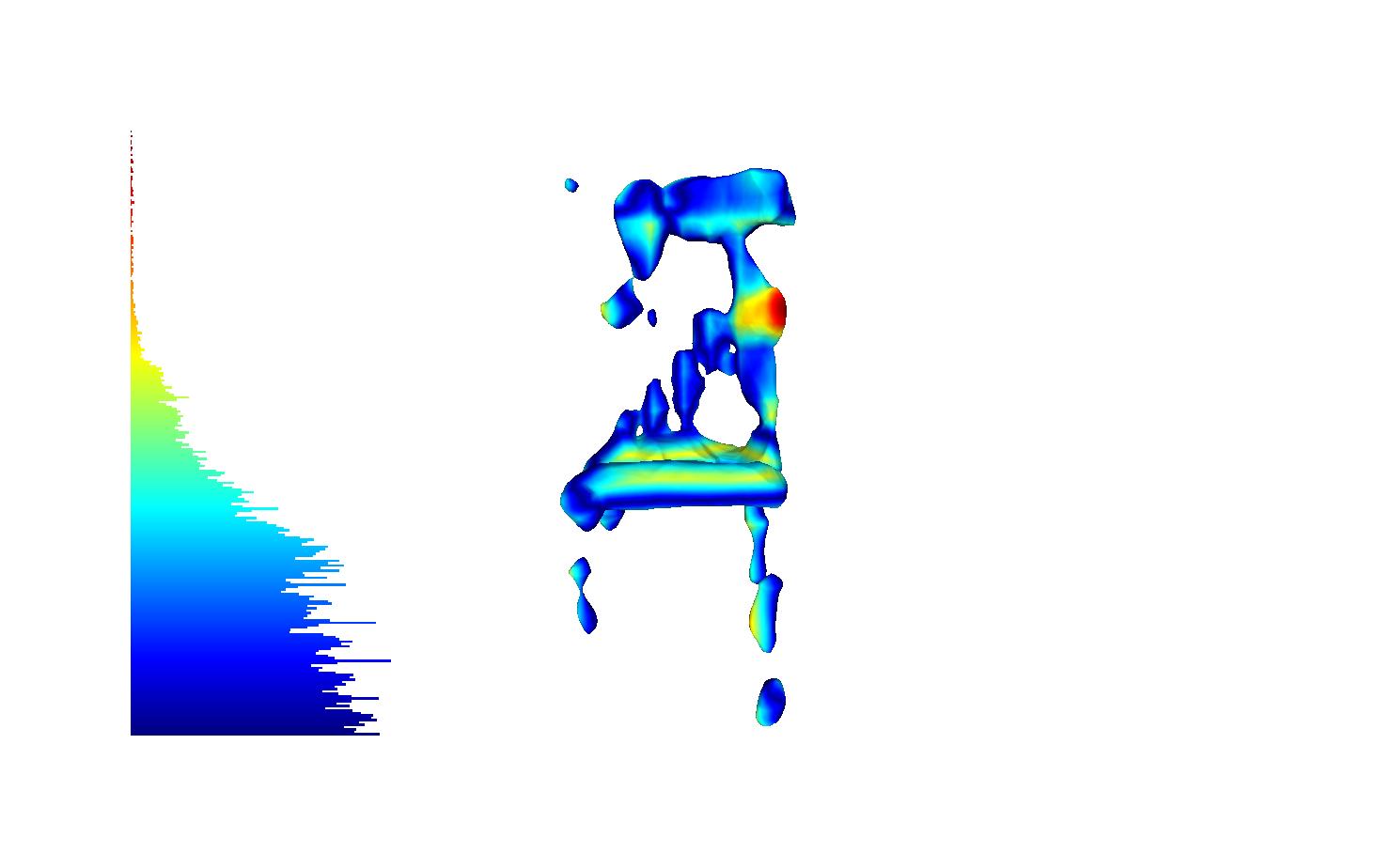}};
        \node[draw=black, below right=of n0.north west] at (-3,2.7)
            {\includegraphics[width=0.06\textwidth,trim={15cm 7cm 15cm 6cm}, clip]{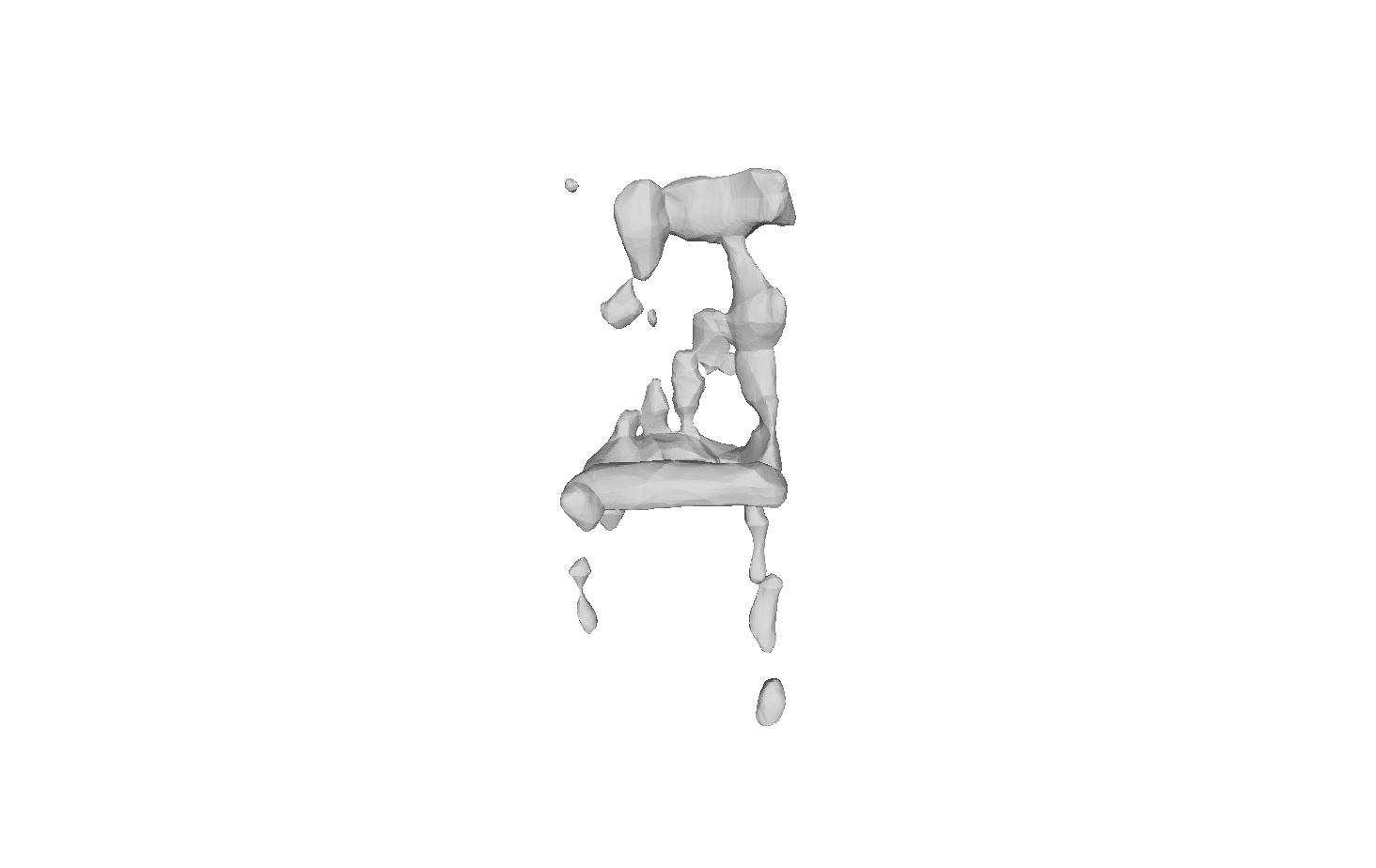}};
        \end{tikzpicture}

        &

         \begin{tikzpicture}
        \node (n0)  {\includegraphics[width=0.4\textwidth]{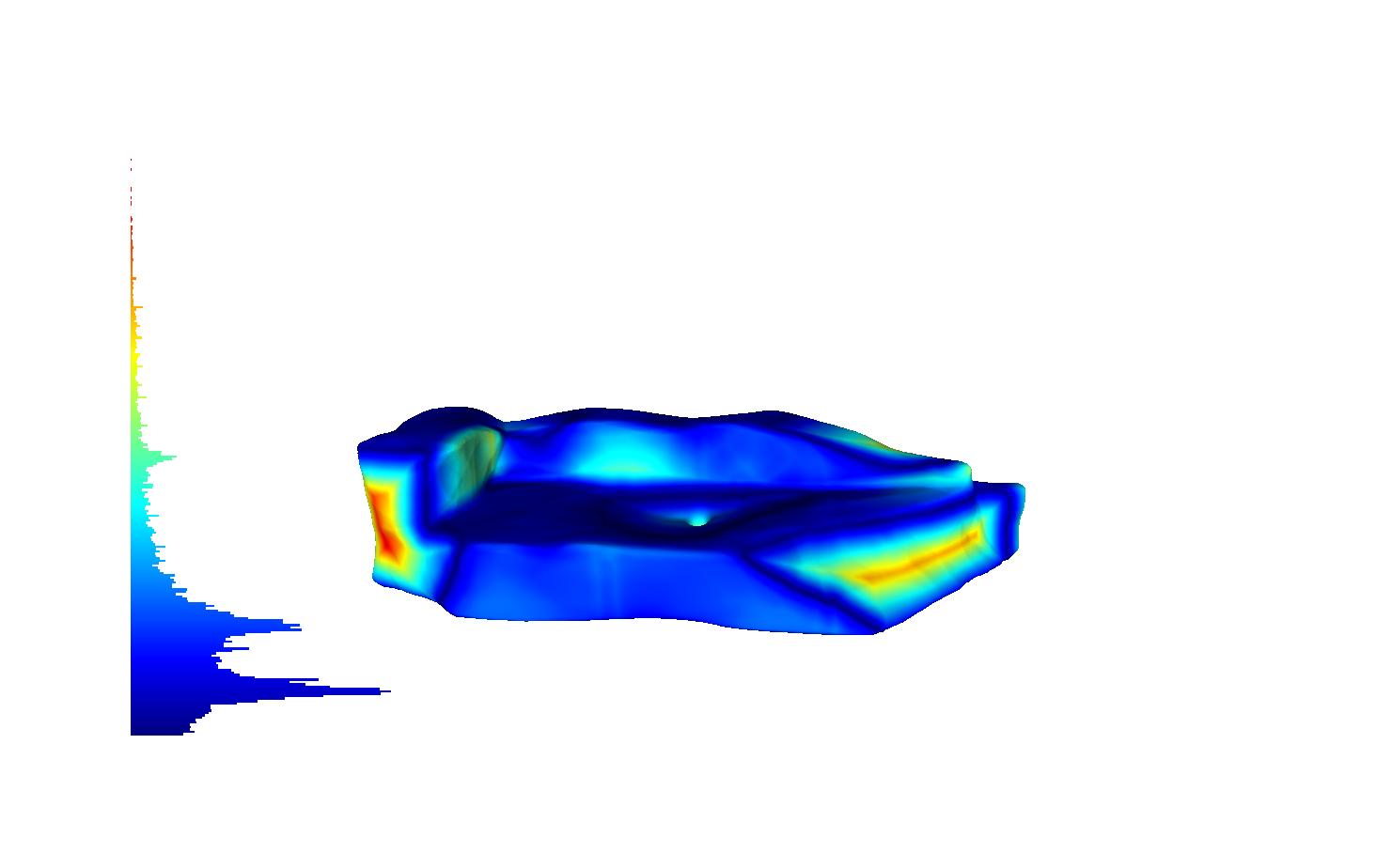}};
        \node[draw=black, below right=of n0.north west] at (-3,2.7)
            {\includegraphics[width=0.15\textwidth,trim={7cm 8cm 7cm 9cm}, clip]{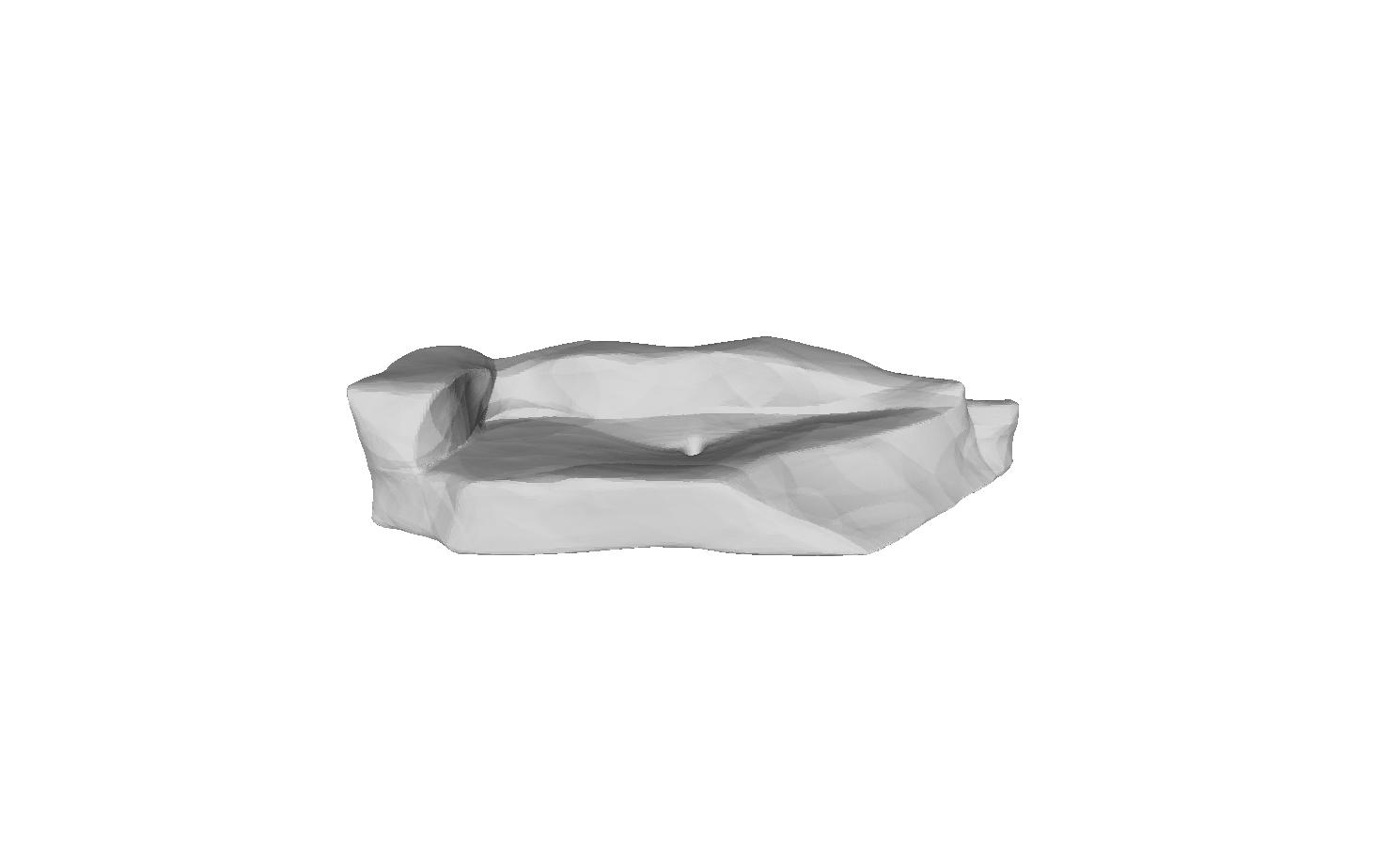}};
        \end{tikzpicture}
        \\
     %    % \includegraphics[width=0.3\textwidth]{inset_images/airplane/gcn_heatmaps_inset_model.png} 
     %    \begin{tikzpicture}
     %    \node (n0)  {\includegraphics[width=0.3\textwidth]{images/inset_images/airplane/orex/heatmap_airplane00.png}};
     %    \node[draw=black, below right=of n0.north west] at (-3,2.7)
     %        {\includegraphics[width=0.15\textwidth,trim={10cm 8cm 10cm 8cm}, clip]{images/inset_images/airplane/orex/recons01.png}};
     %    \end{tikzpicture}
        
     %    &
     %    % \includegraphics[width=0.3\textwidth]{inset_images/chair/chair_heatmaps_gcn_cropped_inset_model.png} 
     %     \begin{tikzpicture}
     %    \node (n0)  {\includegraphics[width=0.28\textwidth]{images/inset_images/chair/orex/chair_recons_heatmap01.png}};
     %    \node[draw=black, below right=of n0.north west] at (-3,2.7)
     %        {\includegraphics[width=0.08\textwidth,trim={15cm 7cm 15cm 6cm}, clip]{images/inset_images/chair/orex/chair_recons00.png}};
     %    \end{tikzpicture}

     %    &
        
     %    % \includegraphics[width=0.3\textwidth]{inset_images/sofa/sofa_inset_gcn_model.png}
     %     \begin{tikzpicture}
     %    \node (n0)  {\includegraphics[width=0.3\textwidth]{images/inset_images/sofa/orex_screenshot_recons_heatmap01.png};
     %    \node[draw=black, below right=of n0.north west] at (-3,2.7)
     %        {\includegraphics[width=0.15\textwidth,trim={7cm 8cm 7cm 9cm}, clip]{images/inset_images/sofa/orex_screenshot_recons00.png}};
     %    \end{tikzpicture}
     %    \\
        & $d_H: 0.022027$ & $d_H: 0.049031$ & $d_H: 0.015033$ \\

    \rotatebox{90}{\textbf{Ours}} & 

        \begin{tikzpicture}
        \node (n0)  {\includegraphics[width=0.3\textwidth]{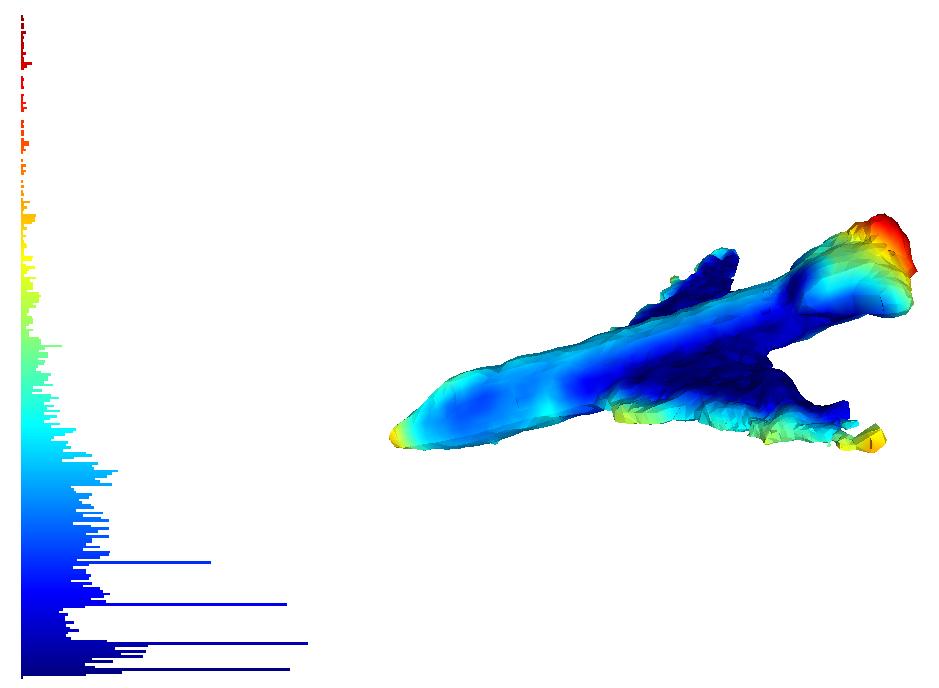}};
        \node[draw=black, below right=of n0.north west] at (-3,2.7)
            {\includegraphics[width=0.15\textwidth,trim={10cm 8cm 10cm 8cm}, clip]{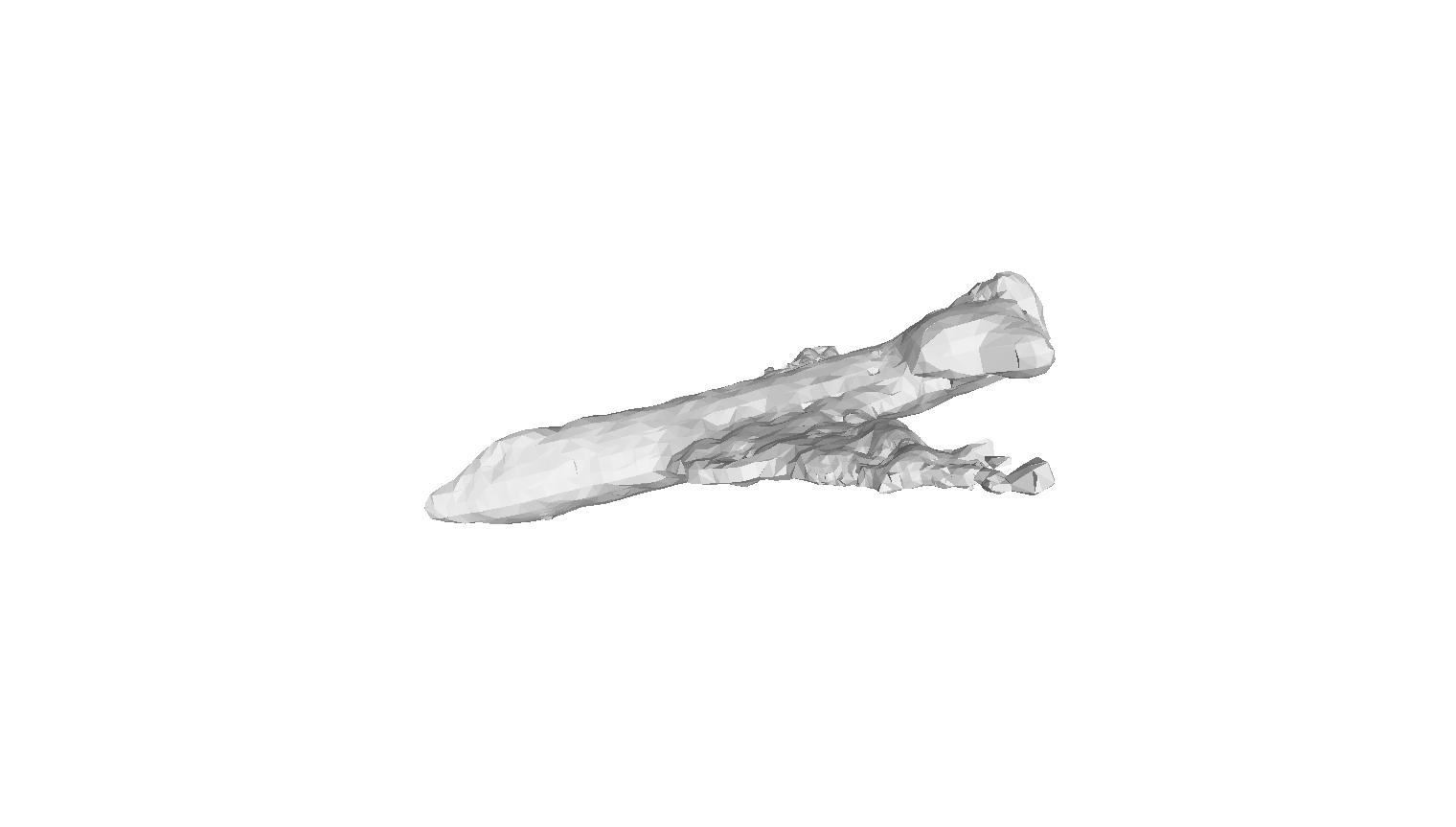}};
        \end{tikzpicture}
        
        &
         \begin{tikzpicture}
        \node (n0)  {\includegraphics[width=0.28\textwidth]{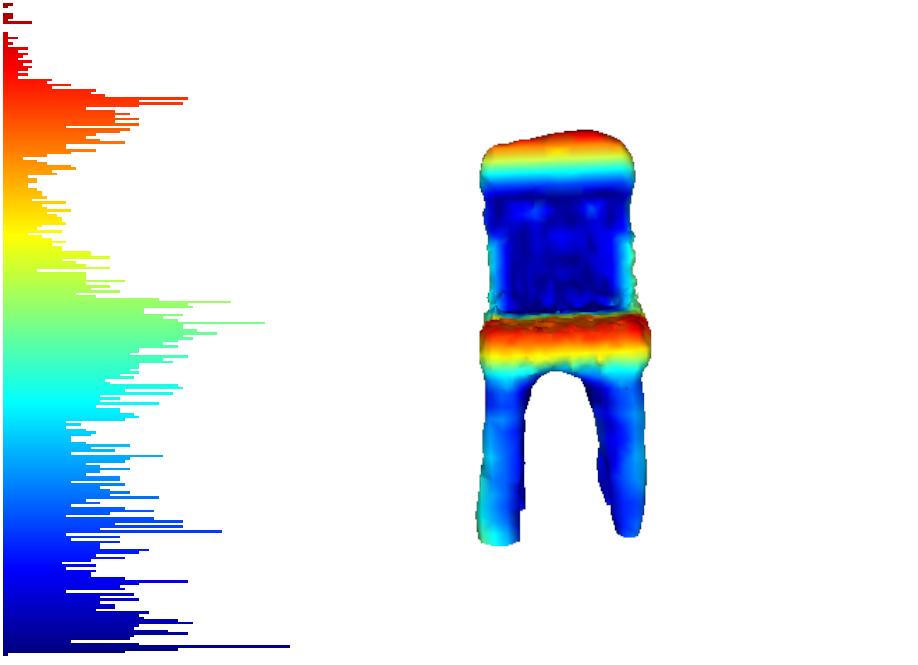}};
        \node[draw=black, below right=of n0.north west] at (-3,2.7)
            {\includegraphics[width=0.08\textwidth,trim={15cm 7cm 15cm 6cm}, clip]{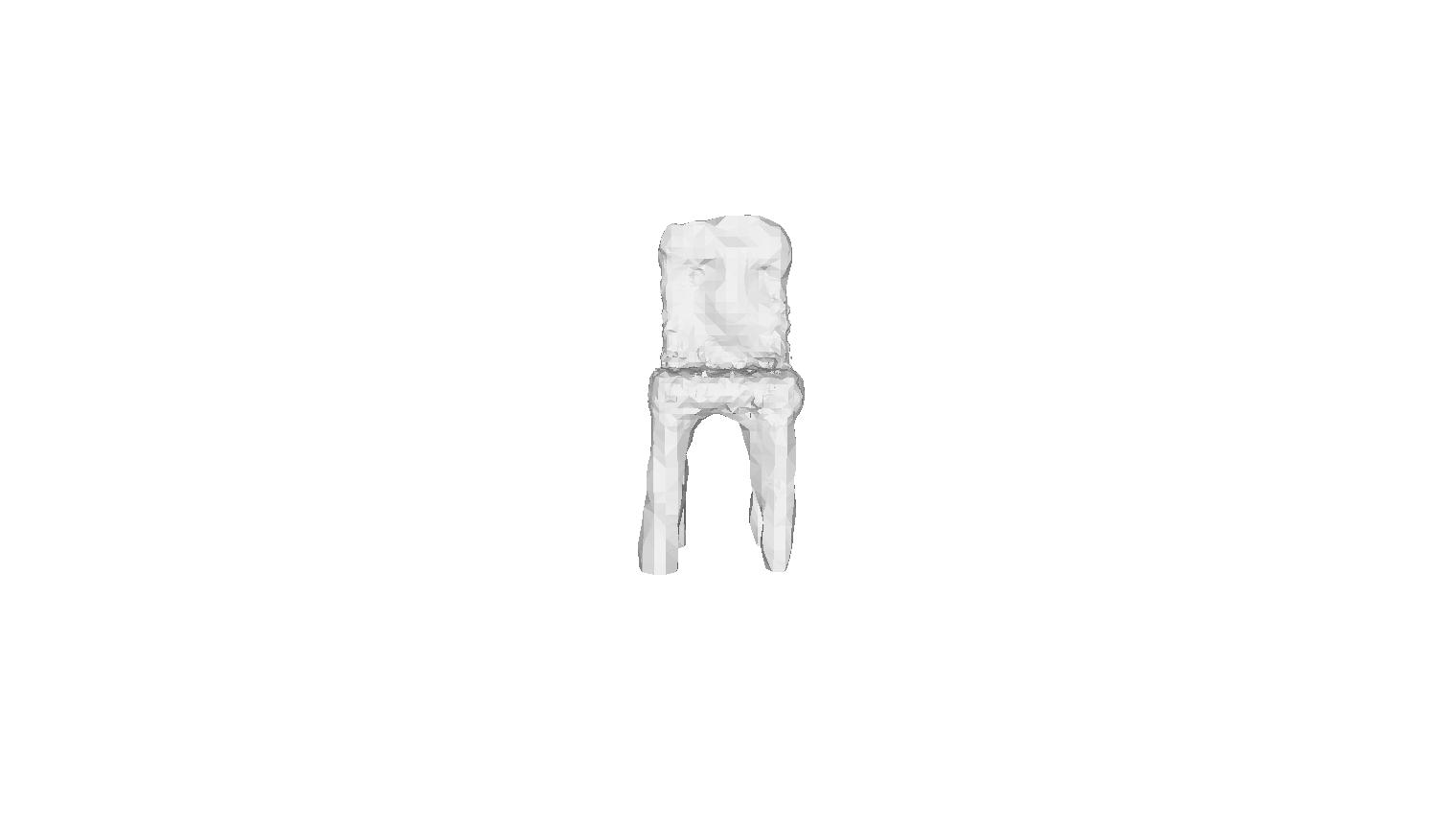}};
        \end{tikzpicture}

        &

         \begin{tikzpicture}
        \node (n0)  {\includegraphics[width=0.3\textwidth]{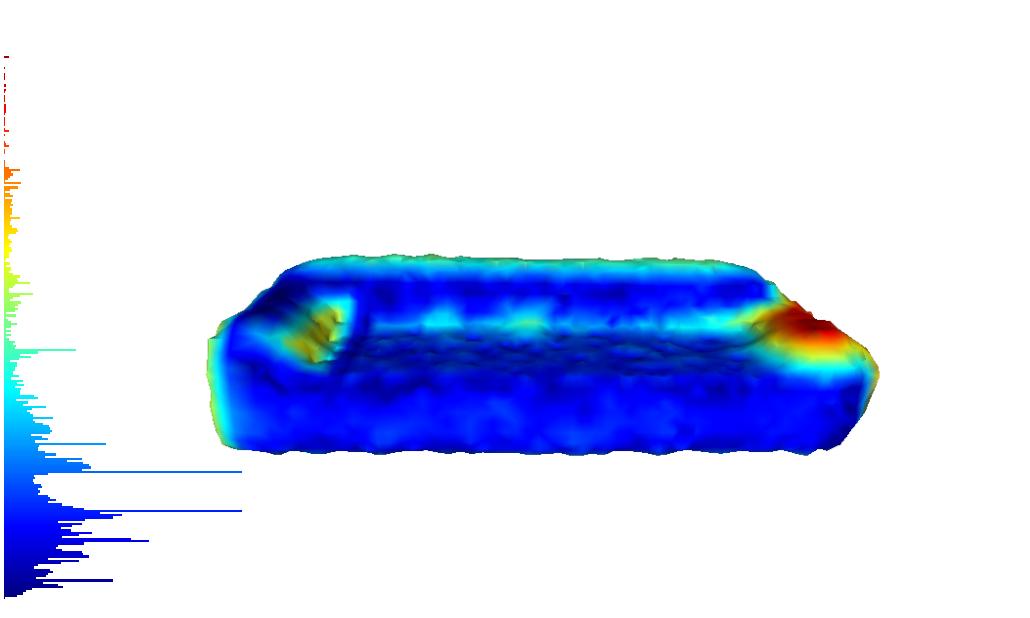}};
        \node[draw=black, below right=of n0.north west] at (-3,2.7)
            {\includegraphics[width=0.15\textwidth,trim={7cm 8cm 7cm 9cm}, clip]{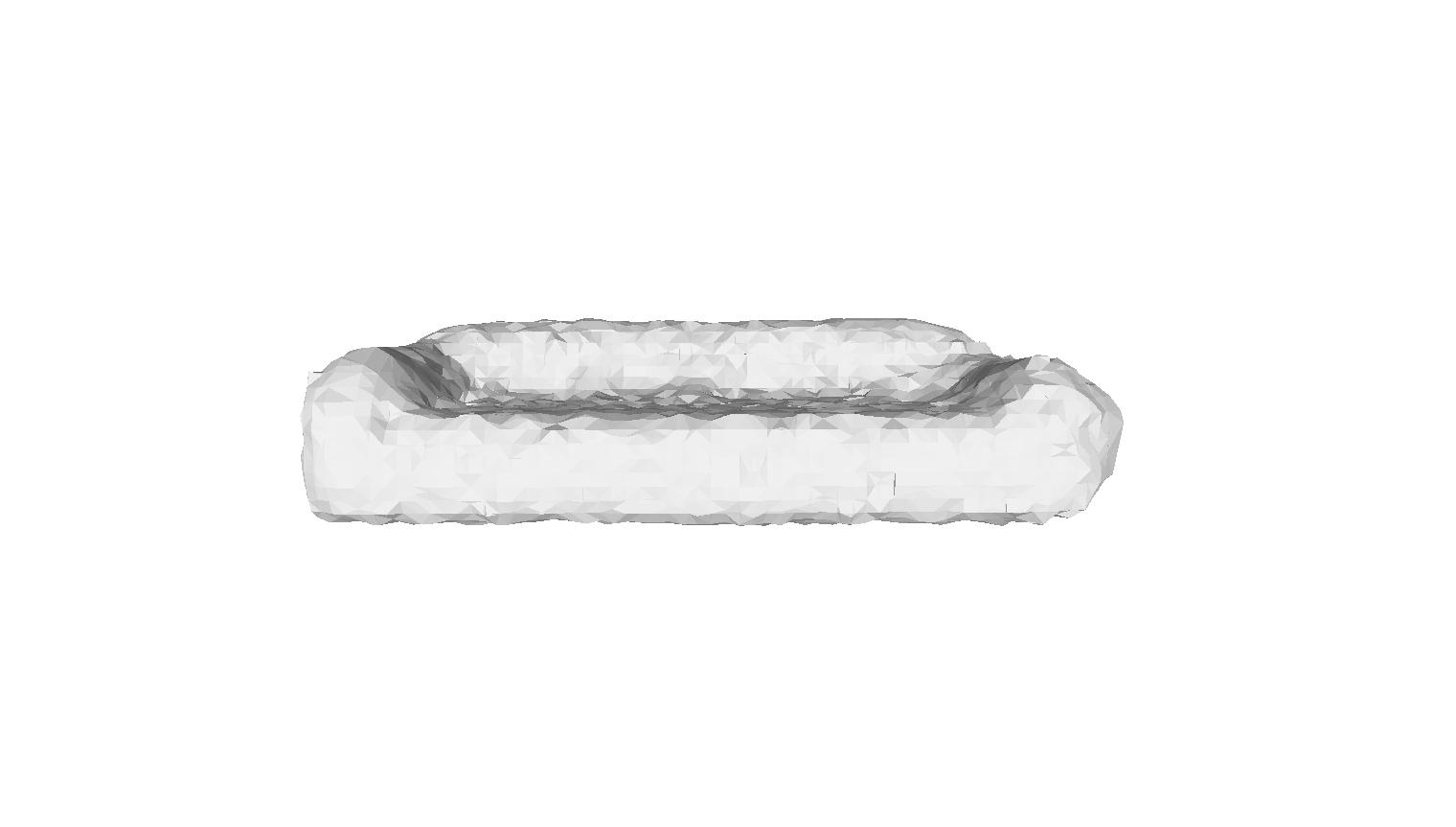}};
        \end{tikzpicture}
        \\
        & $d_H: 0.032111$ & $d_H: 0.049031$ & $d_H: 0.019826$ \\

    \end{tabular} 
    }
    \caption{Comparison with the state-of-the-art methods. Inset shows point-wise surface error compared with the GT.}
    \label{fig:heatmap}
\end{figure*}

We compare our method against 4 methods: 
% Poisson Surface Reconstruction \cite{kahzdan2006poisson} implemented in Open3D \cite{Zhou2018_open3d},  
VIPSS method for variational surface reconstruction from cross-sections \cite{huang2019variational},
surface reconstruction from non-parallel curve networks \cite{liu2008surface},  a state of the art deep learning based method P2P-Net \cite{yin2018p2p} and the recent ORex \cite{sawdayee2022orex} and show results in Figure ~\ref{fig:heatmap}.
Most of these methods suffer from holes and instabilities for sparse cross-sections; therefore, to be fair, we sample more cross-sections in those cases. However, we restrict our method to 10 cross-sections. VIPSS, ORex, and Liu's methods require careful sampling and sometimes tend to fail randomly for sparse cross-sections. We show the best-case results for these methods.
VIPSS is very sensitive to $\lambda$ and requires a large number $\sim 80$ of cross-sections for faithful reconstruction due to failure due to openness in cross-sections; however, we still notice artifacts.
%With VIPSS we use the same points and test for different $\lambda$ values: $\{0,0.01, 0.1\}$ and show the best results. 
We checked the reconstruction with the method proposed in \cite{liu2008surface}, however, the available implementation discards many cross-sections that lead to incorrect results. 
We show results for the cases where we did not observe this issue for a fair comparison. For Orex \cite{sawdayee2022orex} as well, we observe that it performs really well when cross-sections are dense; however, it fails in the case of sparse cross-sections. Therefore, for some samples, we show results in cases where it performs reasonably well.  

We further compare our method against a state-of-the-art deep learning-based method called P2P-Net. We modify P2P Net and train it on points sampled from our cross-sections. We notice that in some cases, despite performing better in terms of metrics, there are still completion issues in several samples, such as the chair shown in Figure ~\ref{fig:heatmap}. 
%This can be attributed to the smoother latent space produced by our method leading to more complete figures. 
Our method generates symmetric structures leading to higher loss value but better perception quality and semantically correct different structures such as the right-hand rest of the sofa and missing leg in the chair.
This also highlights a weakness of our method pertaining to the lack of strict adherence to the cross-sections since our method relies on embedding decoded by the pre-trained decoder. However, we believe that can be circumvented by better pre-training schemes since the performance of the pre-trained decoder forms the lower bound of the reconstruction error and can be swapped with any of the better-performing point cloud generators.

In order to visualize the error in reconstruction from our method and P2P-Net, we perform surface meshing of our resulting point cloud with Poisson Reconstruction \cite{kahzdan2006poisson}
, by computing normals from the ground truth mesh for the best-case-scenario. % and use the resultant mesh to compute the error histograms as a measure for the best-case-scenario. 
For VIPSS, we also modified the method and provided normals from the GT mesh. We show similar histograms for the surfaces obtained from other methods. 
We also note the Hausdorff distance ($d_H$) obtained for different methods in Figure ~\ref{fig:heatmap}. 
We notice that during the generation of the point cloud, since our method does not have hard constraints for precise overlap with input, the shift in point cloud can lead to a relative rise in the Hausdorff distance, as can be seen in the case of the chair in Figure ~\ref{fig:heatmap}. However, it outperforms the other methods in both qualitative and quantitative comparisons in several cases.

% \vspace{-9mm}
% \vspace{-10pt}
\section{Conclusion and Future Scope}
With this work, we open a new direction for the exciting domain of cross-section-based reconstruction. We generate a new dataset that can be used for multiple tasks. The ability to use parametric cross-sections directly in a learning-based setting exempts the use of any sampling-based restrictions in deep learning-based methods.
%; most previous works with point clouds take points in $\mathcal{R}^3$ as inputs leading to a restriction in the network capacity which depends on the number of input points being provided. 
The complete information of the curve is encapsulated in the coefficients of the parametric representation. Further, we utilize GCNs at scale and demonstrate their effectiveness for parametric curves and the ability of the GCNs to capture neighborhood information, which helps deduce better relationships among the cross-sections using attention, adding to the explainability with the flexibility to use any models trained on point cloud generation. We show empirical evidence to analyze the changes in reconstruction, both in terms of the embedding space representation and point cloud reconstruction, to understand the changes with respect to the variation in the amount of information provided to the network. This builds a strong motivation and opens up the field to further research such as the disentanglement of latent features and information-theoretic aspect of cross-section-based reconstruction which we hope to cover in future works. 

\pagebreak
%%%%%%%%% REFERENCES
{\small
\bibliographystyle{ieee_fullname}
\bibliography{egbib}
}

\end{document}